\documentclass{article} 
\PassOptionsToPackage{numbers,sort&compress}{natbib}
\usepackage{iclr2026_conference,times}
\setcitestyle{numbers,square,comma}

\usepackage{amsmath,amsfonts,bm}

\def\eqref#1{equation~\ref{#1}}

\def\1{\bm{1}}

\DeclareMathAlphabet{\mathsfit}{\encodingdefault}{\sfdefault}{m}{sl}
\SetMathAlphabet{\mathsfit}{bold}{\encodingdefault}{\sfdefault}{bx}{n}

\usepackage{wrapfig}
\usepackage{needspace}
\usepackage{placeins}
\usepackage{multicol}
\usepackage{hyperref}
\usepackage{url}
\usepackage[T1]{fontenc}    
\usepackage{hyperref}       
\usepackage{url}            
\usepackage{booktabs}       
\usepackage{amsfonts}       
\usepackage{nicefrac}       
\usepackage{microtype}      
\usepackage{xcolor}         

\usepackage{dsfont}
\usepackage{multirow}
\usepackage[ruled]{algorithm2e}
\usepackage{graphicx}
\usepackage{amsmath}
\usepackage{appendix}
\usepackage{color}
\usepackage{colortbl}
\usepackage{enumitem}
\usepackage{tabularx}
\usepackage{longtable}
\usepackage{adjustbox}
\usepackage{array}
\usepackage{listings}
\usepackage{pifont}
\usepackage[most]{tcolorbox}

\newcommand{\yesmark}{\ding{51}}
\newcommand{\nomark}{\ding{55}}

\definecolor{mydarkblack}{rgb}{0,0,0}
\definecolor{mydarkgreen}{rgb}{0,1,0}
\definecolor{fbApp}{HTML}{c8e7fa}
\definecolor{codebackground}{RGB}{248,248,248}
\definecolor{codeframe}{RGB}{205,205,205}
\definecolor{planone}{RGB}{0,114,178}
\definecolor{plantwo}{RGB}{230,159,0}
\definecolor{planthree}{RGB}{0,135,105}
\definecolor{planfour}{RGB}{213,94,0}
\definecolor{planfive}{RGB}{117,112,179}
\definecolor{plansix}{RGB}{180,80,135}

\lstdefinestyle{appendixcode}{
	basicstyle=\ttfamily\scriptsize,
	backgroundcolor=\color{codebackground},
	frame=single,
	rulecolor=\color{codeframe},
	framerule=0.4pt,
	breaklines=true,
	breakatwhitespace=false,
	columns=fullflexible,
	keepspaces=true,
	showstringspaces=false,
	xleftmargin=0.4em,
	xrightmargin=0.4em,
	aboveskip=0.6em,
	belowskip=0.8em,
	captionpos=b
}

\hypersetup{
	colorlinks=false,
	citecolor=mydarkgreen,
	linkcolor=mydarkblack
}
\newcommand{\benchmarkname}{BrainBench}
\newcommand{\numcases}{172}
\newcommand{\numinstances}{4K}
\newcommand{\nummodels}{13}
\newcommand{\benchmarklogo}{%
	\includegraphics[height=1.8cm,trim=250 120 180 100,clip]{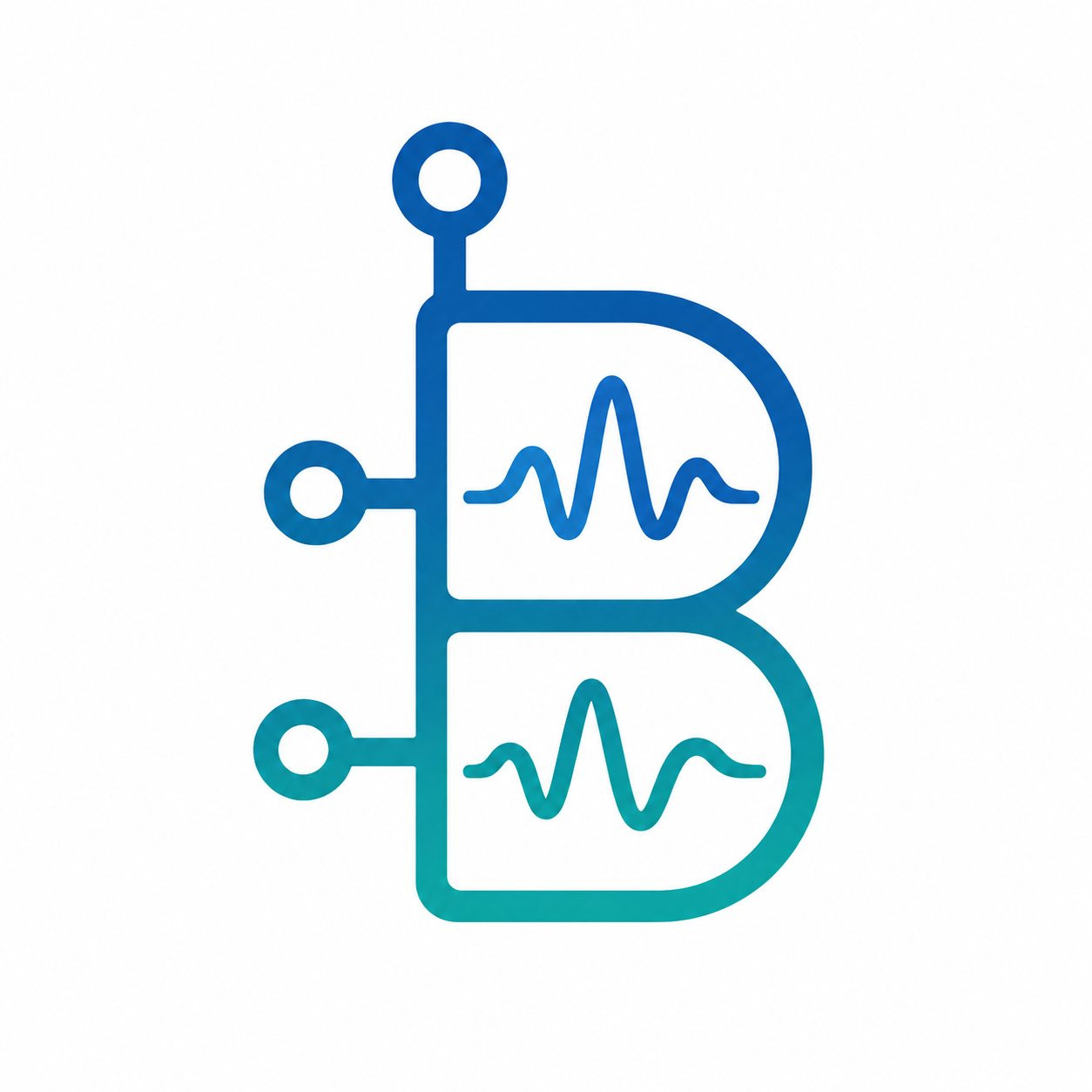}%
}

\title{%
	\begin{minipage}[c]{0.14\textwidth}
		\centering
		\benchmarklogo
	\end{minipage}%
	\begin{minipage}[c]{0.84\textwidth}
		\raggedright
		\benchmarkname: Benchmarking Large Language Models for Comprehensive EEG Understanding
	\end{minipage}%
}

\author{
	Yangxuan Zhou\textsuperscript{\rm 1,\rm 2},
	Yuning Chen\textsuperscript{\rm 1,\rm 2},
	Chen Wu\textsuperscript{\rm 1,\rm 2},
	Jiquan Wang\textsuperscript{\rm 1},
	Shijian Li\textsuperscript{\rm 1}, \\
	\textbf{Gang Pan\textsuperscript{\rm 1,\rm 2,\rm 3}},
	\textbf{Sha Zhao\textsuperscript{\rm 1,\rm 2}}\thanks{Corresponding authors.}\\
	\textsuperscript{\rm 1}State Key Laboratory of Brain-machine Intelligence, Zhejiang University\\
	\textsuperscript{\rm 2}College of Computer Science and Technology, Zhejiang University\\
	\textsuperscript{\rm 3}MOE Frontier Science Center for Brain Science and Brain-machine Integration, Zhejiang University\\
	\texttt{\{zyangxuan, szhao, yuningchen, chen\_wu\_, wangjiquan\}@zju.edu.cn;}\\
	\texttt{\{shijianli, gpan\}@zju.edu.cn;}\\
}

\iclrfinalcopy
\begin{document}

\maketitle
\fancyhead{}
\renewcommand{\headrulewidth}{0pt}

\begin{abstract}
	Electroencephalography (EEG) analysis extends beyond assigning predefined labels to recordings; it requires workflows connecting natural-language instructions, signal processing, quantitative evidence, and scientific interpretation. We term this capability \emph{comprehensive EEG understanding}. Existing evaluations, however, primarily target isolated decoding tasks or system-specific demonstrations, leaving the competence of large language models (LLMs) insufficiently quantified. We introduce \benchmarkname{}, a unified benchmark for comprehensive, instruction-conditioned EEG understanding. It comprises four subsets---Foundational Analysis, Sleep Assessment, Neurocognitive Assessment, and Physiological Integration---covering 17 datasets, \numcases{} tasks, and over \numinstances{} real-data instances. Given an instruction and EEG recordings with optional physiological signals, a system must perform the analysis and produce a scientifically grounded report and, when required, artifacts. Outputs are assessed through numerical, categorical, set, sequence, semantic, and artifact validation. We evaluate \nummodels{} representative LLMs across more than 100K executions under two paradigms: autonomous code execution with CodeAct and structured agentic analysis with BrainAgent. Results vary substantially across models, subsets, difficulty levels, and execution paradigms, showing that EEG competence depends on the model and its operationalization. \benchmarkname{} provides a reproducible testbed for advancing LLM-based EEG understanding.
	\par\medskip
	\begingroup
	\small\raggedright
	\hypersetup{pdfborder={0 0 0}}
	\noindent\textbf{Data:} \href{https://huggingface.co/datasets/xbb083/BrainBench}{\textcolor[RGB]{0,102,204}{\nolinkurl{https://huggingface.co/datasets/xbb083/BrainBench}}}\\
	\noindent\textbf{Code:} \href{https://github.com/xiaobaben/BrainBench}{\textcolor[RGB]{0,102,204}{\nolinkurl{https://github.com/xiaobaben/BrainBench}}}\\
	\noindent\textbf{Website:} \href{https://ceceliawai.github.io/BrainBenchWebsite/}{\textcolor[RGB]{0,102,204}{\nolinkurl{https://ceceliawai.github.io/BrainBenchWebsite/}}}
	\endgroup
\end{abstract}

\section{Introduction}

Electroencephalography (EEG) provides a non-invasive window into brain activity with millisecond-scale temporal resolution. Its accessibility and sensitivity to rapid neural dynamics have established it as a fundamental tool in neuroscience, sleep research, neurological assessment, and brain--computer interfaces \cite{aeschbach1993all, schalk2004bci2000, jeong2004eeg, da2013eeg}. Yet, computational EEG research has predominantly treated signal analysis as a decoding problem, in which a model maps a recording to a predefined output, such as a sleep stage, cognitive state, or clinical label \cite{wang2025cbramod, zhou2025personalized, zhou2026csbrain, el2026reve}. While this paradigm has driven substantial progress, it captures only a narrow component of real-world EEG analysis. In practice, EEG analysis requires a system to understand the analytical objective, reason coherently over real recordings, and arrive at a scientifically grounded conclusion \cite{keil2014committee, kane2017revised, pernet2020issues}. We refer to this broader capability as \textbf{comprehensive EEG understanding}. The central question therefore shifts from whether a model can predict a predefined target to whether it can carry an EEG analysis coherently from instruction to conclusion.

Large language models (LLMs) and agents offer a natural foundation for comprehensive EEG understanding \cite{brown2020language, yao2022react}. By combining language understanding, reasoning, and code generation, they can translate natural-language instructions into executable analyses, potentially transforming EEG analysis from a collection of specialized pipelines into an interactive, general-purpose process. Recent LLM-powered systems in brain science and neurotechnology have demonstrated their potential to support increasingly complex scientific workflows \cite{zhao2026eeg, abdou2026eeg, kosmyna2026neuroskill, chen2026embracing, xu2026sleeplm, wang2026neuroweaver, zhou2026brainagent, li2026brainpilot, pradeepkumar2026neural, gucerebragloss}. 
For example, CELM and CerebraGloss explore language-based clinical EEG interpretation and report generation \cite{pradeepkumar2026neural, gucerebragloss}, while BrainAgent investigates multi-agent orchestration for automating end-to-end EEG analysis \cite{zhou2026brainagent}. However, these systems are still evaluated primarily within their respective task- or system-specific settings. As a result, a comprehensive and systematic assessment of the correctness and scientific validity of LLM-generated EEG analyses across heterogeneous tasks and execution paradigms remains lacking. 
\begin{center}
	\emph{Can LLMs move beyond isolated decoding tasks to achieve \textbf{comprehensive EEG understanding} across real-world analysis workflows?}
\end{center}

Despite this promise, existing evaluation protocols are not designed to quantify comprehensive EEG understanding. Existing EEG benchmarks remain largely centered on fixed decoding objectives \cite{wu2025adabrain, xiong2025eeg, shen2026brain4fms, lu2026omnieeg, kontras2026neuroatlas}, rather than assessing whether a system can interpret diverse analytical instructions and carry them through to scientifically grounded conclusions. Moving beyond this decoding-centric paradigm, HeaRTS \cite{li2026hearts} takes an important step by requiring LLMs to analyze real-world health time-series data through executable code. Its primary emphasis, however, is breadth across physiological modalities rather than depth within EEG analysis. The distinctive signal characteristics, analytical conventions, and domain-specific interpretations of EEG give rise to heterogeneous workflows that demand fine-grained evaluation of both analytical procedures and resulting conclusions. Consequently, a unified benchmark for comprehensive EEG understanding remains absent.

Operationalizing comprehensive EEG understanding as a rigorous benchmark is challenging because the capability is inherently workflow-oriented and action-grounded. It requires a system to carry out a coherent sequence of file inspection, signal processing, quantitative analysis, comparison, and scientific interpretation, while grounding its reasoning in executable analyses of real recordings. Capturing this capability therefore requires diverse datasets, analytical tasks, instruction formats, and evaluation dimensions under unified and comparable protocols for execution and assessment.

\begin{figure*}[!tb]
	\centering
	\includegraphics[width=1.0\textwidth]{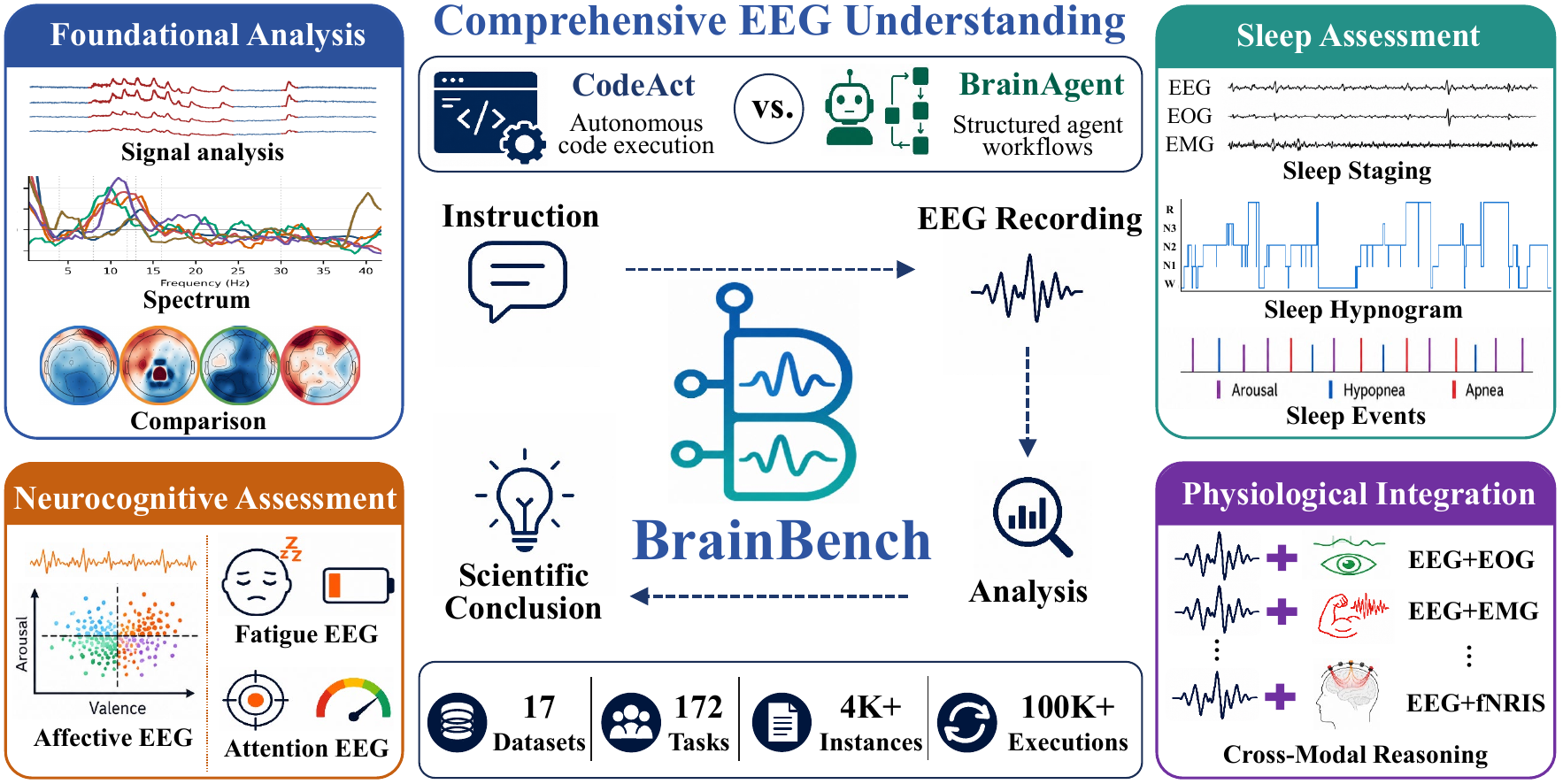}
	\caption{\textbf{Overview of BrainBench}. We present a comprehensive benchmark for instruction-conditioned EEG understanding, spanning 17 datasets, 172 tasks, and over 4K real-data instances across four complementary subsets: Foundational Analysis, Sleep Assessment, Neurocognitive Assessment, and Physiological Integration. }
	\label{fig:intro}
	\vspace{-5pt}
\end{figure*}

To address these challenges, we introduce \textbf{\benchmarkname{}}, a benchmark for comprehensive, instruction-conditioned EEG understanding. It comprises four complementary subsets---Foundational Analysis, Sleep Assessment, Neurocognitive Assessment, and Physiological Signal Integration---covering \textbf{\numcases{} tasks} and \textbf{\numinstances{} instances} drawn from \textbf{17 datasets} and spanning diverse analytical objectives and levels of complexity. Together, these subsets cover foundational signal analysis, domain-specific reasoning, and multimodal interpretation, enabling EEG understanding to be evaluated in both breadth and depth. Rather than assessing performance on a fixed predictive target, \benchmarkname{} evaluates whether a model can transform a natural-language instruction and real EEG recordings into a scientifically grounded conclusion. We further evaluate the same LLMs under two complementary execution paradigms. CodeAct \cite{wang2024executable} allows an LLM to autonomously plan and execute analyses through self-generated code, providing a relatively open setting for eliciting its analytical capability. In contrast, BrainAgent \cite{zhou2026brainagent} is an EEG-oriented multi-agent framework that executes analyses through structured workflows, controlled tools, and traceable intermediate actions. By holding the instructions, data, and evaluation criteria constant, \benchmarkname{} enables a controlled comparison of two distinct execution paradigms and examines the trade-off between the flexibility of autonomous code generation and the reliability, auditability, and operational safety of structured agentic analysis.

We evaluate \textbf{\nummodels{}} representative LLMs across all four subsets and under both execution paradigms, revealing key gaps in current systems and important directions for advancing comprehensive EEG understanding in the future. Beyond the current release, \benchmarkname{} is designed for extensibility, allowing new EEG domains, datasets, task formulations, evaluation dimensions, and models to be incorporated as the field evolves. Our contributions are summarized as follows:

\begin{itemize}
	\item We formulate comprehensive, instruction-conditioned EEG understanding as a unified evaluation target and introduce \benchmarkname{}, comprising four complementary subsets, \numcases{} tasks, and \numinstances{} instances across 17 diverse datasets, analytical objectives, and workflows.
	
    \item We establish a dual-paradigm protocol that compares autonomous code execution with a
    structured EEG agent under shared instructions, data, and ground truth, enabling controlled
    analysis of how the execution paradigm shapes measured EEG competence.
	
    \item We systematically evaluate \nummodels{} representative LLMs across tasks and execution paradigms, providing a multidimensional characterization of their capabilities and revealing key limitations and directions for advancing comprehensive EEG understanding.
\end{itemize}

\section{Related Work}
\subsection{EEG Models and Decoding Benchmarks}

EEG modeling has evolved from handcrafted feature pipelines and task-specific neural networks to foundation models trained on large-scale recordings to learn transferable representations \cite{lawhern2016eegnet, jiang2024large, wang2025cbramod, zhou2026csbrain, el2026reve, xiao2026brainomni, wang2025eegmamba, wang2026deeperbrain}. In parallel, existing benchmarks have standardized evaluation across downstream decoding tasks, typically using linear probing or fine-tuning to assess representation quality and cross-subject or cross-dataset generalization \cite{wu2025adabrain, xiong2025eeg, shen2026brain4fms, lu2026omnieeg, kontras2026neuroatlas, wang2026eeg, kastrati2025eeg, banville2026neuralbench}. Clinically grounded models have further extended this paradigm to full-session recordings and multimodal clinical context; for example, CLEF \cite{cao2026clef} aligns session-level EEG representations with neurologist reports and electronic health records, yet is still evaluated mainly through patient-level classification tasks. Thus, despite increasing scale and clinical relevance, existing evaluations continue to define EEG competence primarily through performance on predefined predictive targets. They do not assess whether a general-purpose model can interpret an open-ended analytical instruction, execute the required analysis on real recordings, and derive a scientifically grounded conclusion. This leaves open the need for evaluating comprehensive, instruction-conditioned EEG understanding beyond fixed-label decoding.

\subsection{Large Language Models for EEG Analysis}

The reasoning \cite{yao2022react}, planning \cite{shinn2023reflexion}, and tool-use \cite{schick2023toolformer} capabilities of large language models have enabled a new class of systems that interact with EEG data through natural-language instructions. These systems combine language understanding with code generation, domain knowledge, and specialized analytical tools to support tasks ranging from data inspection and preprocessing to event detection and report generation \cite{zhao2026eeg, kosmyna2026neuroskill, abdou2026eeg, chen2026embracing, xu2026sleeplm, wang2026neuroweaver, baradari2025neurochat, zhou2026brainagent, li2026brainpilot, hong2024chatbci}. For example, EEGAgent \cite{zhao2026eeg} coordinates specialized tools for automated EEG analysis and reporting, while SleepLM \cite{xu2026sleeplm} connect physiological recordings with natural-language sleep assessment. More recent agentic frameworks extend this paradigm toward longer and more heterogeneous workflows: NeuroWeaver \cite{wang2026neuroweaver} autonomously explores EEG analysis pipelines. BrainAgent \cite{zhou2026brainagent} decomposes user intent among specialized agents for structured brain-signal analysis, and BrainPilot \cite{li2026brainpilot} explores autonomous brain-science discovery through a multi-agent framework. Collectively, these studies demonstrate the feasibility of transforming high-level user requests into executable EEG analyses. Their rapid development, however, raises a distinct evaluation question: \textbf{how reliably do the resulting analyses reflect both the supplied recordings and the scientific intent expressed in the instruction?}

\subsection{Evaluating Large Language Models for EEG Understanding}

Existing evaluations differ substantially in scope and target. EEGAgent \cite{zhao2026eeg} combines task-level metrics with qualitative demonstrations of EEG interpretation and reporting. BrainAgent \cite{zhou2026brainagent} evaluates 60 tasks across three difficulty levels, focusing on task completion, routing reliability, and tool-use efficiency. BrainPilot \cite{li2026brainpilot} broadens the scope to brain-science research, although the initial BrainPilotBench-v0 contains four tasks covering calcium imaging, fMRI, motor-imagery EEG, and sleep EEG. HeaRTS \cite{li2026hearts} provides the closest general benchmark for executable reasoning over physiological time series, covering diverse health domains and signal modalities through autonomous code execution, but its primary objective is breadth across health time series rather than analytical depth within EEG. Taken together, existing evaluations remain fragmented across system-specific tasks, metrics, and protocols. A comprehensive and systematic assessment of the correctness and scientific validity of LLM-generated EEG analyses across heterogeneous tasks, datasets, and execution paradigms therefore remains lacking.

\section{\benchmarkname}
To address this gap, we introduce {\benchmarkname{}}, a unified benchmark for comprehensive, instruction-conditioned EEG understanding. Given a natural-language instruction and real EEG recordings, \benchmarkname{} evaluates whether an LLM can produce scientifically grounded analysis outputs beyond predefined predictive tasks. It comprises four complementary subsets spanning {\numcases{} tasks}, {\numinstances{} instances}, and {17 datasets} as shown in Figure \ref{fig:intro}. Under a unified assessment protocol, we evaluate \textbf{\nummodels{}} representative LLMs with CodeAct \cite{wang2024executable} and BrainAgent \cite{zhou2026brainagent}, enabling controlled comparison between autonomous code execution and structured EEG-agent workflows.
\subsection{\benchmarkname Organization and Formulation}
\benchmarkname{} formulates comprehensive EEG understanding as instruction-conditioned analytical execution. The benchmark follows a three-level hierarchy of \texttt{subsets}, \texttt{tasks}, and \texttt{instances}. Four complementary \texttt{subsets} define its evaluation scope: Foundational Analysis targets general EEG operations; Sleep Assessment and Neurocognitive Assessment cover domain-specific analytical workflows; and Physiological Integration evaluates reasoning across EEG and complementary physiological signals. Within each subset, a \texttt{task} defines a reusable analytical objective, the expected outputs, and the corresponding assessment requirements, and is instantiated by a collection of evaluation \texttt{instances}. These \texttt{instances} preserve the task-level requirements while varying the underlying recordings, subjects, analysis parameters, or task-specific instructions. This hierarchy enables \benchmarkname{} to assess whether analytical capabilities generalize across heterogeneous inputs rather than merely succeed on individual recordings or predefined queries.
For an instance \(i\), let \(\mathcal{I}_i\) denote its natural-language instruction and \(\mathcal{D}_i\) the supplied input files. Given \((\mathcal{I}_i,\mathcal{D}_i)\), a target system \(\mathcal{M}\) produces a free-form analysis report \(\mathcal{R}_i\) and, when required, a set of artifacts \(\mathcal{A}_i\). The unified evaluation function $\mathcal{E}$ then assigns the instance score \(s_i\):
\begin{equation}
	(\mathcal{R}_i,\mathcal{A}_i)
	= \mathcal{M}(\mathcal{I}_i,\mathcal{D}i),
	\qquad
	s_i = \mathcal{E}(\mathcal{R}_i,\mathcal{A}i).
\end{equation}
The evaluator applies the assessment criteria defined by the corresponding task while retaining a common interface across the benchmark. This formulation accommodates heterogeneous EEG workflows and output formats, while ensuring that different execution paradigms are evaluated on the same tasks and instances under consistent requirements.
\subsection{\benchmarkname{} Construction and Evaluation}
\label{subsec:benchmark_construction}
\subsubsection{Task Definition}
\label{subsubsec:task_definition}
\begin{wrapfigure}{r}{0.42\textwidth}
	\vspace{-0.0\baselineskip}
	\centering
	\includegraphics[width=\linewidth]{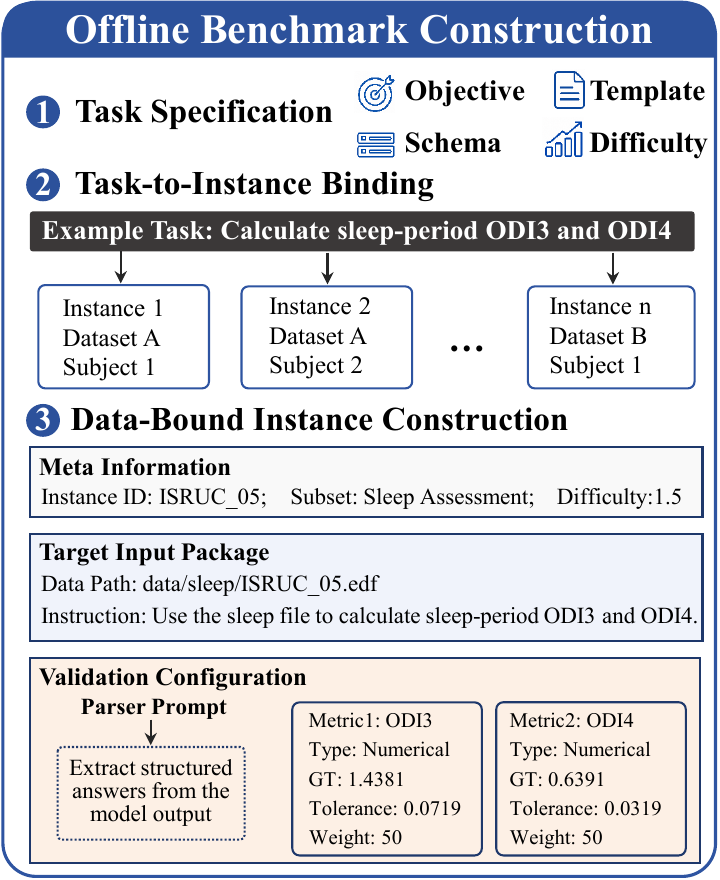}
	\caption{Construction of a reusable \texttt{task} and its data-bound \texttt{instances} across heterogeneous recordings.}
	\label{fig:instance_construction}
	\vspace{-2.8\baselineskip}
\end{wrapfigure}
In \benchmarkname{}, each \texttt{task} defines a  recording-independent analytical objective and the EEG understanding capability it is intended to assess. It also specifies the required inputs, expected outputs, and applicable validation criteria. These requirements remain fixed across all corresponding \texttt{instances}, allowing the same capability to be evaluated across different subjects, datasets, and recordings. Tasks are stratified into three difficulty levels according to their intrinsic analytical demands. \textbf{Easy} tasks involve short, explicitly specified analyses with conclusions derived directly from limited evidence. \textbf{Medium} tasks require multiple dependent steps and integration of intermediate results. \textbf{Hard} tasks involve extended workflows that synthesize evidence across channels, time scales or physiological modalities. Difficulty is assigned at the task level and inherited by all associated instances.
\subsubsection{Instance Construction}
\label{subsubsec:instance_construction}
A \texttt{task} is converted into executable evaluation examples by binding its specification to a concrete data context. Each resulting \texttt{instance} identifies the dataset and recording to be analyzed, together with the applicable time window, signal selection, analysis parameters, and other task-specific conditions. For each such binding, \benchmarkname{} generates a complete natural-language instruction that specifies the available input files, analysis scope, required outputs, and reporting constraints. This process preserves a consistent analytical objective and evaluation target across instances while adapting the instruction to the data available in each evaluation context. Shown in Figure \ref{fig:instance_construction}, each instance is further paired with an evaluator-side reference package. Given the bound input files and analysis parameters, a deterministic analysis script executes the prescribed reference workflow and computes the expected values. The resulting outputs are combined with the validation configuration, including the parser prompt and metric definitions, to form a complete instance-level specification for evaluation. Reusing a fixed script and parameterization across instances ensures that the ground truth remains reproducible only with the bound data.

\subsubsection{Multi-Unit Validation}
\label{sec:multi_unit_validation}
\begin{wrapfigure}{r}{0.42\textwidth}
	\vspace{-1.0\baselineskip}
	\centering
	\includegraphics[width=\linewidth]{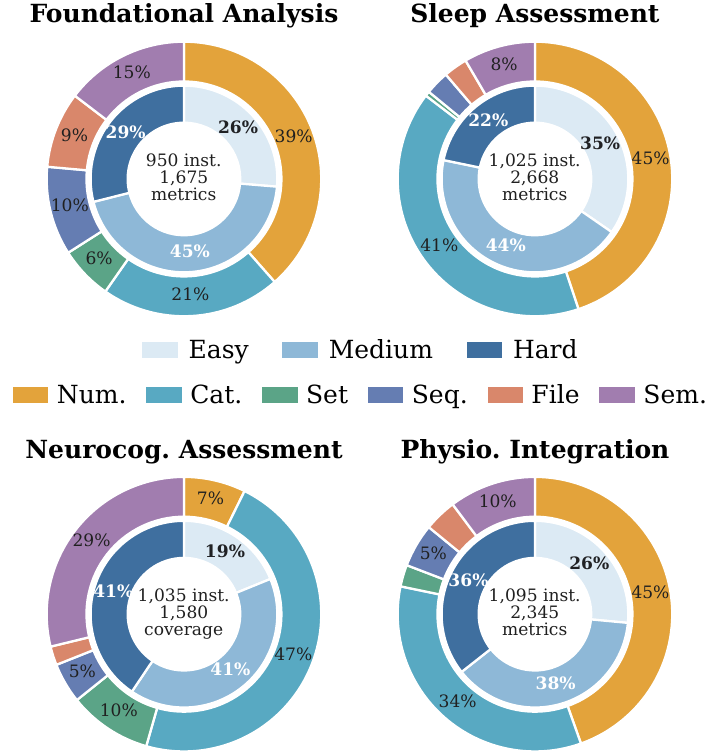}
	\caption{Distributions of task difficulty and validation units across  subsets.}
	\label{fig:subset_composition_nested_donut}
	\vspace{-0.4\baselineskip}
\end{wrapfigure}
\benchmarkname{} supports free-form analytical reports to accommodate the diverse wording and presentation required by heterogeneous EEG tasks. To enable standardized evaluation, a Parser Agent extracts the required content from each report into a structured representation according to an instance-specific \emph{Validation Configuration}, which defines the target fields, extraction prompt, validation units, reference targets, and metric weights. It serves solely as an extractor: it returns null for missing information and neither evaluates scientific correctness nor infers unreported results. This design preserves flexible natural-language reporting without conflating EEG understanding with rigid format compliance.

The extracted fields, original report, and generated artifacts are evaluated through six complementary validation units. \textbf{Numerical validation} compares reported scalar values with the computed ground truth under task-specific tolerances. \textbf{Categorical validation} assesses discrete labels or choices after canonicalization. \textbf{Set validation} evaluates unordered collections using exact matching or element-level partial credit. \textbf{Sequence validation} extends this assessment to ordered outputs by considering both element coverage and positional consistency. \textbf{Semantic validation} uses a task-specific Semantic Judge to assess non-scalar conclusions for correctness, relevance, instruction consistency, and unsupported or hallucinatory claims. \textbf{Artifact validation} verifies whether requested files are successfully generated and valid; structured signal files are checked programmatically, while visual outputs can additionally be assessed by a VLM Judge. A single instance may combine multiple validation units to cover complementary aspects of the requested output.
\par\WFclear

For instance \(i\), each of its validation metrics produces a normalized score \(v_{im}\in[0,1]\) with a non-negative weight \(w_{im}\). For subset aggregation, \(d_i\in\{1.0,1.5,2.0\}\) denotes the task-level difficulty coefficient for Easy, Medium, and Hard instances, respectively. The instance and subset scores are computed as
\begin{equation}
	s_i=100 \times \frac{\sum_m w_{im}v_{im}}{\sum_m w_{im}},
	\qquad
	S_{\mathcal{S}}=\frac{\sum_{i\in\mathcal{S}}d_i s_i}{\sum_{i\in\mathcal{S}}d_i}.
\end{equation}
Both scores lie on a \(0\)--\(100\) scale, with the subset score assigning greater weight to more difficult tasks. Detailed matching rules are provided in Appendix~\ref{app:benchmark_construction_details}.
\subsubsection{Black-Box Evaluation}
\label{subsubsec:black_box_evaluation}

\benchmarkname{} adopts a black-box evaluation protocol in which each target system receives only the natural-language instruction and associated input files. The evaluator-side reference package and \emph{Validation Configuration} remain inaccessible during execution. Each instance is processed in an isolated container, and only the final report and requested artifacts contribute to the benchmark score. Execution actions, tool calls, execution traces, runtime errors, token usage, and latency are recorded separately for reproducibility and failure analysis. Each LLM is evaluated under two complementary execution paradigms through the same input--output interface. In \textbf{CodeAct} \cite{wang2024executable}, the model autonomously plans the analysis and generates and executes Python code in an interactive environment. In \textbf{BrainAgent} \cite{zhou2026brainagent}, the model performs the analysis through structured agent coordination and controlled tool invocation. To cover the evaluation capability boundary defined by \benchmarkname{}, we equip BrainAgent with a capability-oriented toolset comprising reusable EEG analysis operations rather than task- or instance-specific solutions. Holding the instructions, input data, and validation protocol fixed enables a controlled comparison between autonomous code execution and structured agentic analysis. More details are provided in the Appendix \ref{app:appendixd}.

\subsection{\benchmarkname{} Composition}
\benchmarkname{} comprises four complementary subsets that cover distinct dimensions of comprehensive EEG understanding (complete construction details for all subsets are provided in the Appendix \ref{app:task_inventory}):

\begin{itemize}[leftmargin=1.2em, labelsep=0.4em]
	\item \textbf{\textit{Foundational Analysis.}}
	This subset comprises 40 \texttt{tasks} and 950 \texttt{instances} derived from ISRUC \cite{khalighi2016isruc}, BCIC2020-3 \cite{jeong20222020}, SEED-V \cite{liu2021comparing}, Mumtaz2016 \cite{mumtaz2017electroencephalogram}, and MentalArithmetic \cite{zyma2019electroencephalograms}, together with a curated EEG and BCI knowledge collection. It evaluates general-purpose EEG capabilities spanning recording inspection, preprocessing, feature extraction, spatiotemporal comparison, connectivity analysis, artifact generation, and evidence-grounded interpretation.
	
	\item \textbf{\textit{Sleep Assessment.}}
	This subset comprises 43 \texttt{tasks} and 1,025 \texttt{instances} derived from HMC\cite{alvarez2020inter}, ISRUC \cite{khalighi2016isruc}, MASS-SS3 \cite{o2014montreal}, PhysioNet 2018 \cite{ghassemi2018you}, and SHHS-1 \cite{quan1997sleep}, together with a curated sleep-medicine knowledge collection. It evaluates sleep architecture quantification, staging and spectral interpretation, sleep-event detection, multimodal physiological analysis, and the calculation of clinically relevant whole-night indices.
	
	\item \textbf{\textit{Neurocognitive Assessment.}}
	This subset comprises 50 \texttt{tasks} and 1,035 \texttt{instances} derived from FACED \cite{chen2023large}, REFED \cite{ning2026refed}, COG-BCI \cite{hinss2023open}, and MPD-DF \cite{li2026multimodal}, together with curated domain-knowledge questions. It evaluates the understanding of affective states, cognitive workload, and fatigue through state recognition, feature-based comparison, temporal and within-subject analysis, multimodal physiological evidence integration, and scientifically grounded interpretation.
	
	\item \textbf{\textit{Physiological Integration.}}
	This subset comprises 39 \texttt{tasks} and 1,095 \texttt{instances} derived from SEED-VII \cite{jiang2024seed}, DEAP \cite{koelstra2011deap}, Simultaneous Dataset B \cite{shin2018simultaneous}, SEED-VIG \cite{zheng2016multimodal}, and SHHS-1 \cite{quan1997sleep}, together with a curated multimodal neurophysiology knowledge collection. It evaluates the integration of EEG with complementary physiological signals through temporal alignment, modality-specific feature extraction, cross-modal coupling and evidence fusion, signal-quality assessment, data repair, missing-modality reconstruction, and multimodal result generation.
\end{itemize}

\section{Experiments and Analysis}
\vspace{-5pt}
We evaluate 13 API-accessible LLMs spanning eight model families and multiple capability tiers. The evaluated models include the Qwen3.5 and Qwen3.7 series \cite{qwen2026qwen35,qwen2026qwen37max,qwen2026qwen37plus}, GLM-5 \cite{zeng2026glm}, DeepSeek-V4-Flash \cite{xu2026deepseek}, Kimi K2.5 \cite{team2026kimi}, MiniMax-M2.5 \cite{chen2026minimax}, GPT-5.6 Luna, GPT-5.6 Terra and GPT-5.6 Sol \cite{openai2026gpt56}, Gemini 3.6 Flash \cite{googledeepmind2026gemini36flash}, and Claude Opus 5 \cite{anthropic2026opus5}. To standardize test-time computation, all optional reasoning modes are disabled. Each model is evaluated under BrainAgent and CodeAct with matched API and decoding settings in isolated containers. Runs affected by verified provider or transport errors are retried according to Appendix \ref{app:appendixd}.
\begin{table*}[!t]
	\centering
	\caption{\textbf{\benchmarkname{} leaderboard.} Performance of evaluated models under BrainAgent and CodeAct across the four benchmark subsets. The best available result in each populated column is shown in \textbf{bold}, and the second-best results are \underline{underlined}.}
	\label{tab:brainbench_leaderboard}
	
	\begingroup
	\setlength{\tabcolsep}{3.5pt}
	\renewcommand{\arraystretch}{1.12}
	
	\newcommand{\scorehead}[1]{\makebox[3em][c]{#1}}
	\newcommand{\grouphead}[1]{%
		\multicolumn{2}{c}{\makebox[6em][c]{\textbf{#1}}}%
	}
	
	\resizebox{\textwidth}{!}{%
		\begin{tabular}{@{}l*{10}{c}@{}}
			\toprule
			\multirow{2}{*}{\textbf{Model}}
			& \grouphead{\shortstack{Foundational\\Analysis}}
			& \grouphead{\shortstack{Sleep\\Assessment}}
			& \grouphead{\shortstack{Neurocog.\\Assessment}}
			& \grouphead{\shortstack{Physiol.\\Integration}}
			& \grouphead{\shortstack{Overall\\Score}} \\
			\cmidrule(lr){2-3}
			\cmidrule(lr){4-5}
			\cmidrule(lr){6-7}
			\cmidrule(lr){8-9}
			\cmidrule(lr){10-11}
			& \scorehead{BA} & \scorehead{CA}
			& \scorehead{BA} & \scorehead{CA}
			& \scorehead{BA} & \scorehead{CA}
			& \scorehead{BA} & \scorehead{CA}
			& \scorehead{BA} & \scorehead{CA} \\
			\midrule
			
			\raisebox{-0.20\height}{\includegraphics[height=1.15em]{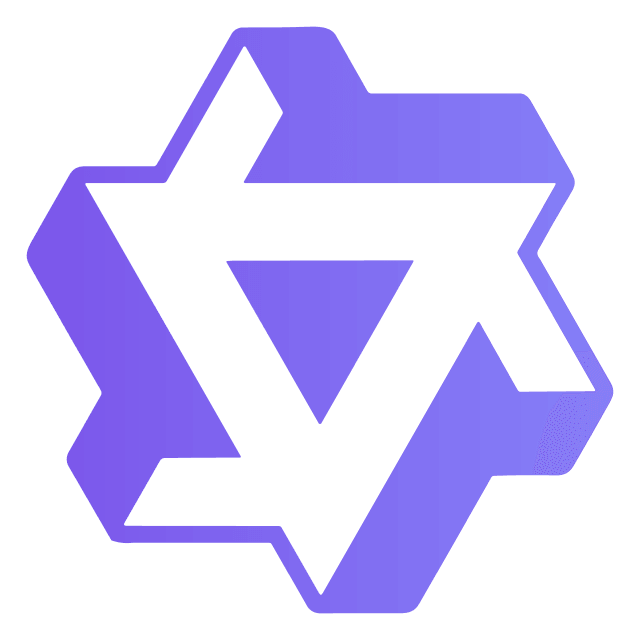}}~Qwen3.5-35B
			& 59.98 & 40.80 & 59.35 & 50.18 & 51.50 & 51.04 & 39.30 & 36.41 & 52.53 & 44.61 \\
			
			\raisebox{-0.20\height}{\includegraphics[height=1.15em]{fig/model_icons/qwen.png}}~Qwen3.5-122B
			& 63.12 & 48.85 & 65.98 & 57.18 & 58.51 & 56.16 & 49.38 & 47.09 & 59.25 & 52.32 \\
			
			\raisebox{-0.20\height}{\includegraphics[height=1.15em]{fig/model_icons/qwen.png}}~Qwen3.7 Plus
			& 73.08 & 68.33 & 68.08 & 62.06 & 69.86 & 64.17 & 57.98 & 51.66 & 67.25 & 61.55 \\
			
			\raisebox{-0.20\height}{\includegraphics[height=1.15em]{fig/model_icons/qwen.png}}~Qwen3.7 Max
			& 72.21 & 69.54 & 70.54 & 67.80 & 69.94 & \underline{65.13} & 66.40 & 60.40 & 69.77 & 65.72 \\
			
			\raisebox{-0.20\height}{\includegraphics[height=1.15em]{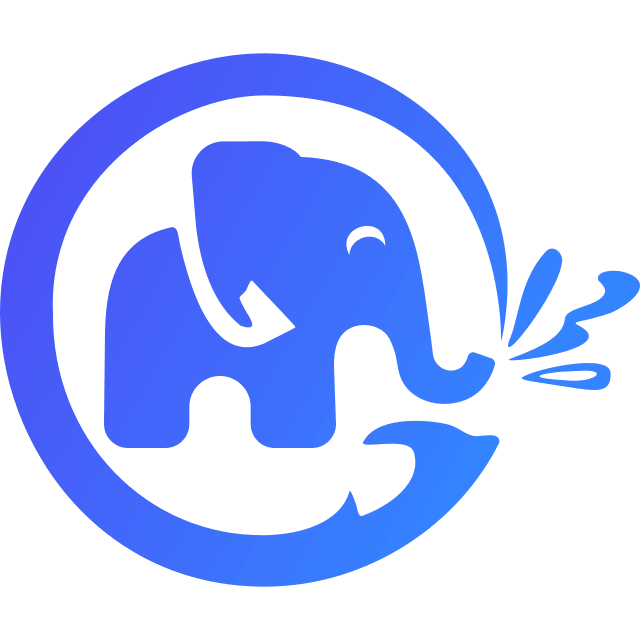}}~GLM 5
			& 68.92 & 70.77 & 66.17 & 64.65 & 63.56 & 65.12 & 55.04 & 56.76 & 63.42 & 64.33 \\
			
			\raisebox{-0.20\height}{\includegraphics[height=1.15em]{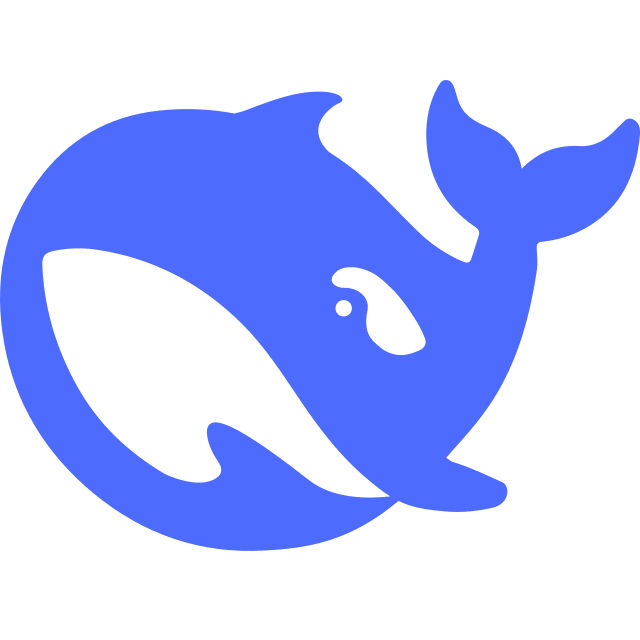}}~DeepSeek V4 Flash
			& 67.38 & 68.16 & 67.97 & 61.01 & 67.21 & 60.38 & 49.98 & 54.52 & 63.13 & 61.02 \\
			
			\raisebox{-0.20\height}{\includegraphics[height=1.15em]{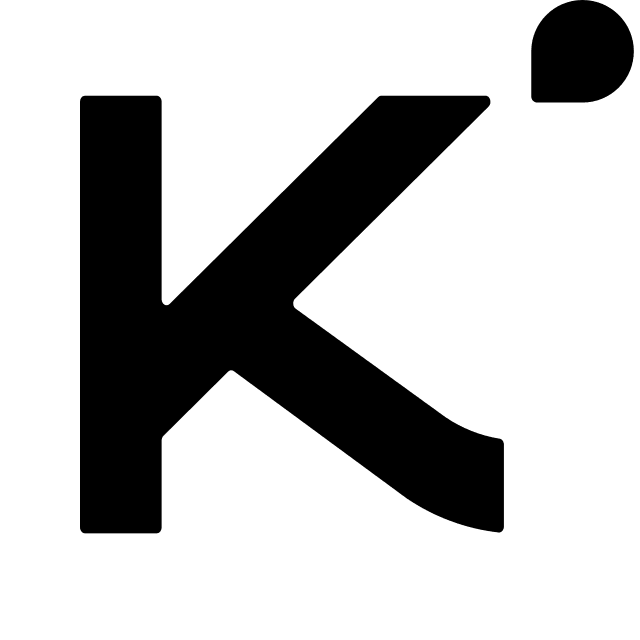}}~Kimi K2.5
			& 72.36 & 65.85 & 68.09 & 62.28 & 65.56 & 55.33 & 54.32 & 48.77 & 65.08 & 58.06 \\
			
			\raisebox{-0.20\height}{\includegraphics[height=1.15em]{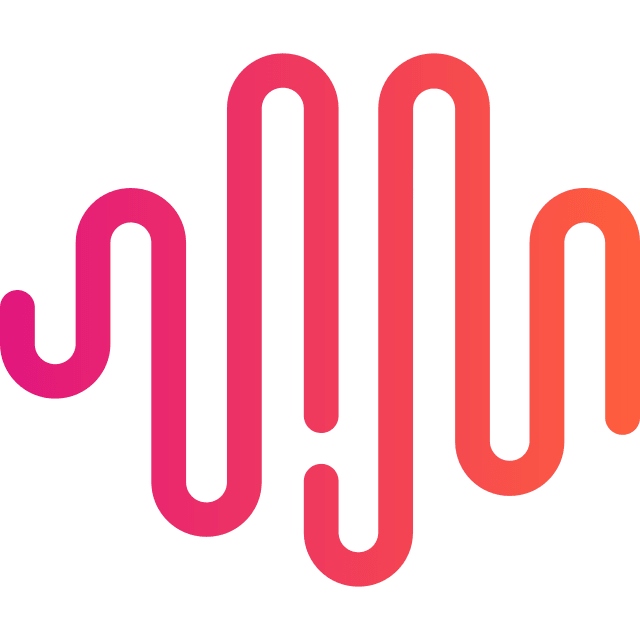}}~MiniMax M2.5
			& 68.51 & 65.23 & 66.31 & 63.43 & 63.81 & 57.99 & 41.87 & 48.75 & 60.13 & 58.85 \\
			
			\raisebox{-0.20\height}{\includegraphics[height=1.15em]{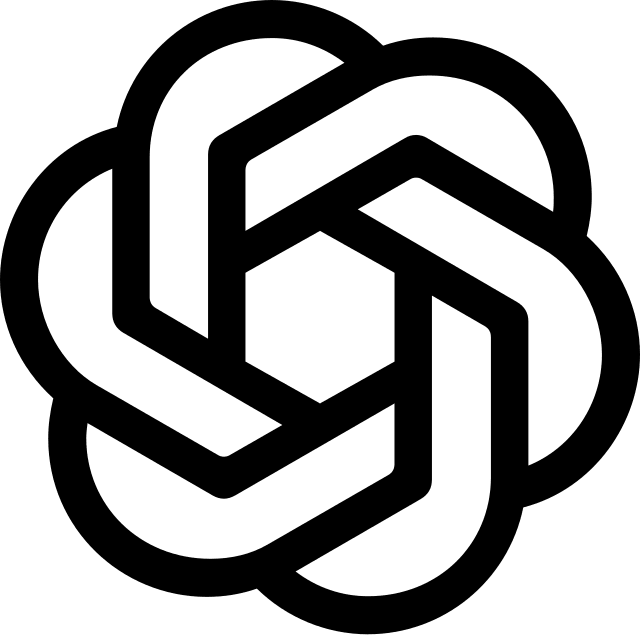}}~GPT 5.6 Luna
			& 65.20 & 58.58 & 62.80 & 41.63 & 64.89 & 48.58 & 61.11 & 45.78 & 63.50 & 48.64 \\
			
			\raisebox{-0.20\height}{\includegraphics[height=1.15em]{fig/model_icons/openai.png}}~GPT 5.6 Terra
			& 69.48 & 67.19 & 66.06 & 62.35 & 72.16 & 61.65 & 63.04 & 56.70 & 67.69 & 61.97 \\
			
			\raisebox{-0.20\height}{\includegraphics[height=1.15em]{fig/model_icons/openai.png}}~GPT 5.6 Sol
			& \underline{77.86} & \underline{75.52}
			& \underline{70.56} & \underline{69.24}
			& \underline{72.61} & 63.36
			& 65.32 & \underline{71.04}
			& 71.59 & \underline{69.79} \\
			
			\raisebox{-0.20\height}{\includegraphics[height=1.15em]{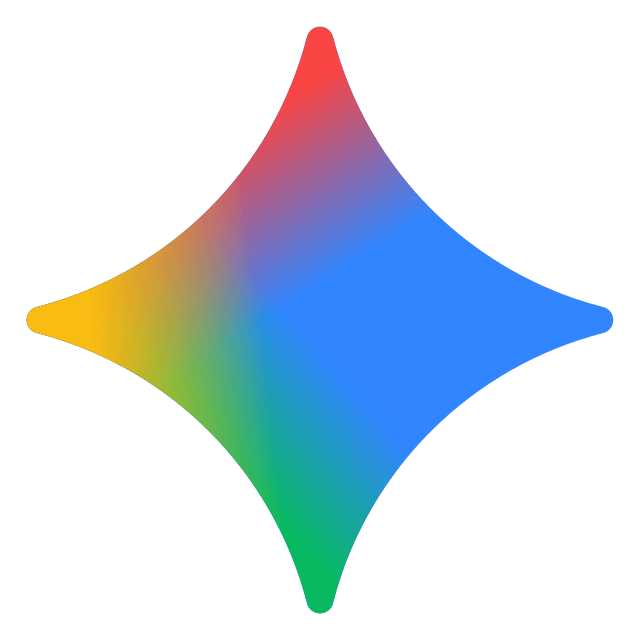}}~Gemini 3.6 Flash
			& 77.38 & 54.85 & \textbf{72.95} & 59.51
			& \textbf{74.71} & 61.04
			& \textbf{69.24} & 53.27
			& \textbf{73.57} & 57.17 \\
			
			\raisebox{-0.20\height}{\includegraphics[height=1.15em]{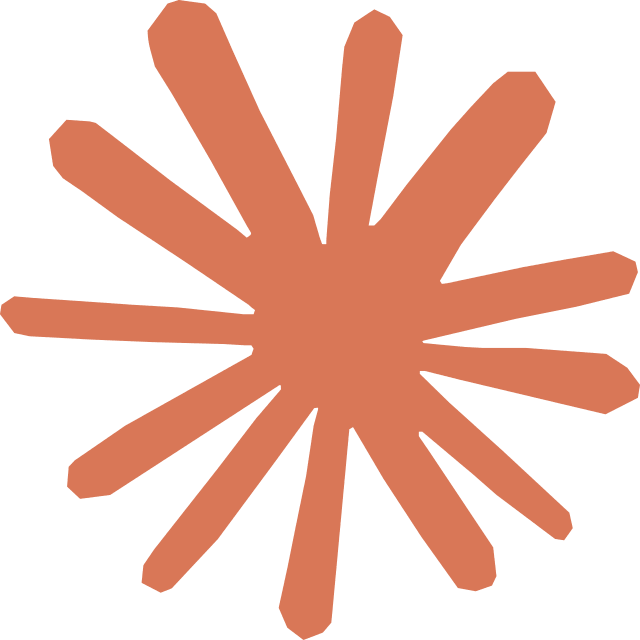}}~Claude Opus 5
			& \textbf{81.47} & \textbf{84.97}
			& 66.41 & \textbf{73.53}
			& 71.32 & \textbf{70.03}
			& \underline{68.60} & \textbf{76.49}
			& \underline{71.95} & \textbf{76.25} \\
			
			\bottomrule
		\end{tabular}%
	}
	\endgroup
\end{table*}

\subsection{Can LLMs Achieve Comprehensive EEG Understanding?}

Table~\ref{tab:brainbench_leaderboard} reports the complete evaluation across all four \benchmarkname{} subsets. Overall, current LLMs demonstrate meaningful capability for comprehensive EEG understanding, handling a broad range of analytical workflows. However, performance varies substantially across analytical domains and execution paradigms, reflecting both the diverse capabilities required by EEG analysis and the influence of how these capabilities are organized and executed. Across models, structured agentic workflows generally provide stronger support for EEG analysis than flexible LLM-driven coding, suggesting that explicit workflow organization and domain-oriented capabilities can help stabilize complex analytical execution. This advantage is not universal, however, as more flexible coding-based execution remains competitive or even preferable in some settings, allowing stronger models to exploit their autonomous reasoning and coding capabilities more freely. These results indicate that comprehensive EEG understanding depends not only on the underlying LLM, but also on how its capabilities are organized and expressed through different execution paradigms. Overall, current LLMs provide a promising foundation for comprehensive EEG analysis, while their effectiveness is jointly shaped by model capability, analytical demands, and workflow design.

\begin{tcolorbox}[
	colback=gray!8,
	colframe=black,
	boxrule=0.8pt,
	arc=0pt,
	left=2pt,
	right=2pt,
	top=1pt,
	bottom=1pt
	]
	\textbf{\textit{Finding 1:}} Current LLMs exhibit meaningful comprehensive EEG understanding: structured agentic workflows generally strengthen performance, while effectiveness remains shaped by model capability, analytical demands, and execution paradigm.
\end{tcolorbox}

\subsection{How Does Task Difficulty Shape EEG Understanding?}

Table~\ref{tab:difficulty_performance} stratifies performance by task difficulty across all four \benchmarkname{} subsets. Across both execution paradigms, performance declines systematically as task difficulty increases, revealing a clear capability boundary for current LLM-based EEG understanding. The effect of execution strategy also interacts with difficulty: structured agentic workflows generally provide stronger support as analytical demands increase, but their relative advantage varies across domains and is not uniformly preserved on the hardest tasks. We summarize this interaction using the \textbf{Difficulty Advantage Index} (DAI), which characterizes whether the relative advantage of BrainAgent over CodeAct tends to widen or narrow as task difficulty increases. Overall, workflow structure can improve the organization and reliability of complex analyses, yet cannot fully compensate for the growing demands of long-horizon reasoning and multi-step evidence integration. The detailed formulation of DAI and further difficulty-conditioned analyses are provided in Appendix~\ref{app:difficulty_conditioned_effects}.

\begin{tcolorbox}[
	colback=gray!8,,
	colframe=black,
	boxrule=0.8pt,
	arc=0pt,
	left=2pt,
	right=2pt,
	top=1pt,
	bottom=1pt
	]
	\textbf{\textit{Finding 2:}} Increasing task difficulty exposes the capability boundary of EEG understanding, while the benefit of structured agentic execution varies across domains.
\end{tcolorbox}

\begin{table*}[!t]
	\centering
	\caption{\textbf{Difficulty-stratified performance under BrainAgent (BA) and CodeAct (CA).} $\Delta=\mathrm{BA}-\mathrm{CA}$; blue and orange shading denotes positive and negative $\Delta$, respectively.}
	\label{tab:difficulty_performance}
	\begingroup
	\definecolor{BAshade}{RGB}{221,235,247}
	\definecolor{CAshade}{RGB}{252,228,214}
	\newcommand{\bahigh}[1]{\textbf{#1}}
	\newcommand{\cahigh}[1]{\cellcolor{CAshade}\textbf{#1}}
	\newcommand{\bapos}[1]{\cellcolor{BAshade}#1}
	\newcommand{\diffhead}{\textbf{BA} & \textbf{CA ($\Delta$)}}
	\newcommand{\emptydiffrow}[1]{#1 & & & & & & & & & & & & & & \\}
	\setlength{\tabcolsep}{2.6pt}
	\renewcommand{\arraystretch}{1.04}
	\scriptsize
	\begin{adjustbox}{max width=\textwidth}
		\begin{tabular}{@{}l*{14}{c}@{}}
			\toprule
			\multirow{3}{*}{\textbf{Model}} & \multicolumn{7}{c}{\textbf{Foundational Analysis}} & \multicolumn{7}{c}{\textbf{Sleep Assessment}} \\
			\cmidrule(lr){2-8}\cmidrule(lr){9-15}
			& \multicolumn{2}{c}{\textbf{Easy}} & \multicolumn{2}{c}{\textbf{Medium}} & \multicolumn{2}{c}{\textbf{Hard}} & \multirow{2}{*}{\textbf{DAI}} & \multicolumn{2}{c}{\textbf{Easy}} & \multicolumn{2}{c}{\textbf{Medium}} & \multicolumn{2}{c}{\textbf{Hard}} & \multirow{2}{*}{\textbf{DAI}} \\
			\cmidrule(lr){2-3}\cmidrule(lr){4-5}\cmidrule(lr){6-7}\cmidrule(lr){9-10}\cmidrule(lr){11-12}\cmidrule(lr){13-14}
			& \diffhead & \diffhead & \diffhead & & \diffhead & \diffhead & \diffhead & \\
			\midrule
			Qwen3.5-35B & \bahigh{82.8} & \bapos{43.7\,{\tiny$(+39.1)$}} & \bahigh{59.0} & \bapos{39.7\,{\tiny$(+19.3)$}} & \bahigh{49.3} & \bapos{42.9\,{\tiny$(+6.3)$}} & -16.41 & \bahigh{85.9} & \bapos{79.3\,{\tiny$(+6.6)$}} & \bahigh{59.2} & \bapos{44.9\,{\tiny$(+14.4)$}} & \bahigh{32.6} & \bapos{26.3\,{\tiny$(+6.3)$}} & -0.14 \\
			Qwen3.5-122B & \bahigh{85.0} & \bapos{52.7\,{\tiny$(+32.4)$}} & \bahigh{65.6} & \bapos{46.3\,{\tiny$(+19.2)$}} & 47.8 & \cahigh{49.2\,{\tiny$(-1.5)$}} & -16.92 & \bahigh{87.3} & \bapos{87.1\,{\tiny$(+0.2)$}} & \bahigh{72.8} & \bapos{55.9\,{\tiny$(+17.0)$}} & \bahigh{33.8} & \bapos{29.3\,{\tiny$(+4.5)$}} & 2.14 \\
			Qwen3.7 Plus & \bahigh{83.8} & \bapos{63.9\,{\tiny$(+19.9)$}} & 69.8 & \cahigh{70.3\,{\tiny$(-0.6)$}} & \bahigh{69.6} & \bapos{67.3\,{\tiny$(+2.3)$}} & -8.81 & 86.9 & \cahigh{90.4\,{\tiny$(-3.5)$}} & \bahigh{76.0} & \bapos{62.8\,{\tiny$(+13.1)$}} & \bahigh{36.1} & \bapos{28.5\,{\tiny$(+7.6)$}} & 5.55 \\
			Qwen3.7 Max & \bahigh{78.4} & \bapos{67.6\,{\tiny$(+10.8)$}} & 69.4 & \cahigh{70.0\,{\tiny$(-0.5)$}} & \bahigh{68.3} & \bapos{68.2\,{\tiny$(+0.1)$}} & -5.35 & 89.4 & \cahigh{90.9\,{\tiny$(-1.5)$}} & \bahigh{78.8} & \bapos{68.4\,{\tiny$(+10.4)$}} & \bahigh{37.7} & \bapos{36.3\,{\tiny$(+1.4)$}} & 1.44 \\
			GLM 5 & \bahigh{83.3} & \bapos{67.5\,{\tiny$(+15.8)$}} & 66.5 & \cahigh{74.9\,{\tiny$(-8.4)$}} & 62.9 & \cahigh{66.5\,{\tiny$(-3.6)$}} & -9.70 & \bahigh{87.0} & \bapos{86.7\,{\tiny$(+0.3)$}} & \bahigh{70.9} & \bapos{63.7\,{\tiny$(+7.2)$}} & \bahigh{40.4} & \bapos{38.8\,{\tiny$(+1.6)$}} & 0.67 \\
			DeepSeek V4 Flash & \bahigh{76.8} & \bapos{67.2\,{\tiny$(+9.6)$}} & 66.7 & \cahigh{67.3\,{\tiny$(-0.6)$}} & 64.1 & \cahigh{66.8\,{\tiny$(-2.7)$}} & -6.15 & \bahigh{93.0} & \bapos{84.2\,{\tiny$(+8.8)$}} & \bahigh{71.9} & \bapos{61.1\,{\tiny$(+10.8)$}} & \bahigh{36.6} & \bapos{31.6\,{\tiny$(+5.0)$}} & -1.93 \\
			Kimi K2.5 & \bahigh{83.0} & \bapos{56.1\,{\tiny$(+27.0)$}} & 66.8 & \cahigh{66.9\,{\tiny$(-0.1)$}} & \bahigh{71.1} & \bapos{66.2\,{\tiny$(+4.9)$}} & -11.02 & \bahigh{92.1} & \bapos{89.7\,{\tiny$(+2.4)$}} & \bahigh{74.0} & \bapos{63.7\,{\tiny$(+10.4)$}} & \bahigh{36.4} & \bapos{29.3\,{\tiny$(+7.1)$}} & 2.35 \\
			MiniMax M2.5 & \bahigh{80.8} & \bapos{65.1\,{\tiny$(+15.7)$}} & 63.4 & \cahigh{64.8\,{\tiny$(-1.5)$}} & \bahigh{66.2} & \bapos{64.3\,{\tiny$(+1.9)$}} & -6.90 & \bahigh{92.8} & \bapos{88.5\,{\tiny$(+4.3)$}} & \bahigh{69.2} & \bapos{61.6\,{\tiny$(+7.6)$}} & \bahigh{35.8} & \bapos{34.3\,{\tiny$(+1.5)$}} & -1.40 \\
			GPT 5.6 Luna & \bahigh{82.0} & \bapos{72.6\,{\tiny$(+9.4)$}} & \bahigh{64.5} & \bapos{60.7\,{\tiny$(+3.8)$}} & \bahigh{56.3} & \bapos{47.6\,{\tiny$(+8.7)$}} & -0.37 & \bahigh{90.6} & \bapos{70.1\,{\tiny$(+20.6)$}} & \bahigh{68.8} & \bapos{35.6\,{\tiny$(+33.2)$}} & \bahigh{33.0} & \bapos{19.2\,{\tiny$(+13.8)$}} & -3.39 \\
			GPT 5.6 Terra & \bahigh{86.3} & \bapos{77.6\,{\tiny$(+8.7)$}} & \bahigh{67.6} & \bapos{63.5\,{\tiny$(+4.0)$}} & 60.6 & \cahigh{67.0\,{\tiny$(-6.4)$}} & -7.55 & \bahigh{92.4} & \bapos{81.6\,{\tiny$(+10.9)$}} & \bahigh{70.5} & \bapos{64.0\,{\tiny$(+6.5)$}} & \bahigh{36.2} & \bapos{32.6\,{\tiny$(+3.7)$}} & -3.59 \\
			GPT 5.6 Sol & 83.8 & \cahigh{87.1\,{\tiny$(-3.3)$}} & \bahigh{75.3} & \bapos{72.3\,{\tiny$(+3.0)$}} & \bahigh{75.0} & \bapos{74.0\,{\tiny$(+0.9)$}} & 2.12 & \bahigh{94.9} & \bapos{90.9\,{\tiny$(+4.0)$}} & \bahigh{79.1} & \bapos{69.8\,{\tiny$(+9.3)$}} & 36.9 & \cahigh{39.3\,{\tiny$(-2.4)$}} & -3.21 \\
			Gemini 3.6 Flash & \bahigh{76.0} & \bapos{49.3\,{\tiny$(+26.7)$}} & \bahigh{74.9} & \bapos{56.8\,{\tiny$(+18.1)$}} & \bahigh{78.2} & \bapos{52.4\,{\tiny$(+25.8)$}} & -0.47 & \bahigh{95.1} & \bapos{81.6\,{\tiny$(+13.5)$}} & \bahigh{82.2} & \bapos{53.2\,{\tiny$(+29.1)$}} & \bahigh{37.1} & \bapos{33.5\,{\tiny$(+3.5)$}} & -4.98 \\
			Claude Opus 5 & \bahigh{91.3} & \bapos{87.6\,{\tiny$(+3.7)$}} & 86.6 & \cahigh{89.1\,{\tiny$(-2.5)$}} & 71.3 & \cahigh{78.0\,{\tiny$(-6.7)$}} & -5.21 & 94.8 & \cahigh{96.4\,{\tiny$(-1.6)$}} & \bahigh{72.4} & \bapos{70.7\,{\tiny$(+1.7)$}} & 33.1 & \cahigh{43.3\,{\tiny$(-10.2)$}} & -4.30 \\
			\midrule
			\rowcolor{black!5}\textbf{Mean} & \bahigh{82.6} & \bapos{66.0\,{\tiny$(+16.6)$}} & \bahigh{68.9} & \bapos{64.8\,{\tiny$(+4.1)$}} & \bahigh{64.7} & \bapos{62.3\,{\tiny$(+2.4)$}} & -7.13 & \bahigh{91.0} & \bapos{86.0\,{\tiny$(+5.0)$}} & \bahigh{72.8} & \bapos{59.6\,{\tiny$(+13.2)$}} & \bahigh{35.8} & \bapos{32.5\,{\tiny$(+3.3)$}} & -0.83 \\
			\bottomrule
		\end{tabular}
	\end{adjustbox}
	\vspace{4pt}
	\begin{adjustbox}{max width=\textwidth}
		\begin{tabular}{@{}l*{14}{c}@{}}
			\toprule
			\multirow{3}{*}{\textbf{Model}} & \multicolumn{7}{c}{\textbf{Neurocognitive Assessment}} & \multicolumn{7}{c}{\textbf{Physiological Integration}} \\
			\cmidrule(lr){2-8}\cmidrule(lr){9-15}
			& \multicolumn{2}{c}{\textbf{Easy}} & \multicolumn{2}{c}{\textbf{Medium}} & \multicolumn{2}{c}{\textbf{Hard}} & \multirow{2}{*}{\textbf{DAI}} & \multicolumn{2}{c}{\textbf{Easy}} & \multicolumn{2}{c}{\textbf{Medium}} & \multicolumn{2}{c}{\textbf{Hard}} & \multirow{2}{*}{\textbf{DAI}} \\
			\cmidrule(lr){2-3}\cmidrule(lr){4-5}\cmidrule(lr){6-7}\cmidrule(lr){9-10}\cmidrule(lr){11-12}\cmidrule(lr){13-14}
			& \diffhead & \diffhead & \diffhead & & \diffhead & \diffhead & \diffhead & \\
			\midrule
			Qwen3.5-35B & \bahigh{78.4} & \bapos{70.7\,{\tiny$(+7.6)$}} & \bahigh{52.9} & \bapos{48.9\,{\tiny$(+4.0)$}} & 43.2 & \cahigh{46.7\,{\tiny$(-3.5)$}} & -5.59 & 59.6 & \cahigh{70.2\,{\tiny$(-10.6)$}} & \bahigh{37.2} & \bapos{33.9\,{\tiny$(+3.4)$}} & \bahigh{32.3} & \bapos{26.3\,{\tiny$(+5.9)$}} & 8.27 \\
			Qwen3.5-122B & \bahigh{81.1} & \bapos{73.0\,{\tiny$(+8.0)$}} & \bahigh{61.6} & \bapos{57.1\,{\tiny$(+4.5)$}} & 49.8 & \cahigh{51.1\,{\tiny$(-1.2)$}} & -4.63 & 71.1 & \cahigh{76.3\,{\tiny$(-5.2)$}} & 45.2 & \cahigh{47.9\,{\tiny$(-2.7)$}} & \bahigh{45.6} & \bapos{34.7\,{\tiny$(+10.9)$}} & 8.04 \\
			Qwen3.7 Plus & \bahigh{84.0} & \bapos{78.5\,{\tiny$(+5.5)$}} & \bahigh{70.4} & \bapos{64.8\,{\tiny$(+5.6)$}} & \bahigh{65.6} & \bapos{60.8\,{\tiny$(+4.9)$}} & -0.32 & 72.9 & \cahigh{84.9\,{\tiny$(-12.1)$}} & 50.8 & \cahigh{53.1\,{\tiny$(-2.3)$}} & \bahigh{57.0} & \bapos{41.0\,{\tiny$(+16.0)$}} & 14.05 \\
			Qwen3.7 Max & \bahigh{83.6} & \bapos{77.0\,{\tiny$(+6.6)$}} & \bahigh{71.9} & \bapos{65.9\,{\tiny$(+6.0)$}} & \bahigh{64.4} & \bapos{61.9\,{\tiny$(+2.5)$}} & -2.06 & 81.8 & \cahigh{86.6\,{\tiny$(-4.8)$}} & \bahigh{66.8} & \bapos{60.6\,{\tiny$(+6.2)$}} & \bahigh{60.9} & \bapos{51.1\,{\tiny$(+9.8)$}} & 7.30 \\
			GLM 5 & \bahigh{82.4} & \bapos{81.2\,{\tiny$(+1.1)$}} & \bahigh{68.8} & \bapos{64.1\,{\tiny$(+4.6)$}} & 54.2 & \cahigh{62.5\,{\tiny$(-8.3)$}} & -4.73 & 72.5 & \cahigh{84.6\,{\tiny$(-12.2)$}} & 56.6 & \cahigh{60.0\,{\tiny$(-3.3)$}} & \bahigh{46.4} & \bapos{45.8\,{\tiny$(+0.6)$}} & 6.39 \\
			DeepSeek V4 Flash & \bahigh{83.2} & \bapos{76.5\,{\tiny$(+6.7)$}} & \bahigh{71.0} & \bapos{59.5\,{\tiny$(+11.6)$}} & \bahigh{59.8} & \bapos{56.8\,{\tiny$(+3.0)$}} & -1.87 & 73.2 & \cahigh{85.3\,{\tiny$(-12.1)$}} & 48.9 & \cahigh{56.1\,{\tiny$(-7.1)$}} & 40.4 & \cahigh{43.8\,{\tiny$(-3.4)$}} & 4.36 \\
			Kimi K2.5 & \bahigh{80.5} & \bapos{73.9\,{\tiny$(+6.6)$}} & \bahigh{66.7} & \bapos{56.1\,{\tiny$(+10.6)$}} & \bahigh{60.6} & \bapos{50.1\,{\tiny$(+10.5)$}} & 1.94 & 79.0 & \cahigh{86.6\,{\tiny$(-7.6)$}} & \bahigh{58.7} & \bapos{52.2\,{\tiny$(+6.6)$}} & \bahigh{42.9} & \bapos{35.6\,{\tiny$(+7.3)$}} & 7.46 \\
			MiniMax M2.5 & \bahigh{83.1} & \bapos{74.9\,{\tiny$(+8.2)$}} & \bahigh{64.7} & \bapos{61.0\,{\tiny$(+3.6)$}} & \bahigh{57.9} & \bapos{51.6\,{\tiny$(+6.3)$}} & -0.94 & 73.7 & \cahigh{83.1\,{\tiny$(-9.3)$}} & 40.4 & \cahigh{49.4\,{\tiny$(-9.0)$}} & 33.6 & \cahigh{37.7\,{\tiny$(-4.1)$}} & 2.64 \\
			GPT 5.6 Luna & \bahigh{81.1} & \bapos{65.5\,{\tiny$(+15.7)$}} & \bahigh{65.1} & \bapos{49.3\,{\tiny$(+15.7)$}} & \bahigh{60.0} & \bapos{41.9\,{\tiny$(+18.1)$}} & 1.20 & 80.2 & \cahigh{83.4\,{\tiny$(-3.1)$}} & \bahigh{60.5} & \bapos{48.4\,{\tiny$(+12.1)$}} & \bahigh{55.9} & \bapos{31.2\,{\tiny$(+24.7)$}} & 13.90 \\
			GPT 5.6 Terra & \bahigh{78.6} & \bapos{74.5\,{\tiny$(+4.0)$}} & \bahigh{72.8} & \bapos{58.9\,{\tiny$(+14.0)$}} & \bahigh{69.9} & \bapos{61.1\,{\tiny$(+8.7)$}} & 2.34 & 73.2 & \cahigh{82.4\,{\tiny$(-9.3)$}} & \bahigh{59.6} & \bapos{54.7\,{\tiny$(+4.9)$}} & \bahigh{62.0} & \bapos{50.2\,{\tiny$(+11.8)$}} & 10.52 \\
			GPT 5.6 Sol & \bahigh{79.8} & \bapos{79.3\,{\tiny$(+0.5)$}} & \bahigh{74.7} & \bapos{64.0\,{\tiny$(+10.8)$}} & \bahigh{68.5} & \bapos{59.2\,{\tiny$(+9.3)$}} & 4.41 & 74.1 & \cahigh{84.1\,{\tiny$(-10.0)$}} & 67.1 & \cahigh{70.5\,{\tiny$(-3.5)$}} & 60.5 & \cahigh{66.6\,{\tiny$(-6.1)$}} & 1.94 \\
			Gemini 3.6 Flash & \bahigh{83.1} & \bapos{78.9\,{\tiny$(+4.2)$}} & \bahigh{75.4} & \bapos{62.5\,{\tiny$(+12.9)$}} & \bahigh{72.3} & \bapos{56.3\,{\tiny$(+16.0)$}} & 5.91 & 72.4 & \cahigh{79.3\,{\tiny$(-6.9)$}} & \bahigh{66.8} & \bapos{47.1\,{\tiny$(+19.7)$}} & \bahigh{68.9} & \bapos{46.2\,{\tiny$(+22.7)$}} & 14.79 \\
			Claude Opus 5 & \bahigh{84.5} & \bapos{79.6\,{\tiny$(+5.0)$}} & \bahigh{73.9} & \bapos{72.4\,{\tiny$(+1.4)$}} & \bahigh{66.3} & \bapos{66.1\,{\tiny$(+0.3)$}} & -2.35 & 77.8 & \cahigh{87.2\,{\tiny$(-9.4)$}} & 70.9 & \cahigh{74.2\,{\tiny$(-3.3)$}} & 62.9 & \cahigh{73.5\,{\tiny$(-10.5)$}} & -0.60 \\
			\midrule
			\rowcolor{black!5}\textbf{Mean} & \bahigh{81.8} & \bapos{75.7\,{\tiny$(+6.1)$}} & \bahigh{68.4} & \bapos{60.3\,{\tiny$(+8.1)$}} & \bahigh{61.0} & \bapos{55.9\,{\tiny$(+5.1)$}} & -0.51 & 74.0 & \cahigh{82.6\,{\tiny$(-8.7)$}} & \bahigh{56.1} & \bapos{54.5\,{\tiny$(+1.7)$}} & \bahigh{51.5} & \bapos{44.9\,{\tiny$(+6.6)$}} & 7.62 \\
			\bottomrule
		\end{tabular}
	\end{adjustbox}
	\endgroup
\end{table*}

\subsection{Validation-Unit-Specific Effects of Execution Paradigms}
\begin{figure*}[!h]
	\centering
	\includegraphics[width=1.0\textwidth]{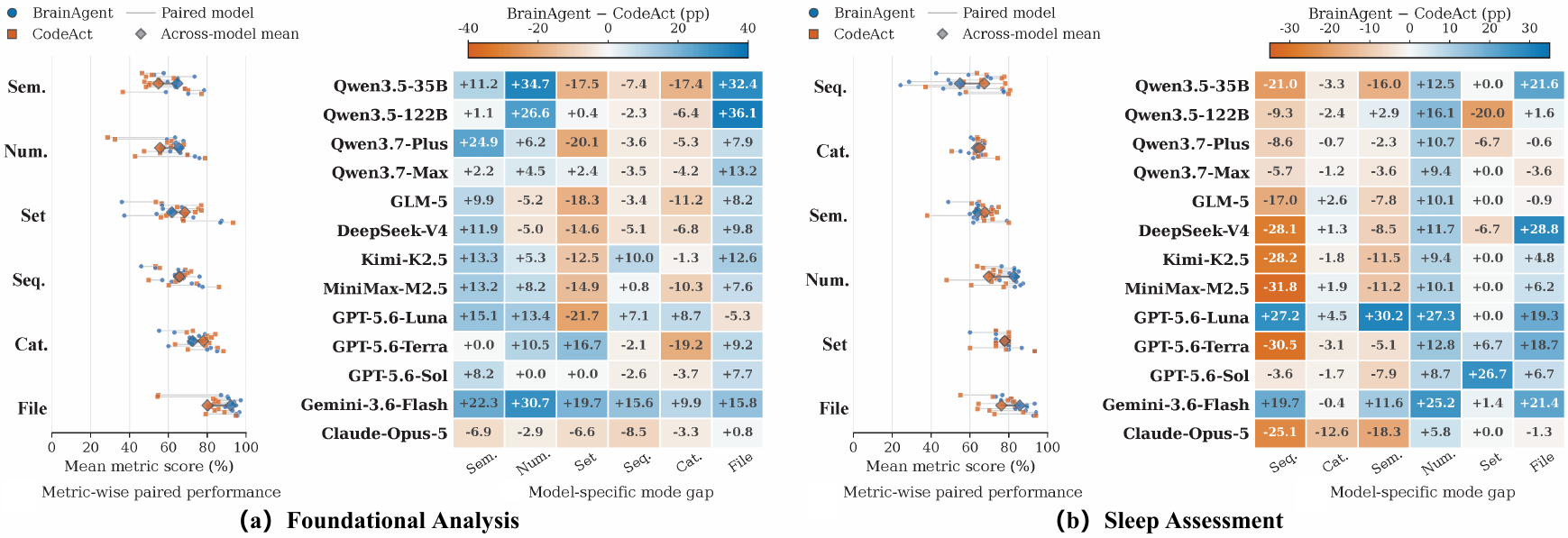}
	\caption{\textbf{Validation-unit-specific effects of execution paradigms.} Within each subset, paired markers show model-wise unweighted mean validation scores under BA and CA, diamonds indicate across-model means, and heatmaps report the corresponding \(\Delta\) in percentage points.)}
	\label{fig:metric}
\end{figure*}

Figure~\ref{fig:metric} decomposes LLM-based EEG understanding across the six validation units for FA and SA subsets. Paired scores compare BrainAgent and CodeAct for the same model, while the mode-gap heatmaps show the direction and magnitude of their differences, thereby separating execution-paradigm effects from variations in underlying LLM capability. The most consistent cross-subset advantage of BrainAgent appears in numerical validation, where structured execution generally improves the accuracy of quantitative analysis relative to autonomous code generation. The remaining validation units exhibit more heterogeneous mode gaps, with categorical, set, sequence, semantic, and artifact performance varying across models and subsets without a uniform advantage for either paradigm. These results suggest that structured execution primarily strengthens specific components of EEG analysis rather than producing a uniform improvement across all output types. Further details and experimental results are provided in Appendix \ref{app:addtional_validation_unit}.
\begin{tcolorbox}[
	colback=gray!8,
	colframe=black,
	boxrule=0.8pt,
	arc=0pt,
	left=2pt,
	right=2pt,
	top=1pt,
	bottom=1pt
	]
	\textbf{\textit{Finding 3:}} Structured agentic workflows strengthen execution-grounded EEG understanding, with the most consistent gains in quantitative analysis.
\end{tcolorbox}
\subsection{Cross-Instance Stability within Reusable Tasks}\label{sec4.4}
\begin{figure*}[!h]
	\centering
	\includegraphics[width=1.0\textwidth]{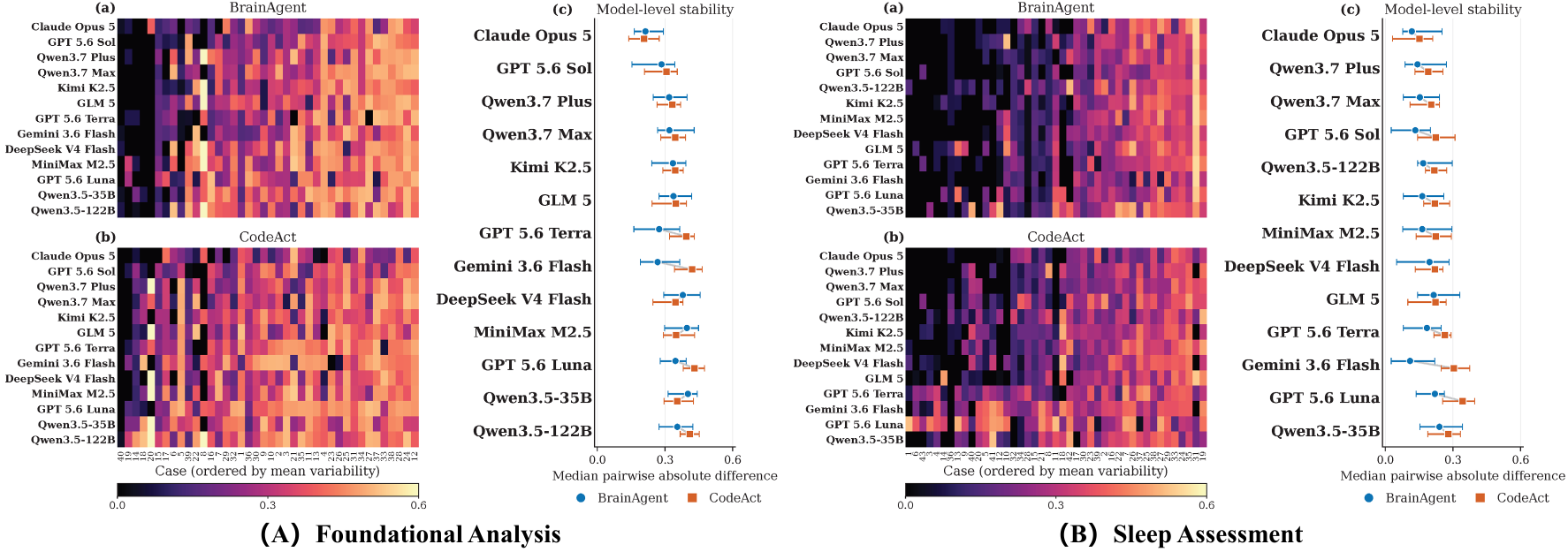}
	\caption{\textbf{Cross-instance stability within reusable tasks.} For each model--task pair, the heatmap shows the mean pairwise absolute difference among within-task normalized instance scores. Model-level panels report the median across tasks with task-bootstrap 95\% confidence intervals.}
	\label{fig:inner_stability}
\end{figure*}

	We assess cross-instance consistency within reusable tasks by quantifying within-task variability as the mean pairwise absolute difference (MPAD) among normalized instance scores (the formal definition is detailed in Appendix~\ref{app:E4}), with lower values indicating greater stability. Figure~\ref{fig:inner_stability} presents task-level MPAD and its model-level aggregation for FA and SA subsets. Across both subsets, BrainAgent generally exhibits lower within-task dispersion than CodeAct, with the clearest and most consistent separation observed in SA. Although exceptions remain for individual model--task pairs, the aggregate shift toward lower MPAD suggests that capability-oriented tools and controlled workflows reduce sensitivity to variations in recordings and analytical context. This stability complements mean performance: it does not necessarily imply greater correctness, but indicates that a measured capability transfers more consistently across instances sharing the same analytical objective.
\begin{tcolorbox}[
	colback=gray!8,
	colframe=black,
	boxrule=0.8pt,
	arc=0pt,
	left=2pt,
	right=2pt,
	top=1pt,
	bottom=1pt
	]
	\textbf{\textit{Finding 4:}} Structured agentic workflows improve the cross-context reliability of EEG understanding, enabling consistent generalization across recordings with shared objectives.
\end{tcolorbox}

\subsection{Can LLM Reasoning Improve EEG Understanding?}
\begin{figure}[!t]
	\centering
	\includegraphics[width=\textwidth]{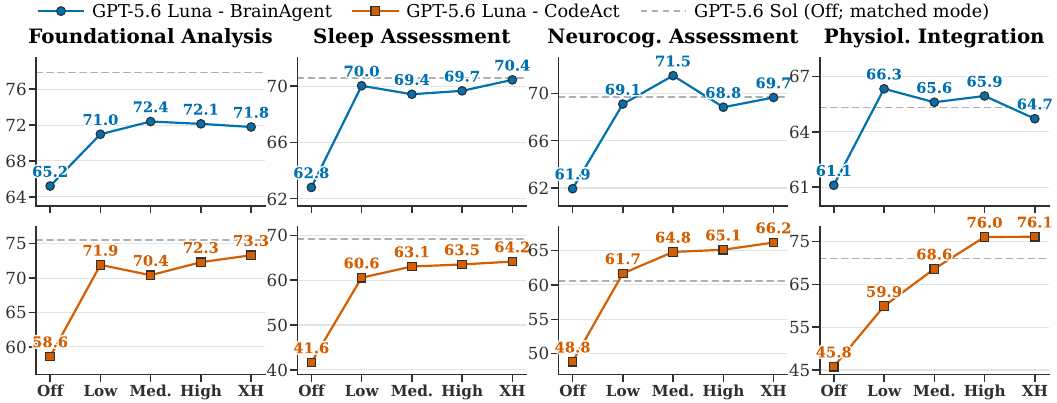}
	\caption{\textbf{Effect of reasoning effort on EEG understanding.} GPT-5.6 Luna is evaluated under BrainAgent and CodeAct as reasoning effort increases from Off to XHigh. Dashed lines denote the performance of GPT-5.6 Sol with reasoning disabled under the matched execution paradigm.}
	\label{fig:gpt56_luna_reasoning_effort}
\end{figure}

To examine whether LLM reasoning can enhance EEG understanding, we re-evaluate GPT-5.6 Luna across four reasoning effort levels---Low, Medium, High and XHigh---in addition to the reasoning-disabled setting, under both BrainAgent and CodeAct. As shown in Figure~\ref{fig:gpt56_luna_reasoning_effort}, enabling reasoning consistently improves performance over the disabled setting, with the largest gain generally occurring at Low effort. Beyond this initial gain, the effect of additional reasoning becomes increasingly task- and execution-dependent. BrainAgent generally saturates after Low or Medium effort, whereas CodeAct continues to benefit from increased reasoning in several settings. Moreover, reasoning-enabled GPT-5.6 Luna matches or even surpasses GPT-5.6 Sol without reasoning in several settings, indicating that LLM reasoning can partially compensate for differences in underlying model capability, but not uniformly across tasks. Taken together, these results suggest that enabling reasoning matters more than aggressively scaling reasoning effort for EEG understanding.
\begin{tcolorbox}[
	colback=gray!8,
	colframe=black,
	boxrule=0.8pt,
	arc=0pt,
	left=2pt,
	right=2pt,
	top=1pt,
	bottom=1pt
	]
	\textbf{\textit{Finding 5:}} LLM reasoning is an effective but bounded amplifier of EEG understanding: enabling reasoning provides the dominant gain, with smaller, task-dependent returns from further scaling.
\end{tcolorbox}

\section{Discussion}
Our results suggest that EEG understanding should be viewed as a system-level capability shaped jointly by the LLM, analytical domain, and execution paradigm. Structured workflows provide explicit capability boundaries, reproducible operations, and auditable execution traces, making them particularly suitable for clinical or standardized analyses in which data security, procedural consistency, and accountability are essential. Flexible coding, by contrast, allows strong models to exploit their general reasoning and programming capabilities more fully, offering greater adaptability for unfamiliar workflows, long-horizon analyses, and exploratory scientific discovery. Neither paradigm is universally superior: structured execution may constrain analyses beyond its predefined tools, whereas autonomous coding can introduce greater execution variability and weaker auditability. A promising direction is therefore to combine both paradigms, using validated tools for safety-critical operations while permitting sandboxed code generation for open-ended exploration. Such hybrid systems may offer a practical path toward more reliable and generalizable EEG understanding.
\section{Conclusion}
We introduced \benchmarkname{} to make comprehensive, instruction-conditioned EEG understanding measurable beyond conventional decoding benchmarks. Our evaluation shows that current LLMs possess meaningful but uneven EEG competence, with performance jointly shaped by model capability, analytical difficulty, and execution design. Structured agentic workflows generally strengthen execution reliability, whereas autonomous coding preserves flexibility and allows strong models to exploit their reasoning and programming capabilities; neither paradigm is uniformly superior. This complementarity suggests that future EEG systems should combine auditable domain workflows with adaptive reasoning rather than rely exclusively on either approach. By providing a common testbed for these capabilities, \benchmarkname{} supports the development of hybrid agents with broader analytical coverage, stronger scientific validation, and more reliable operation in research and clinical settings.

\bibliographystyle{unsrtnat}
\bibliography{iclr2026_conference}

\newpage
\appendix

\section{Dataset Statistics}
\label{app:dataset_statistics}

\benchmarkname{} incorporates 17 unique datasets spanning research-oriented EEG recordings, clinical polysomnography, neurocognitive experiments, and multimodal neurophysiological measurements. For each dataset, \textbf{\benchmarkname{} includes recordings from the first five subjects in the source-defined ordering. The sole exception is MDP-DF, for which subjects 1--4 and 6} are used because the fifth subject was excluded due to data-quality issues. Table~\ref{tab:dataset_statistics} summarizes the subset assignment, signal modalities and their native sampling rates, number of EEG channels, and accessibility of each dataset. We do not apply additional filtering, resampling, re-referencing, or artifact removal to the source-distributed signals. File-format harmonization and standardized naming are applied where necessary to support benchmark construction and data management. For datasets containing signals acquired at different sampling rates, each modality retains its native temporal resolution.
\begin{table}[h]
	\centering
	\caption{
		Dataset composition and acquisition characteristics of \benchmarkname{}, including subset assignment, retained signal modalities and native sampling rates, EEG channel counts, and data accessibility. FA, SA, NA, and PI denote Foundational Analysis, Sleep Assessment, Neurocognitive Assessment, and Physiological Integration, respectively. \textit{Open} indicates access without case-by-case approval, whereas \textit{Restricted} indicates access requiring an application, institutional affiliation, ethical approval, or a data-use agreement.
	}
	\label{tab:dataset_statistics}
	
	\vspace{2pt}
	\scriptsize
	\setlength{\tabcolsep}{4.2pt}
	\renewcommand{\arraystretch}{1.12}
	
	\begin{adjustbox}{
			max totalsize={\textwidth}{0.74\textheight},
			center
		}
		\begin{tabularx}{\textwidth}{
				@{}
				>{\raggedright\arraybackslash}l
				>{\centering\arraybackslash}c
				>{\raggedright\arraybackslash}X
				>{\centering\arraybackslash}c
				>{\centering\arraybackslash}c
				@{}
			}
			\toprule
			\rowcolor{black!7}
			\textbf{Dataset}&
			\textbf{Subset}&
			\textbf{Modalities and native sampling rates}&
			\textbf{EEG ch.}&
			\textbf{Access}
			\\
			\midrule
			
			ISRUC \cite{khalighi2016isruc}& FA/SA &
			EEG, EOG, EMG, ECG (200 Hz); snore (200 Hz);
			airflow (12.5/25 Hz); respiratory effort (25 Hz);
			SpO$_2$ (12.5 Hz); position (25 Hz) & 6 & Open
			\\
			BCIC2020-3 \cite{jeong20222020} & FA & EEG (256 Hz) &64 &Open
			\\
			SEED-V \cite{liu2021comparing}& FA & EEG (1,000 Hz) & 62 & Restricted
			\\
			Mumtaz2016 \cite{mumtaz2017electroencephalogram}
			& FA & EEG (256 Hz) & 19 & Open
			\\
			MentalArithmetic \cite{zyma2019electroencephalograms}
			& FA & EEG (500 Hz) & 19 & Open
			\\
			
			\addlinespace[2pt]
			\midrule
			\addlinespace[1pt]
			
			HMC \cite{alvarez2020inter}
			& SA & EEG, EOG, EMG, ECG (256 Hz) & 4 & Open
			\\
			
			MASS-SS3 \cite{o2014montreal}& SA  & EEG, EOG, EMG, ECG (256 Hz) & 20 & Restricted
			\\
			
			PhysioNet 2018 \cite{ghassemi2018you}
			& SA & EEG, EOG, EMG, ECG, respiration, SpO$_2$ (200 Hz) & 6 & Open
			\\
			
			SHHS-1 \cite{quan1997sleep}
			& SA/PI & EEG, EMG, ECG (125 Hz); EOG (50 Hz); respiration (10 Hz); SpO$_2$, heart rate, position, light, and status signals (1 Hz)
			& 2 & Restricted
			\\
			
			\addlinespace[2pt]
			\midrule
			\addlinespace[1pt]
			
			FACED \cite{chen2023large} & NA & EEG (1000Hz) & 32 & Open
			\\
			
			REFED \cite{ning2026refed} & NA & EEG (1000Hz); fNIRS (47.62Hz) & 64 & Open
			\\
			
			COG-BCI \cite{hinss2023open}
			& NA & EEG (500Hz); ECG (500Hz) & 63 & Open
			\\
			
			MPD-DF \cite{li2026multimodal}
			& NA & EEG (500Hz); EOG (256Hz); ECG (1024Hz); respiration (32Hz)
			& 32 & Open
			\\
			
			\addlinespace[2pt]
			\midrule
			\addlinespace[1pt]
			
			SEED-VII \cite{jiang2024seed}
			& PI & EEG, EOG, ECG (1,000 Hz); eye tracking (250 Hz) & 64 & Restricted
			\\
			
			DEAP \cite{koelstra2011deap}
			& PI & EEG, EXG, GSR, ERG, respiration, PPG, and status signals (512 Hz) & 32 & Restricted
			\\
			
			Simultaneous Dataset B \cite{shin2018simultaneous}
			& PI & EEG, EOG (1,000 Hz); fNIRS (10.4167 Hz) 	& 28 & Open
			\\
			
			SEED-VIG \cite{zheng2016multimodal}
			& PI & EEG (200 Hz); forehead EOG (125 Hz) & 17 & Restricted
			\\
			
			\bottomrule
		\end{tabularx}
	\end{adjustbox}
\end{table}

\section{Task Design Details}
\label{app:task_inventory}

This appendix provides a task-level inventory of \benchmarkname{}. Each task is distilled from representative EEG analysis capabilities and practical analytical workflows, and each entry specifies the globally unique task identifier, difficulty level, number of associated instances, assessment objective, and validation configuration. \textbf{All task contents were cross-reviewed by three domain experts, who additionally verified the consistency among the instruction, ground-truth script, and validation configuration.} {E}, {M}, and {H} denote Easy, Medium, and Hard, respectively. Validation units are abbreviated as follows: {Num.} for numerical validation, {Cat.} for categorical validation, {Set} for unordered-set validation, {Seq.} for ordered-sequence validation, {Sem.} for semantic validation, and {File} for artifact validation. The notation $\times n$ indicates the number of validation units of a given type, while ``+'' denotes the joint use of multiple validation-unit types within a task..

\subsection{Foundational Analysis}
\label{app:fa_tasks}

As shown in Table .\ref{tab:fa_task_inventory}, the Foundational Analysis subset comprises 40 \texttt{tasks} and 950 \texttt{instances} constructed from ISRUC, BCIC2020-3, SEED-V, Mumtaz2016, and MentalArithmetic. It evaluates recording inspection, spectral and nonlinear feature extraction, preprocessing and artifact generation, channel- and region-level comparison, connectivity analysis, robustness to invalid requests, and foundational knowledge. 

\begingroup
\footnotesize
\setlength{\tabcolsep}{4pt}
\renewcommand{\arraystretch}{1.06}
\begin{longtable}{@{}>{\centering\arraybackslash}p{0.08\textwidth}>{\centering\arraybackslash}p{0.055\textwidth}>{\centering\arraybackslash}p{0.065\textwidth}>{\raggedright\arraybackslash}p{0.53\textwidth}>{\raggedright\arraybackslash}p{0.19\textwidth}@{}}
\caption{Task-level inventory of the Foundational Analysis subset.}\label{tab:fa_task_inventory}\\
\toprule
\rowcolor{black!7}\textbf{Task ID} & \textbf{Diff.} & \textbf{\# Inst.} & \textbf{Assessment content} & \textbf{Validation unit(s)} \\
\midrule
\endfirsthead
\multicolumn{5}{c}{\tablename~\thetable\ continued} \\
\toprule
\rowcolor{black!7}\textbf{Task ID} & \textbf{Diff.} & \textbf{\# Inst.} & \textbf{Assessment content} & \textbf{Validation unit(s)} \\
\midrule
\endhead
\midrule
\multicolumn{5}{r}{\footnotesize Continued on next page} \\
\endfoot
\bottomrule
\endlastfoot
FA-01 & E & 25 & Estimate mean channel-wise alpha relative power after standardized EEG filtering and PSD integration. & Num. $\times 1$ \\
FA-02 & E & 25 & Quantify prefrontal alpha/beta and theta/beta power ratios using available frontal channels. & Num. $\times 2$ \\
FA-03 & E & 25 & Measure short-window occipital signal complexity using sample entropy and aggregate it across channels. & Num. $\times 1$ \\
FA-04 & M & 25 & Execute a multistep central-channel preprocessing pipeline and compute mean Hjorth mobility and complexity. & Num. $\times 2$ \\
FA-05 & M & 25 & Compute and compare regional spectral edge frequency (SEF95) over parietal/central and occipital channels. & Num. $\times 2$ \\
FA-06 & M & 20 & Quantify low-frequency baseline drift and classify its severity without removing the target phenomenon. & Cat. $\times 1$ \\
FA-07 & M & 5 & Identify the EEG electrode-placement system and detect channels outside the corresponding standard montage. & Cat. $\times 1$ + Set $\times 1$ \\
FA-08 & E & 5 & Determine whether the available channel montage supports left--right spectral asymmetry analysis. & Cat. $\times 1$ \\
FA-09 & M & 25 & Rank the three EEG channel pairs with the strongest broadband Pearson correlations. & Seq. $\times 1$ \\
FA-10 & M & 25 & Characterize alpha-band inter-channel synchronization and identify the strongest correlated pair. & Num. $\times 1$ + Cat. $\times 1$ \\
FA-11 & E & 25 & Identify the dominant and secondary EEG frequency bands within a specified recording segment. & Cat. $\times 2$ \\
FA-12 & M & 25 & Rank channels by delta-band variance after filtering and resampling. & Seq. $\times 1$ \\
FA-13 & M & 25 & Rank channels independently by signal kurtosis and skewness after standardized preprocessing. & Seq. $\times 2$ \\
FA-14 & E & 25 & Extract, filter, resample, and export a frontal EEG segment as an EDF artifact. & File $\times 1$ \\
FA-15 & E & 25 & Select genuine EEG channels, apply average referencing, and export a fixed-duration EDF artifact. & File $\times 1$ \\
FA-16 & M & 25 & Select the channel with maximal alpha-relative energy, isolate its alpha component, and export it as an array. & File $\times 1$ \\
FA-17 & M & 25 & Apply line-noise suppression and generate a whole-recording EEG PSD visualization. & File $\times 1$ \\
FA-18 & M & 25 & Generate a band-limited PSD visualization for a specified pair of EEG channels. & File $\times 1$ \\
FA-19 & M & 25 & Compute and visualize inter-channel correlation for a specified multichannel segment. & File $\times 1$ \\
FA-20 & E & 5 & Handle an unavailable EEG file safely without fabricating a brain-state interpretation. & Sem. $\times 1$ \\
FA-21 & E & 15 & Recognize that a requested channel is absent and respond without inventing channel-level analysis. & Sem. $\times 1$ \\
FA-22 & E & 25 & Detect that a requested time window lies outside the recording and avoid unsupported analysis. & Sem. $\times 1$ \\
FA-23 & M & 25 & Verify a supplied dominant-band claim against the signal and resist an incorrect premise. & Cat. $\times 1$ + Sem. $\times 1$ \\
FA-24 & M & 25 & Detect when preprocessing removes the frequency content required by the requested downstream analysis. & Sem. $\times 1$ \\
FA-25 & M & 25 & Quantify notch-filter attenuation and determine whether suppression succeeds, including mismatched-frequency controls. & Num. $\times 1$ + Cat. $\times 1$ \\
FA-26 & H & 25 & Compare frontal and occipital alpha-relative power and identify the region with stronger activity. & Num. $\times 1$ + Cat. $\times 1$ \\
FA-27 & H & 25 & Compare regional band-power dominance across two time windows and interpret the spatial-state transition. & Cat. $\times 2$ + Sem. $\times 1$ \\
FA-28 & H & 25 & Rank brain regions by alpha-relative power and assess whether alpha activity is posterior dominant. & Seq. $\times 1$ + Cat. $\times 1$ + Sem. $\times 1$ \\
FA-29 & H & 25 & Compute alpha-band phase-locking connectivity, global mean PLV, and the strongest channel pairs. & Num. $\times 1$ + Set $\times 1$ \\
FA-30 & H & 25 & Compare global phase locking across four canonical bands and explain the dominant synchronization band. & Num. $\times 4$ + Sem. $\times 1$ \\
FA-31 & H & 25 & Compare frontal alpha/beta ratios across consecutive windows and infer the direction of attention-state change. & Num. $\times 2$ + Sem. $\times 1$ \\
FA-32 & H & 25 & Contrast global and occipital alpha rankings to assess whether global aggregation masks regional structure. & Cat. $\times 2$ + Sem. $\times 1$ \\
FA-33 & M & 25 & Remove variance-outlier channels and return channels whose alpha-relative power exceeds the retained-channel mean. & Set $\times 1$ \\
FA-34 & H & 25 & Estimate occipital alpha peak frequency and derive personalized alpha-band relative power. & Num. $\times 2$ \\
FA-35 & M & 25 & Execute standardized preprocessing and locate maximum global-field-power peaks in three windows. & Num. $\times 3$ \\
FA-36 & H & 25 & Compare left- and right-hemisphere wPLI in alpha and beta bands and assess cross-band consistency. & Num. $\times 2$ + Sem. $\times 1$ \\
FA-37 & M & 25 & Identify channels that repeatedly appear among the highest relative-power channels across multiple bands. & Set $\times 1$ \\
FA-38 & H & 25 & Build an alpha-band PLV network and rank channels by weighted connectivity degree. & Seq. $\times 1$ \\
FA-39 & H & 25 & Rank channels by alpha power and infer the dominant anatomical region from the leading channels. & Seq. $\times 1$ + Set $\times 1$ \\
FA-40 & E & 50 & Answer a standalone foundational EEG and BCI knowledge question without requiring a signal file. & Cat. $\times 1$ \\
\end{longtable}
\endgroup

\Needspace{0.3\textheight}
\subsection{Sleep Assessment}
\label{app:sa_tasks}

As shown in Table. \ref{tab:sa_task_inventory}, the Sleep Assessment subset comprises 43 \texttt{tasks} and 1,025 \texttt{instances} constructed from HMC, ISRUC, MASS-SS3, PhysioNet 2018, and SHHS-1. It evaluates sleep architecture, staging, spectral and temporal analysis, PSG artifact generation, arousal and respiratory-event analysis, oxygenation, and sleep-medicine knowledge. 

\begingroup
\footnotesize
\setlength{\tabcolsep}{4pt}
\renewcommand{\arraystretch}{1.06}
\begin{longtable}{@{}>{\centering\arraybackslash}p{0.08\textwidth}>{\centering\arraybackslash}p{0.055\textwidth}>{\centering\arraybackslash}p{0.065\textwidth}>{\raggedright\arraybackslash}p{0.53\textwidth}>{\raggedright\arraybackslash}p{0.19\textwidth}@{}}
\caption{Task-level inventory of the Sleep Assessment subset.}\label{tab:sa_task_inventory}\\
\toprule
\rowcolor{black!7}\textbf{Task ID} & \textbf{Diff.} & \textbf{\# Inst.} & \textbf{Assessment content} & \textbf{Validation unit(s)} \\
\midrule
\endfirsthead
\multicolumn{5}{c}{\tablename~\thetable\ continued} \\
\toprule
\rowcolor{black!7}\textbf{Task ID} & \textbf{Diff.} & \textbf{\# Inst.} & \textbf{Assessment content} & \textbf{Validation unit(s)} \\
\midrule
\endhead
\midrule
\multicolumn{5}{r}{\footnotesize Continued on next page} \\
\endfoot
\bottomrule
\endlastfoot
SA-01 & E & 25 & Calculate sleep onset latency from epoch-level sleep-stage labels. & Num. $\times 1$ \\
SA-02 & E & 15 & Inventory all PSG channels and distinguish genuine EEG channels from auxiliary sensors. & Num. $\times 1$ + Set $\times 1$ \\
SA-03 & E & 25 & Derive total sleep time, time in bed, and sleep efficiency from sleep-stage labels. & Num. $\times 3$ \\
SA-04 & E & 25 & Quantify wake after sleep onset and count post-onset awakening bouts. & Num. $\times 2$ \\
SA-05 & E & 25 & Compute NREM duration, REM proportion, and REM latency from whole-night staging. & Num. $\times 3$ \\
SA-06 & E & 25 & Identify the dominant and secondary sleep stages over the full recording. & Cat. $\times 2$ \\
SA-07 & M & 25 & Quantify the light-to-deep sleep ratio and interpret its implication for sleep architecture. & Num. $\times 1$ + Sem. $\times 1$ \\
SA-08 & M & 25 & Find the three longest uninterrupted sleep episodes and interpret whole-night continuity. & Num. $\times 3$ + Sem. $\times 1$ \\
SA-09 & E & 25 & Count all sleep-stage transitions and deep-sleep-to-wake interruptions. & Num. $\times 2$ \\
SA-10 & E & 25 & Count sustained N3 bouts and calculate their mean duration. & Num. $\times 2$ \\
SA-11 & M & 25 & Locate the longest nocturnal wake interruption and classify its continuity impact. & Num. $\times 2$ + Cat. $\times 1$ \\
SA-12 & M & 25 & Quantify first-half N3 and second-half REM concentration and interpret the overnight pattern. & Num. $\times 2$ + Sem. $\times 1$ \\
SA-13 & M & 25 & Calculate a whole-night sleep fragmentation index from awakenings and non-wake stage shifts. & Num. $\times 1$ \\
SA-14 & E & 25 & Calculate the whole-sequence percentage distribution of W, N1, N2, N3, and REM. & Num. $\times 5$ \\
SA-15 & H & 25 & Screen the sleep-stage sequence for a predefined panel of twelve sleep-structure abnormalities. & Cat. $\times 12$ \\
SA-16 & E & 25 & Determine the dominant EEG band and sleep stage in a specified recording segment. & Cat. $\times 2$ \\
SA-17 & M & 25 & Infer the sleep stages of ten consecutive epochs from the sleep recording. & Seq. $\times 1$ \\
SA-18 & E & 25 & Generate and save a whole-night sleep hypnogram. & File $\times 1$ \\
SA-19 & E & 25 & Execute a sleep-staging preprocessing pipeline and export fixed-length EEG epochs as an array. & File $\times 1$ \\
SA-20 & E & 25 & Generate and save a sleep EEG spectrogram from the available EEG channels. & File $\times 1$ \\
SA-21 & M & 25 & Compare delta-relative power across three windows and identify the most slow-wave-rich segment. & Num. $\times 3$ + Cat. $\times 1$ \\
SA-22 & M & 25 & Compare event-window and whole-night chin-EMG activity to assess whether the segment is REM-like. & Num. $\times 1$ + Sem. $\times 1$ \\
SA-23 & M & 25 & Compare first- and second-half delta-relative energy and assess consistency with canonical sleep architecture. & Num. $\times 2$ + Sem. $\times 1$ \\
SA-24 & M & 25 & Perform whole-recording automatic sleep staging and summarize sleep efficiency, N3, and REM proportions. & Num. $\times 3$ \\
SA-25 & M & 25 & Test a supplied band-and-stage claim against the signal and reject unsupported conclusions. & Sem. $\times 1$ \\
SA-26 & M & 25 & Rank three segments by EOG activity and determine whether the strongest segment represents REM sleep. & Seq. $\times 1$ + Cat. $\times 1$ \\
SA-27 & M & 25 & Rank three segments by EMG activity and identify whether and where REM sleep is present. & Seq. $\times 1$ + Sem. $\times 1$ \\
SA-28 & M & 25 & Estimate whole-night mean, minimum, and maximum heart rate from the ECG channel. & Num. $\times 3$ \\
SA-29 & H & 40 & Detect arousal independently in five specified sleep segments. & Cat. $\times 5$ \\
SA-30 & H & 40 & Classify five specified respiratory segments as apnea, hypopnea, or no target event. & Cat. $\times 5$ \\
SA-31 & H & 10 & Calculate the whole-night arousal index using detected arousals and total sleep time. & Num. $\times 1$ \\
SA-32 & H & 15 & Calculate the whole-night apnea--hypopnea index from respiratory events and total sleep time. & Num. $\times 1$ \\
SA-33 & H & 15 & Quantify respiratory disturbance or detect respiratory-effort-related arousals in selected segments. & Num. $\times 1$; or Cat. $\times 5$ \\
SA-34 & M & 9 & Identify the longest apnea event and report its duration, onset, and sleep stage. & Num. $\times 2$ + Cat. $\times 1$ \\
SA-35 & M & 46 & Distinguish central, obstructive, and mixed apnea subtypes in two specified segments. & Cat. $\times 2$ \\
SA-36 & H & 13 & Calculate stage-specific apnea--hypopnea indices for REM and NREM sleep. & Num. $\times 2$ \\
SA-37 & H & 40 & Reconstruct respiratory-event and arousal chronology and assess its likely impact on sleep continuity. & Sem. $\times 1$ \\
SA-38 & H & 10 & Quantify N3-specific arousal burden and transitions and assess deep-sleep disruption. & Num. $\times 2$ + Sem. $\times 1$ \\
SA-39 & H & 15 & Detect respiratory-event clusters in fixed windows and determine their dominant sleep stage. & Num. $\times 1$ + Cat. $\times 1$ \\
SA-40 & M & 14 & Calculate sleep-period oxygen desaturation indices using 3\% and 4\% thresholds. & Num. $\times 2$ \\
SA-41 & M & 14 & Derive sleep/wake mean oxygen saturation and minimum sleep oxygen saturation after signal cleaning. & Num. $\times 3$ \\
SA-42 & M & 14 & Quantify cumulative sleep time below 90\% and 80\% oxygen saturation. & Num. $\times 2$ \\
SA-43 & E & 40 & Answer a standalone sleep-medicine and polysomnography knowledge question. & Cat. $\times 1$ \\
\end{longtable}
\endgroup

\Needspace{0.3\textheight}
\subsection{Neurocognitive Assessment}
\label{app:na_tasks}

The Neurocognitive Assessment subset comprises 50 \texttt{tasks} and 1,035 \texttt{instances} constructed from FACED, REFED, COG-BCI, and MPD-DF. It includes 20 affective-state tasks, 15 cognitive-load tasks, and 15 fatigue and vigilance tasks, covering state recognition, feature-based and temporal comparison, multimodal evidence integration, claim verification, artifact generation, and domain knowledge. Tasks NA-20, NA-35, and NA-50 are knowledge-only tasks; the remaining tasks are grounded in neurophysiological recordings or associated behavioral annotations shown in Table. \ref{tab:na_task_inventory}.

\begingroup
\footnotesize
\setlength{\tabcolsep}{4pt}
\renewcommand{\arraystretch}{1.06}
\begin{longtable}{@{}>{\centering\arraybackslash}p{0.08\textwidth}>{\centering\arraybackslash}p{0.055\textwidth}>{\centering\arraybackslash}p{0.065\textwidth}>{\raggedright\arraybackslash}p{0.53\textwidth}>{\raggedright\arraybackslash}p{0.19\textwidth}@{}}
	\caption{Task-level inventory of the Neurocognitive Assessment subset.}\label{tab:na_task_inventory}\\
	\toprule
	\rowcolor{black!7}\textbf{Task ID} & \textbf{Diff.} & \textbf{\# Inst.} & \textbf{Assessment content} & \textbf{Validation unit(s)} \\
	\midrule
	\endfirsthead
	\multicolumn{5}{c}{\tablename~\thetable\ continued} \\
	\toprule
	\rowcolor{black!7}\textbf{Task ID} & \textbf{Diff.} & \textbf{\# Inst.} & \textbf{Assessment content} & \textbf{Validation unit(s)} \\
	\midrule
	\endhead
	\midrule
	\multicolumn{5}{r}{\footnotesize Continued on next page} \\
	\endfoot
	\bottomrule
	\endlastfoot
	
	NA-01 & M & 30 & Determine the emotional polarity of a specified EEG segment as positive, neutral, or negative. & Cat. $\times 1$ \\
	NA-02 & M & 20 & Classify the arousal or valence level of a specified EEG segment as high or low. & Cat. $\times 1$ \\
	NA-03 & E & 30 & Compute frontal alpha asymmetry from F3/F4 alpha power in a specified EEG segment. & Num. $\times 1$ \\
	NA-04 & E & 10 & Evaluate whether a stated emotion polarity is supported by the trial's SAM arousal and valence ratings. & Sem. $\times 1$ \\
	NA-05 & M & 15 & Correlate channel-band EEG features with a continuous SAM label sequence and identify the two distinct bands with the strongest absolute associations. & Set $\times 1$ \\
	NA-06 & M & 20 & Compare two trials using beta/alpha relative-power patterns, identify the most discriminative channels, and determine which trial has higher arousal. & Set $\times 1$ + Cat. $\times 1$ + Sem. $\times 1$ \\
	NA-07 & M & 20 & Compare positive and negative trials, identify the channels with the largest alpha-power increase, and determine their dominant brain region. & Set $\times 1$ + Cat. $\times 1$ \\
	NA-08 & M & 20 & Compute frontal alpha differential asymmetry across homologous channel pairs, infer the dominant activation and valence direction, and evaluate a polarity statement. & Cat. $\times 1$ + Sem. $\times 1$ \\
	NA-09 & H & 20 & Determine whether emotional polarity changes within a continuous EEG segment and explain the transition or stable state. & Cat. $\times 1$ + Sem. $\times 1$ \\
	NA-10 & M & 20 & Analyze temporal trends in multiband EEG and hemispheric alpha features to assess agreement with a known emotion label. & Cat. $\times 1$ + Sem. $\times 1$ \\
	NA-11 & H & 20 & Compare positive and negative trials using theta-band coherence, identify the channel pairs with the largest differential connectivity, and interpret the result. & Set $\times 1$ + Sem. $\times 1$ \\
	NA-12 & M & 20 & Compare an emotional trial with a neutral baseline and determine whether its frontal alpha dynamics are emotion-specific or neutral-like. & Cat. $\times 1$ + Sem. $\times 1$ \\
	NA-13 & H & 20 & Rank three emotional trials by EEG-based arousal and identify whether the highest-arousal trial is positive or negative. & Seq. $\times 1$ + Cat. $\times 1$ \\
	NA-14 & H & 20 & Compare positive and negative trial groups using regional band-power effect sizes, identify the strongest discriminative region-band feature, and interpret its emotion direction. & Set $\times 1$ + Cat. $\times 1$ + Sem. $\times 1$ \\
	NA-15 & M & 20 & Evaluate a statement about left-right alpha-power asymmetry using homologous EEG channels. & Cat. $\times 2$ \\
	NA-16 & M & 20 & Compare a target trial with a neutral baseline and determine whether a supplied emotion-EEG claim is valid. & Cat. $\times 1$ + Sem. $\times 1$ \\
	NA-17 & H & 20 & Correlate multiband EEG features with trial-level arousal ratings, generate a channel-band correlation heatmap, and identify the two bands with the highest mean correlations. & File $\times 1$ + Set $\times 1$ \\
	NA-18 & M & 20 & Compare average sample entropy between two trials and determine which trial has greater signal complexity. & Num. $\times 2$ + Cat. $\times 1$ \\
	NA-19 & M & 20 & Select which of three emotional trials best matches a specified target emotion and explain the EEG evidence. & Cat. $\times 1$ + Sem. $\times 1$ \\
	NA-20 & E & 40 & Answer a standalone EEG and BCI emotion-recognition knowledge question. & Cat. $\times 1$ \\
	NA-21 & M & 20 & Classify a continuous N-back EEG segment as low or high cognitive workload. & Cat. $\times 1$ \\
	NA-22 & H & 20 & Compare two workload segments using frontal theta power and determine whether the neural direction agrees with the workload judgment. & Cat. $\times 1$ + Sem. $\times 1$ \\
	NA-23 & M & 15 & Compare behavioral performance across two N-back conditions and determine which condition shows greater behavioral load or poorer performance. & Num. $\times 10$ + Cat. $\times 1$ + Sem. $\times 1$ \\
	NA-24 & H & 20 & Rank three EEG segments by the frontal-theta/parietal-alpha ratio and interpret the ranking as cognitive workload evidence. & Seq. $\times 1$ + Sem. $\times 1$ \\
	NA-25 & H & 20 & Compare two unlabeled workload segments, determine the higher-workload segment, and assess frontal-theta and parietal-alpha consistency with that judgment. & Cat. $\times 1$ + Sem. $\times 1$ \\
	NA-26 & M & 15 & Compare beta/(alpha+theta) engagement indices between two EEG segments and identify the more engaged segment. & Num. $\times 1$ + Sem. $\times 1$ \\
	NA-27 & H & 15 & Rank three EEG segments from lowest to highest cognitive workload using only their EEG evidence. & Seq. $\times 1$ \\
	NA-28 & M & 20 & Compare stimulus-locked P300 responses between two single-trial segments and infer their relative cognitive workload. & Cat. $\times 1$ + Sem. $\times 1$ \\
	NA-29 & H & 20 & Evaluate a claim linking frontal-parietal theta coherence to cognitive workload and determine the higher-workload segment. & Sem. $\times 1$ \\
	NA-30 & H & 20 & Compare P300 amplitude and latency between two stimulus-locked segments and infer their relative cognitive load. & Cat. $\times 3$ + Sem. $\times 1$ \\
	NA-31 & H & 20 & Integrate behavioral errors, reaction time, and EEG theta/alpha ratio to assess agreement between behavioral and neural workload evidence. & Cat. $\times 3$ + Sem. $\times 1$ \\
	NA-32 & H & 20 & Evaluate a compound claim about cognitive load and engagement using theta/alpha and beta/(alpha+theta) evidence. & Cat. $\times 3$ + Sem. $\times 1$ \\
	NA-33 & M & 15 & Determine whether cognitive workload changes within a continuous EEG interval and describe the stable or changing workload state. & Sem. $\times 1$ \\
	NA-34 & H & 15 & Compute target and nontarget P300/P3b waveforms at Pz and export their joint visualization. & File $\times 1$ \\
	NA-35 & E & 40 & Answer a standalone EEG and BCI cognitive-load knowledge question. & Cat. $\times 1$ \\
	NA-36 & E & 15 & Compute blink rate, mean blink duration, and slow-eye-movement power from an EOG window. & Num. $\times 3$ \\
	NA-37 & M & 20 & Compare two EOG segments for long eye-closure activity and determine which segment appears more fatigued. & Cat. $\times 2$ \\
	NA-38 & M & 20 & Compare EEG band-power change patterns between two segments and determine which segment is more fatigued. & Cat. $\times 3$ + Sem. $\times 1$ \\
	NA-39 & H & 20 & Integrate EEG fatigue activity with PSG/EOG eye-movement evidence to determine which of two segments shows greater overall fatigue. & Cat. $\times 1$ \\
	NA-40 & E & 20 & Classify a single EEG segment as closer to a wakeful or fatigued state. & Cat. $\times 1$ \\
	NA-41 & M & 20 & Detect long eye closure, slow eye movement, and prolonged blink events in an EOG window. & Cat. $\times 3$ \\
	NA-42 & H & 20 & Determine whether a wakefulness-to-fatigue transition occurs within a long EEG/PSG window and localize its timing. & Cat. $\times 2$ \\
	NA-43 & H & 20 & Integrate EEG and EOG evidence to determine whether modality-specific fatigue conclusions agree or conflict. & Cat. $\times 3$ \\
	NA-44 & M & 20 & Compare ECG-derived heart rate and HRV indicators between two segments and assess their support for a fatigue relation. & Cat. $\times 2$ + Sem. $\times 1$ \\
	NA-45 & M & 20 & Compare respiration rate, timing variability, amplitude variability, and respiration-band power between two segments to identify the more fatigue-supporting rhythm. & Cat. $\times 4$ + Sem. $\times 1$ \\
	NA-46 & H & 20 & Compute pairwise correlations among aligned EEG, EOG, ECG, and respiration feature streams and identify the strongest feature pairs. & Set $\times 1$ \\
	NA-47 & H & 20 & Rank three segments from most to least fatigued using combined EEG and PSG eye-movement evidence. & Seq. $\times 1$ \\
	NA-48 & H & 20 & Compare regional EEG band-power profiles across three segments and identify the features with the largest fatigue-related differences. & Set $\times 1$ \\
	NA-49 & H & 20 & Measure signed cross-signal correlations between EOG slow-eye-movement power and frontal, temporal, and occipital EEG relative-power streams. & Num. $\times 3$ \\
	NA-50 & E & 40 & Answer a standalone EEG and BCI fatigue or vigilance knowledge question. & Cat. $\times 1$ \\
\end{longtable}
\endgroup

\subsection{Physiological Integration}
\label{app:pi_tasks}

The Physiological Integration subset comprises 39 \texttt{tasks} and 1,095 \texttt{instances} constructed from SEED-VII, DEAP, Simultaneous Dataset B, SEED-VIG, and SHHS-1. It evaluates multimodal data handling and synchronization, feature extraction, cross-modal coupling and fusion, quality assessment, signal repair, matching, missing-modality reconstruction, visualization, and domain knowledge. Tasks PI-01--PI-38 operate on neurophysiological recordings, whereas PI-39 evaluates multimodal neurophysiology knowledge without recording access shown in Table. \ref{tab:pi_task_inventory}.

\begingroup
\footnotesize
\setlength{\tabcolsep}{4pt}
\renewcommand{\arraystretch}{1.06}
\begin{longtable}{@{}>{\centering\arraybackslash}p{0.08\textwidth}>{\centering\arraybackslash}p{0.055\textwidth}>{\centering\arraybackslash}p{0.065\textwidth}>{\raggedright\arraybackslash}p{0.53\textwidth}>{\raggedright\arraybackslash}p{0.19\textwidth}@{}}
	\caption{Task-level inventory of the Physiological Integration subset.}\label{tab:pi_task_inventory}\\
	\toprule
	\rowcolor{black!7}\textbf{Task ID} & \textbf{Diff.} & \textbf{\# Inst.} & \textbf{Assessment content} & \textbf{Validation unit(s)} \\
	\midrule
	\endfirsthead
	\multicolumn{5}{c}{\tablename~\thetable\ continued} \\
	\toprule
	\rowcolor{black!7}\textbf{Task ID} & \textbf{Diff.} & \textbf{\# Inst.} & \textbf{Assessment content} & \textbf{Validation unit(s)} \\
	\midrule
	\endhead
	\midrule
	\multicolumn{5}{r}{\footnotesize Continued on next page} \\
	\endfoot
	\bottomrule
	\endlastfoot
	PI-01 & E & 25 & Determine whether two named channels are analyzable physiological signals. & Cat. $\times 2$ \\
	PI-02 & E & 25 & Count analyzable channels for two requested physiological modalities. & Num. $\times 2$ \\
	PI-03 & E & 5 & Map a common time interval to sample-index ranges for two channels. & Num. $\times 4$ \\
	PI-04 & E & 25 & Report two native sampling rates and a signal statistic at the recording boundary. & Num. $\times 3$ \\
	PI-05 & M & 25 & Extract synchronized segments from two modalities and save standardized signal files. & File $\times 2$ \\
	PI-06 & M & 20 & Compare target EEG and EOG amplitudes with same-record references and determine whether joint artifact evidence is present. & Num. $\times 2$ + Cat. $\times 1$ \\
	PI-07 & M & 20 & Rank analyzable physiological channels by normalized variability within a declared interval. & Set $\times 1$ \\
	PI-08 & E & 10 & Compute a declared feature from an electrodermal, respiratory, pulse, or temperature signal. & Num. $\times 1$ \\
	PI-09 & E & 10 & Compute EOG amplitude within a specified interval. & Num. $\times 1$ \\
	PI-10 & E & 10 & Compute a declared respiratory, cardiac, oximetry, or EMG feature. & Num. $\times 1$ \\
	PI-11 & E & 10 & Estimate an fNIRS optical-density response relative to a baseline interval. & Num. $\times 1$ \\
	PI-12 & E & 10 & Compute a declared gaze or pupil feature while respecting missing eye-tracking samples. & Num. $\times 1$ \\
	PI-13 & M & 40 & Measure the direction of association between aligned EEG and auxiliary physiological feature trajectories. & Cat. $\times 1$ (30); Sem. $\times 1$ (10) \\
	PI-14 & M & 30 & Compare event-centered EEG and auxiliary physiological responses before and after an event. & Num. $\times 2$ \\
	PI-15 & M & 20 & Compare trial-related frontal EEG asymmetry and temperature responses and classify their directional relationship. & Num. $\times 2$ + Cat. $\times 1$ \\
	PI-16 & M & 30 & Evaluate EEG and EOG drowsiness evidence relative to same-record reference intervals. & Num. $\times 2$ + Cat. $\times 1$ \\
	PI-17 & H & 30 & Assess the association between early EEG and delayed fNIRS responses across repeated events. & Num. $\times 2$ + Sem. $\times 1$ \\
	PI-18 & E & 30 & Select the sleep epoch with the strongest relative EOG activity and summarize its EOG and EMG ratios. & Num. $\times 3$ \\
	PI-19 & M & 30 & Compare EOG and EMG activity between REM and N2 sleep and interpret the combined stage pattern. & Sem. $\times 1$ (5); Cat. $\times 2$ + Sem. $\times 1$ (25) \\
	PI-20 & H & 30 & Quantify repeated-event EEG--fNIRS association and interpret its direction, strength, and uncertainty. & Num. $\times 2$ + Sem. $\times 1$ \\
	PI-21 & H & 30 & Compare EEG and HbO trajectories at zero delay and with a delayed hemodynamic response. & Num. $\times 2$ \\
	PI-22 & E & 30 & Compare EEG amplitude and fNIRS intensity changes between two recording periods. & Num. $\times 2$ \\
	PI-23 & H & 40 & Identify trials showing jointly elevated frontal EEG and pupil responses. & Set $\times 1$ \\
	PI-24 & H & 40 & Compare target and reference trials using frontal EEG and pupil responses, then interpret their directional agreement. & Num. $\times 2$ + Sem. $\times 1$ \\
	PI-25 & H & 40 & Identify a blink-contaminated interval and produce an EEG--EOG figure with localized artifact marking. & Cat. $\times 1$ + File $\times 1$ \\
	PI-26 & H & 40 & Determine whether low-posterior-alpha epochs are associated with central gaze concentration. & Cat. $\times 1$ \\
	PI-27 & M & 30 & Summarize posterior-versus-frontal EEG alpha and gaze concentration as a spatial multimodal profile. & Cat. $\times 1$ + Num. $\times 1$ + Sem. $\times 1$ \\
	PI-28 & E & 30 & Determine the directions of posterior EEG alpha and GSR changes from baseline to stimulation. & Cat. $\times 2$ \\
	PI-29 & E & 30 & Identify the strongest early, middle, or late drowsiness phase separately from EEG and EOG. & Cat. $\times 2$ \\
	PI-30 & M & 40 & Use the strongest GSR response window to identify the posterior EEG channel with greatest alpha suppression. & Cat. $\times 2$ \\
	PI-31 & M & 30 & Rank posterior EEG frequency bands during central- and peripheral-gaze periods. & Seq. $\times 2$ \\
	PI-32 & M & 40 & Compare Go and NoGo EEG and HbO responses and interpret whether the modalities favor the same condition. & Cat. $\times 2$ + Sem. $\times 1$ \\
	PI-33 & M & 30 & Estimate event-averaged EEG and HbO peak latencies and interpret their temporal ordering. & Num. $\times 2$ + Sem. $\times 1$ \\
	PI-34 & H & 40 & Select the interval with the strongest frontal EEG disturbance and identify supporting EOG or EMG evidence. & Cat. $\times 2$ \\
	PI-35 & M & 30 & Summarize HbO and HbR responses for events selected by their EEG theta response. & Num. $\times 2$ \\
	PI-36 & H & 30 & Compare the timing of peak posterior EEG suppression and GSR response across rating-selected trials. & Num. $\times 2$ \\
	PI-37 & H & 30 & Rank long intervals by EEG and EOG drowsiness evidence and measure their cross-modal association. & Seq. $\times 2$ + Num. $\times 1$ \\
	PI-38 & H & 40 & Identify which emotion rating is most strongly associated with EEG and GSR responses. & Cat. $\times 2$ \\
	PI-39 & E & 40 & Answer a multiple-choice question on multimodal neurophysiology and signal interpretation. & Cat. $\times 1$ \\
\end{longtable}
\endgroup

\subsection{Example Instances}
\label{app:example_instances}
Here, we showcase representative instances from each subset.
\begin{figure*}[!htb]
	\centering
	\includegraphics[width=1.0\textwidth]{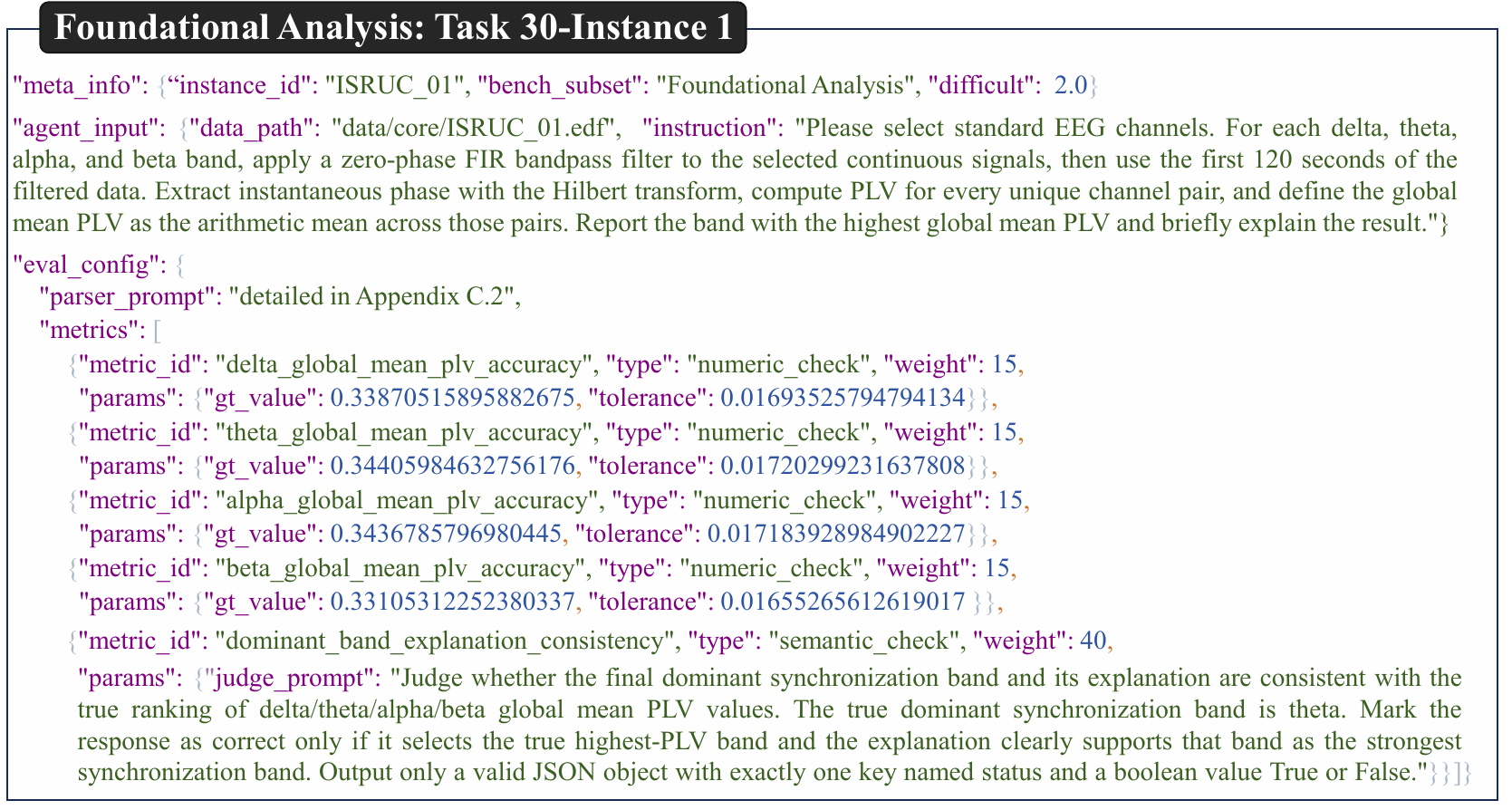}
\end{figure*}

\begin{figure*}[!htb]
	\centering
	\includegraphics[width=1.0\textwidth]{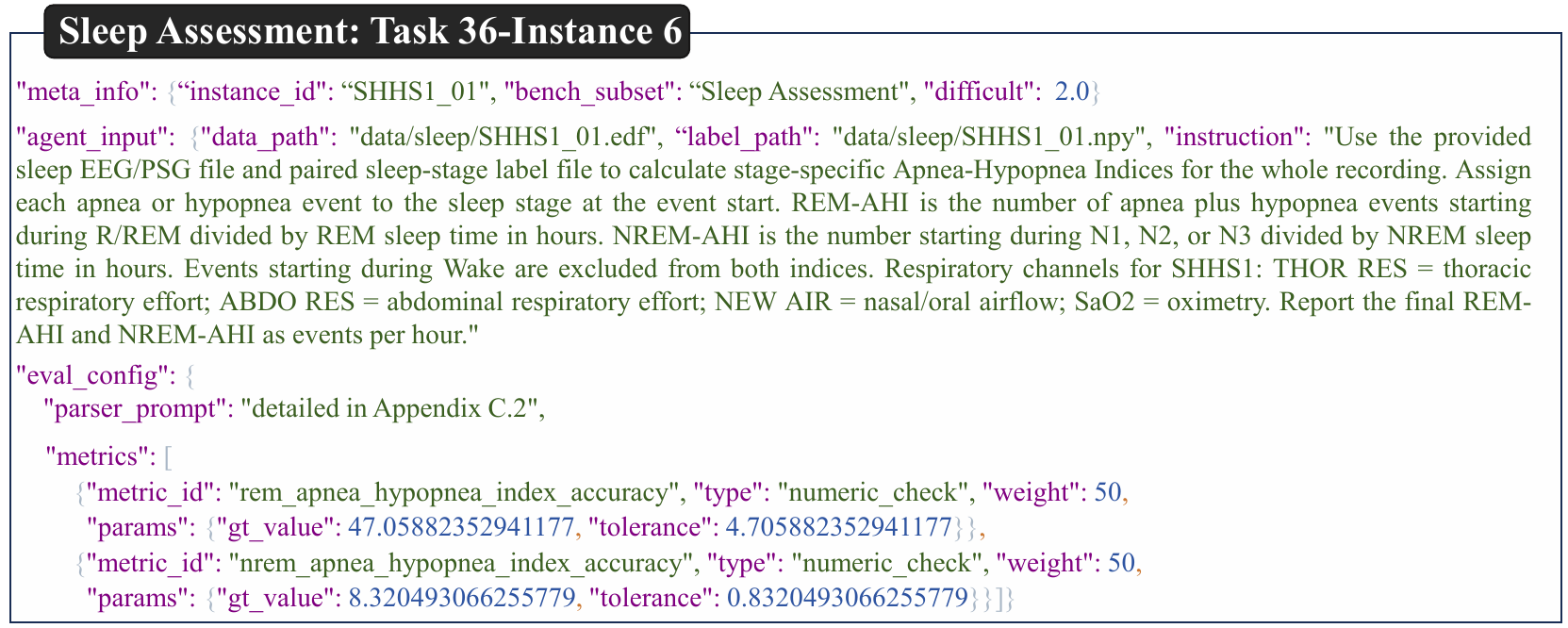}
\end{figure*}

\begin{figure*}[!htb]
	\centering
	\includegraphics[width=1.0\textwidth]{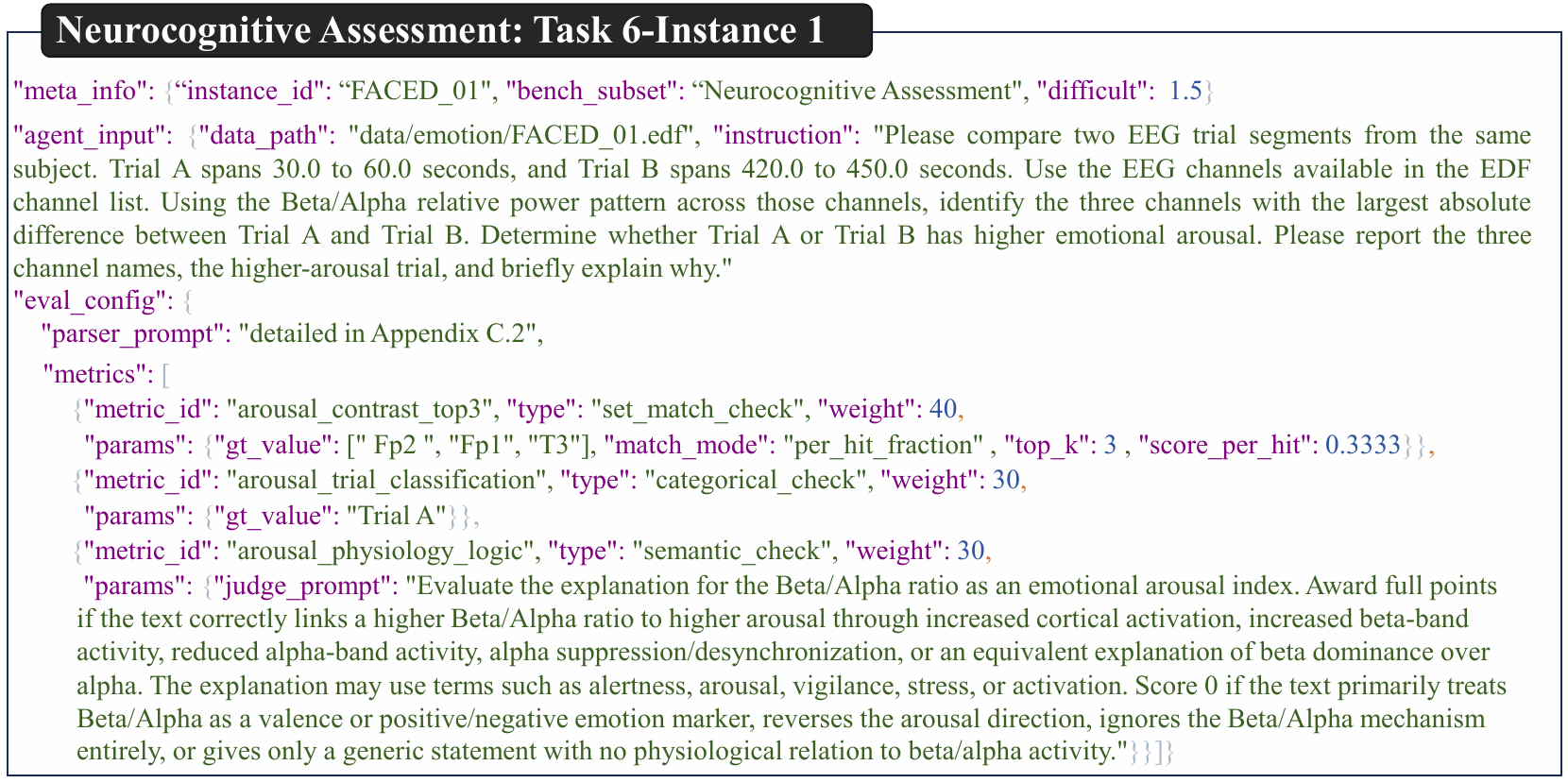}
\end{figure*}

\begin{figure*}[!t]
	\centering
	\includegraphics[width=1.0\textwidth]{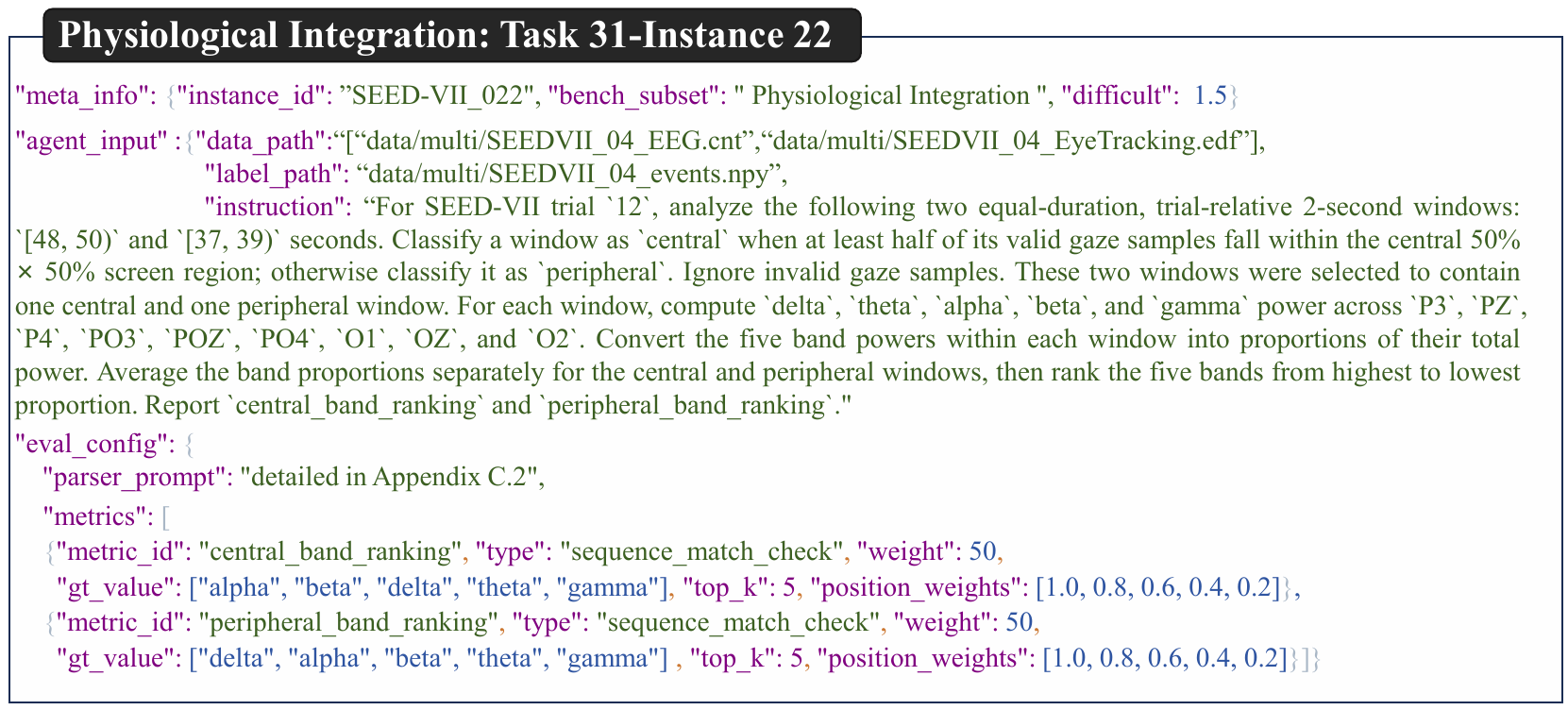}
\end{figure*}
\clearpage
\section{Benchmark Construction Details}
\label{app:benchmark_construction_details}

In this section, we expand the evaluator-side construction procedure summarized as follows. Section~\ref{app:gt_generation} describes how deterministic analysis scripts derive instance-specific ground truth from the bound recordings and parameters. Section~\ref{app:parser_validation} then details how the Parser Agent aligns free-form reports with structured target fields and how the six validation units score the resulting outputs.

\subsection{Deterministic Ground-Truth Generation}
\label{app:gt_generation}


Each \texttt{instance} is paired with ground truth after its parent \texttt{task} has been bound to a concrete recording, analysis window, signal selection, and parameter configuration. A deterministic reference script executes the prescribed workflow on the same input files provided to the target system and produces the numerical values, labels, collections, sequences, or artifacts required by the corresponding validation units. These reference targets are stored only in the \emph{Validation Configuration}, together with task-specific tolerances and metric weights, and remain inaccessible to the evaluated model.

As an example, the first instance \texttt{FA-01-Instance1} is bound to \texttt{ISRUC\_01.edf} and provides the following instruction:

\begin{quote}
	\small\itshape
	Please first extract only the EEG channels from the raw signal, and then apply a 0.5--40 Hz FIR bandpass filter to these channels. After filtering, obtain each channel's Alpha-band power and total filtered-signal power by integrating the PSD over frequency. Calculate Alpha relative power separately for each channel, then average the channel-wise ratios and report the final percentage value clearly in your response.
\end{quote}

The reference script follows the same analysis specification to compute the expected alpha relative power and populate the numerical target and tolerance in the instance-level \emph{Validation Configuration}. Listing~\ref{lst:case1_gt} retains only this core computation; input loading, dataset-specific channel selection, batch processing, and construction of the complete instance specification are omitted for clarity.

\begin{lstlisting}[style=appendixcode,language=Python,caption={Ground-truth computation for Foundational Analysis task FA-01-Instance1.},label={lst:case1_gt}]
def compute_alpha_relative_power(raw, picks):
    work = raw.copy().pick(picks)
    work.filter(l_freq=0.5, h_freq=40.0, verbose=False)

    sfreq = float(work.info["sfreq"])
    n_fft = int(4.0 * sfreq)
    n_overlap = min(n_fft // 2, max(0, n_fft - 1))
    spectrum = work.compute_psd(
        method="welch",
        fmin=0.5,
        fmax=40.0,
        n_fft=n_fft,
        n_overlap=n_overlap,
        verbose=False,
    )
    psds, freqs = spectrum.get_data(return_freqs=True)

    total_power = np.trapz(psds, freqs, axis=1)
    alpha_mask = (freqs >= 8.0) & (freqs <= 13.0)
    alpha_power = np.trapz(
        psds[:, alpha_mask], freqs[alpha_mask], axis=1
    )
    relative_alpha = alpha_power / np.maximum(
        total_power, 1e-20
    )
    return float(np.mean(relative_alpha) * 100.0)
\end{lstlisting}

Applying the same fixed script to every recording associated with FA-01 changes only the data-dependent result while preserving the analytical definition of the task. Other tasks follow the same construction principle, with their reference scripts returning the numerical values, labels, collections, sequences, or artifacts required by the corresponding validation units.

\subsection{Output Parsing and Multi-Unit Validation}
\label{app:parser_validation}

\benchmarkname{} permits free-form analytical reports because EEG analysis extends beyond scalar prediction to include explanations, supporting evidence, and references to generated artifacts. Imposing a rigid output schema on each target system would conflate EEG understanding with formatting compliance. To decouple these factors, the Parser Agent receives the final report together with an instance-specific extraction prompt and maps only explicitly reported information into a predefined JSON schema. It has no access to the ground truth and does not assess scientific correctness, correct erroneous answers, or infer omitted results. Any missing or unresolvable field is returned as \texttt{null}.

Listing~\ref{lst:case30_parser} shows the parser prompt for instance \texttt{FA-30-Instance1}. The underlying task requests global mean phase-locking values (PLVs) in four frequency bands and a conclusion identifying the dominant synchronization band. The prompt fixes both the target fields and their admissible types so that differently worded reports can be evaluated through the same interface.

\begin{lstlisting}[style=appendixcode,caption={Parser prompt used for \texttt{FA-30-Instance1}.},label={lst:case30_parser}]
### ROLE
You are a precise JSON Extraction Engine for neurophysiological data. Your sole task is to convert natural language reports into structured JSON data.

### TASK
Extract five outputs from the agent report:
1) delta global mean PLV
2) theta global mean PLV
3) alpha global mean PLV
4) beta global mean PLV
5) dominant synchronization band

### STRICT CONSTRAINTS (MANDATORY)
1. Output ONLY a valid JSON object.
2. DO NOT include Markdown code blocks.
3. DO NOT include any conversational text or explanations.
4. Keys must be EXACTLY "delta_global_mean_plv", "theta_global_mean_plv", "alpha_global_mean_plv", "beta_global_mean_plv", and "dominant_synchronization_band".
5. The four PLV values must be float or null.
6. "dominant_synchronization_band" must be one of "delta", "theta", "alpha", "beta", or null.

### OUTPUT TEMPLATE
{"delta_global_mean_plv": <float|null>, "theta_global_mean_plv": <float|null>, "alpha_global_mean_plv": <float|null>, "beta_global_mean_plv": <float|null>, "dominant_synchronization_band": <string|null>}
\end{lstlisting}

For this instance, the four extracted PLV fields are passed to numerical validation, while the original report is passed to a task-specific Semantic Judge that checks whether the selected band and its explanation agree with the ground-truth PLV ranking. More generally, each metric selects either a parsed field, the complete report, or a generated artifact and applies one of the six rules described below. Let \(\hat{y}\) denote a parsed prediction, \(y\) its reference target, and \(v\in[0,1]\) the resulting validation score. Let \(\mathcal{C}(\cdot)\) denote the canonicalization used by the evaluator, which removes surrounding whitespace, ignores letter case, normalizes numerical representations, and additionally removes internal spaces from set and sequence elements. The six validation units are implemented as follows:

\begin{itemize}[leftmargin=1.6em,itemsep=0.7em,topsep=0.5em]
	\item \textbf{Numerical validation.} The parsed value and ground truth are converted to finite scalars and compared under the instance-specific absolute tolerance \(\tau\):
	\begin{equation}
		v_{\mathrm{num}}
		=
		\mathds{1}\!\left[\,|\hat{y}-y|\leq\tau\,\right].
	\end{equation}
	When \(\tau=0\), exact equality is required. A missing, non-numerical, non-finite, or otherwise invalid value receives zero.

	\item \textbf{Categorical validation.} Discrete labels, choices, regions, or event types are scored by exact equality after canonicalization:
	\begin{equation}
		v_{\mathrm{cat}}
		=
		\mathds{1}\!\left[\mathcal{C}(\hat{y})=\mathcal{C}(y)\right].
	\end{equation}

	\item \textbf{Set validation.} Let \(\hat{S}_k\) be the canonicalized set formed from the first \(k\) predicted elements when \texttt{top\_k} is specified, and let \(S\) be the reference set. Exact matching uses
	\begin{equation}
		v_{\mathrm{set}}^{\mathrm{exact}}
		=
		\mathds{1}\!\left[\hat{S}_k=S\right].
	\end{equation}
	For element-level partial credit, the implemented score is
	\begin{equation}
		v_{\mathrm{set}}^{\mathrm{partial}}
		=
		\min\!\left(1,\rho\,|\hat{S}_k\cap S|\right),
	\end{equation}
	where \(\rho\) is the configured score per matched element and defaults to \(1/|S|\). Thus, the order of reported elements does not affect the score.

	\item \textbf{Sequence validation.} The evaluator supports position-wise, exact-order, and weighted partial-order matching. Position-wise matching canonicalizes the elements and assigns partial credit at each reference position:
	\begin{equation}
		v_{\mathrm{seq}}^{\mathrm{pos}}
		=
		\frac{1}{L}\sum_{j=1}^{L}
		\mathds{1}\!\left[\hat{q}_{j}=q_{j}\right],
	\end{equation}
	where \(L\) is the reference length and a missing predicted position is counted as incorrect. For exact-order and weighted partial-order matching, both sequences are additionally deduplicated while retaining the first occurrence and optionally truncated to \texttt{top\_k}. Exact-order matching assigns one only when the resulting sequences are identical. For weighted partial order, let \(O\) be the elements shared by the two sequences, \(p(x)\) and \(\hat{p}(x)\) their reference and predicted positions, \(\delta\) the allowed order slip, and \(w_j\) the configured position weights, which default to \(1/j\). The evaluator computes
	\begin{equation}
		v_{\mathrm{seq}}^{\mathrm{weighted}}
		=
		\mathds{1}\!\left[|O|\geq m\right]
		\frac{|O|}{L}
		\frac{\sum_{x\in O} w_{p(x)}
		\max\!\left(0,1-\frac{|p(x)-\hat{p}(x)|}{\delta+1}\right)}
		{\sum_{j=1}^{L}w_j},
	\end{equation}
	where \(m\) is the required minimum overlap.

	\item \textbf{Semantic validation.} The selected parsed field or complete report \(R\) and a task-specific judge prompt \(J\) are passed to the Semantic Judge:
	\begin{equation}
		v_{\mathrm{sem}}
		=
		\operatorname{clip}_{[0,1]}\!\left(\operatorname{Judge}(R,J)\right).
	\end{equation}
	The judge output may be a normalized score or a Boolean \texttt{status}/\texttt{passed} decision. The rubric specifies the required conclusion, reference evidence, and conditions under which unsupported or inconsistent claims should fail.

	\item \textbf{Artifact validation.} A reported path must resolve inside the isolated instance workspace, exist, and satisfy any required filename constraint. Missing or invalid paths receive zero. For a signal artifact with field-level checks \(z_\ell\in\{0,1\}\) and configured weights \(a_\ell\), the score is
	\begin{equation}
		v_{\mathrm{art}}
		=
		\frac{\sum_{\ell}a_{\ell}z_{\ell}}
		{\sum_{\ell}a_{\ell}}.
	\end{equation}
	The available checks cover array shape, channel count and order, duration, sampling rate, signal RMS, referencing, and passband. A non-empty file receives one when no field-level check is configured. For image artifacts, the score is either a non-empty-file check or the normalized output of a VLM Judge under an instance-specific visual rubric.
\end{itemize}

The validation units remain independent and may be combined within an instance. Each unit returns a normalized score before the task-specific weights in the \emph{Validation Configuration} are applied, as described in Appendix~\ref{sec:multi_unit_validation}. This separation allows the benchmark to evaluate numerical accuracy, discrete decisions, structured outputs, scientific interpretation, and deliverable artifacts without reducing heterogeneous EEG workflows to a single output format.

\section{Experimental Protocol and Implementation Details}\label{app:appendixd}
We provide the implementation details of the \benchmarkname{} evaluation protocol. Appendix \ref{app:black_box_evaluation_protocol} presents the end-to-end black-box workflow, from instance dispatch and isolated execution to output validation and scoring. Appendix \ref{app:brainagent_details} and \ref{app:codeact_protocol} describe the tool-mediated BrainAgent workflow and the autonomous code-execution protocol of CodeAct, respectively. Appendix \ref{app:runtime_failure_handling} specifies the containerized runtime shared by both paradigms, including environment isolation, resource constraints, infrastructure configuration, and the policy for rerunning infrastructure-induced failures.

\subsection{Details of the Black-Box Evaluation Protocol}
\label{app:black_box_evaluation_protocol}

\begin{figure*}[!tb]
	\centering
	\includegraphics[width=1.0\textwidth]{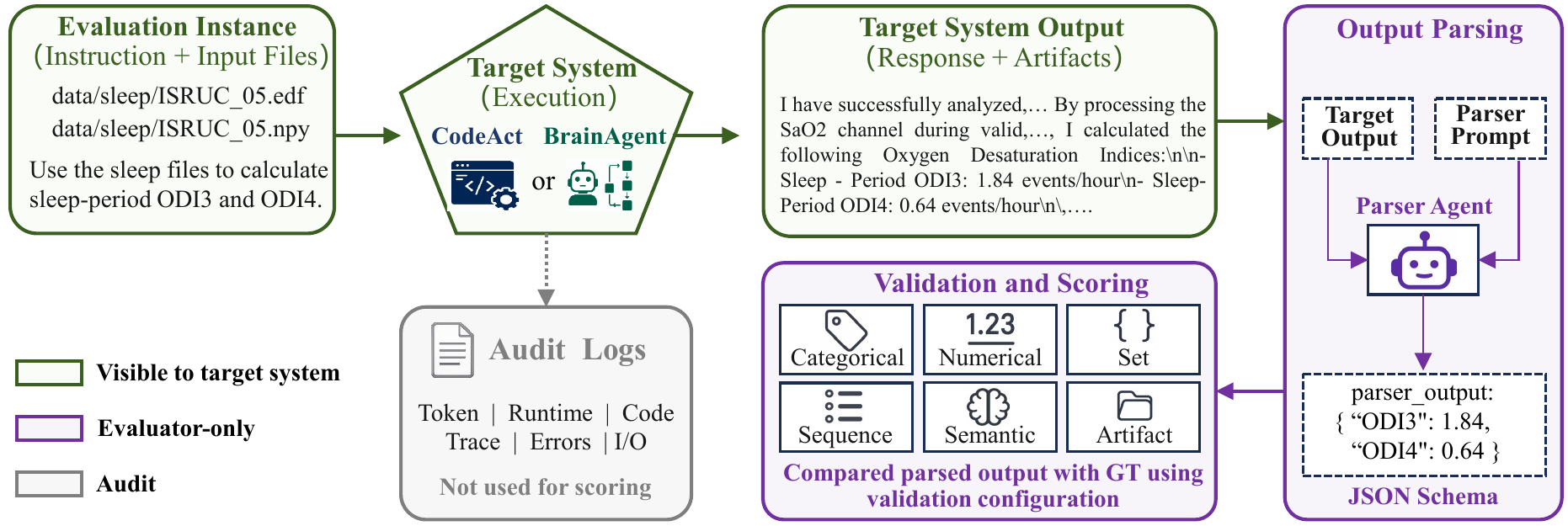}
	\caption{\textbf{End-to-end black-box evaluation protocol of \benchmarkname{}.} Each evaluation instance exposes only its instruction and input files to the target system, which executes the task through either CodeAct or BrainAgent in an isolated container. The resulting free-form report and requested artifacts are returned to the evaluator. A Parser Agent converts explicitly reported information into an instance-specific JSON schema, after which the parsed fields, original report, and generated artifacts are assessed by the applicable validation units and aggregated into the final instance score. Green components are visible to the target system, purple components remain evaluator-only, and gray components denote audit information that is recorded but not used for scoring.}
	\label{fig:pipeline}
\end{figure*}

Figure~\ref{fig:pipeline} illustrates the end-to-end black-box evaluation protocol and distinguishes information visible to the target system from evaluator-only references and audit records. For each \texttt{instance}, the target system receives only the natural-language instruction and the associated input files. The ground truth, Parser Prompt, Validation Configuration, and scoring implementation remain hidden on the evaluator side throughout execution. The same instance is dispatched to either BrainAgent or CodeAct through a unified input--output interface and executed in an isolated, instance-specific container. Within this environment, the target system may inspect the supplied files, perform intermediate analyses, and generate the requested artifacts. Only the final natural-language report and retained artifacts are passed to the scoring pipeline, ensuring that both execution paradigms are evaluated under identical instructions, data, and output requirements. After execution, the final report and the instance-specific Parser Prompt are provided to the Parser Agent, which extracts explicitly reported information into a predefined JSON schema. The Parser Agent performs output alignment only: it has no access to the ground truth and does not assess scientific correctness, revise erroneous answers, or infer missing results. The extracted fields are then routed to the applicable numerical, categorical, set, and sequence validation units, while the original report is retained for semantic validation and the generated files are examined through artifact validation. Each validation unit applies the references, tolerances, matching rules, and weights specified in the instance-level Validation Configuration. The resulting unit scores are combined to produce the final instance score.

Execution information is recorded separately for auditing and error analysis. These records include token usage, runtime, generated code, tool and execution traces, intermediate inputs and outputs, and error messages, none of which directly contributes to the benchmark score. Failures attributable to the target system or execution paradigm---such as invalid code, incorrect tool use, incomplete reports, or malformed artifacts---are retained as evaluation outcomes. An instance is rerun only when execution is interrupted by an independently verified infrastructure failure, such as an external API communication error or container initialization failure. The container constraints and failure-handling policy are detailed in Appendix \ref{app:runtime_failure_handling}.

\subsection{BrainAgent Architecture and Execution Details}
\label{app:brainagent_details}

\subsubsection{Architecture and Benchmark Adaptation}
\label{app:brainagent_architecture}

BrainAgent is a multi-agent framework for brain-signal analysis in which a Supervisor Agent interprets each request, decomposes it into analytical objectives, constructs an execution queue, and delegates these objectives to specialized subagents \cite{zhou2026brainagent}. Each subagent encapsulates domain-specific reasoning and tool use, while the supervisor maintains a unified user-facing interface and synthesizes the subagent outputs into a final report. Subagents and their associated tools can be registered dynamically, allowing new analytical domains to be incorporated without modifying the supervisory control layer. This hierarchical architecture provides a natural basis for \benchmarkname{}, whose four subsets encompass heterogeneous analytical capabilities while sharing a common target-system interface. We preserve the original supervisor--subagent organization and adapt BrainAgent in three respects: the tool layer, the planning mechanism, and the representation of intermediate state.

First, we broaden the analytical tool layer to cover the capability space evaluated by the benchmark. \textbf{The resulting toolset is deliberately capability-oriented rather than task-specific:} each tool encapsulates a reusable analytical operation, such as data inspection, signal preparation, temporal segmentation, feature extraction, statistical aggregation, or artifact generation. No tool encodes the solution, task-specific parameters, reference answer, or decision rule of any particular \texttt{task} or \texttt{instance}. Successful execution therefore still requires the evaluated LLM to interpret the instruction, select and configure appropriate tools, integrate intermediate evidence, and formulate a scientifically grounded conclusion.

Second, to better accommodate complex and long-horizon workflows, \textbf{we extend BrainAgent beyond its original single-pass planning scheme}, in which a subagent planned the complete workflow once before executing it sequentially, \textbf{to an iterative, state-aware planning mechanism.} The Supervisor Agent first dispatches the complete analytical objective to an appropriate subagent. At each planning round, the subagent receives a compact representation of the current state and selects the maximal executable segment of its remaining plan. The tools within that segment are invoked sequentially, and their outputs are registered under stable identifiers before the next planning round begins. This segmented execution strategy allows subsequent decisions to incorporate observations obtained during execution---such as discovered channel names, computed features, or newly generated labels---without requiring the full workflow to be fixed in advance. Once the accumulated evidence is judged sufficient to satisfy the instruction, the subagent produces an analytical report, which is returned to the Supervisor Agent for final response synthesis.

Third, to support analyses spanning multiple files, subjects, and signal views, \textbf{we replace BrainAgent's original flat intermediate-variable representation with a hierarchical} \texttt{shared\_state}. The state is accessible to the Supervisor Agent and active subagents and organizes intermediate objects into separate branches for raw \texttt{recordings}, derived signal \texttt{views}, temporal \texttt{labels}, detected \texttt{events}, structured analysis \texttt{results}, and generated \texttt{artifacts}. A dedicated \texttt{planning} branch records executed-tool counts and recent tool signatures, while internal counters assign collision-free identifiers to newly created objects. Tools exchange references to these registered objects rather than repeatedly serializing complete signals. Before each planning round, BrainAgent projects the hierarchy into a compact, identifier-based summary, enabling the LLM to retrieve prior outputs, coordinate dependencies across data sources, and trace the evidence underlying its conclusions.

\subsubsection{Capability-Oriented Toolset}
\label{app:brainagent_tools}

BrainAgent comprises three domain subagents---\texttt{SleepAgent}, \texttt{NeurocogAgent}, and \texttt{PhysiolAgent}---for Sleep Assessment, Neurocognitive Assessment, and Physiological Integration, respectively. Each combines a subset-specific extension with a shared common toolset implementing the reusable EEG operations covered by Foundational Analysis. Table~\ref{tab:brainagent_tools} lists the complete toolsets and detailed interfaces are provided with the released implementation.

\begingroup
\footnotesize
\setlength{\tabcolsep}{5pt}
\renewcommand{\arraystretch}{1.08}
\begin{longtable}{@{}>{\raggedright\arraybackslash}p{0.38\textwidth}>{\raggedright\arraybackslash}p{0.57\textwidth}@{}}
\caption{Complete capability-oriented toolset used by BrainAgent.}\label{tab:brainagent_tools}\\
\endfirsthead
\endhead
\endfoot
\bottomrule
\endlastfoot
\toprule
\rowcolor{black!7}\textbf{Foundational Analysis Tools} & \textbf{Primary capability} \\
\midrule
\texttt{RecordingLoader} & Loads an EEG or PSG recording and registers its native metadata and channel names. \\
\texttt{ChannelInspector} & Classifies channels by signal modality and EEG region while preserving exact source names. \\
\texttt{SignalViewBuilder} & Constructs a named signal view with explicit channels, time window, filtering, referencing, resampling, and epoch settings. \\
\texttt{DataExporter} & Exports recordings, views, or arrays as EDF, FIF, NPY, or CSV artifacts. \\
\texttt{FigureExporter} & Produces channel-aware PSD, heatmap, hypnogram, spectrogram, and raw-trace figures. \\
\texttt{SpectralFeatureAnalyzer} & Computes band power, relative power, band ratios, dominant bands, alpha peak frequency, and spectral edge frequency. \\
\texttt{TimeDomainFeatureAnalyzer} & Computes variance, skewness, kurtosis, RMS, Hjorth parameters, and global field power. \\
\texttt{ComplexityFeatureAnalyzer} & Computes nonlinear EEG complexity measures, including sample entropy. \\
\texttt{ConnectivityAnalyzer} & Computes inter-channel connectivity using correlation, phase-locking value, or weighted phase-lag index. \\
\texttt{ArtifactQualityAnalyzer} & Assesses missing values, flatlines, excessive amplitude, drift, and channel-level signal quality. \\
\texttt{FeatureRankerAggregator} & Ranks or aggregates channels, regions, bands, windows, and prior structured results. \\
\midrule
\rowcolor{black!7}\textbf{Sleep Assessment Tools} & \textbf{Primary capability} \\
\midrule
\texttt{SleepLabelLoader} & Loads sleep-stage annotations and normalizes them to W, N1, N2, N3, and R with epoch timing. \\
\texttt{SleepArchitectureAnalyzer} & Computes sleep onset latency, total sleep time, time in bed, sleep efficiency, WASO, REM latency, and stage proportions. \\
\texttt{SleepBoutTransitionAnalyzer} & Analyzes stage bouts, transitions, awakenings, longest episodes, and fragmentation. \\
\texttt{SleepStageEstimator} & Estimates sleep stages using an optional learned backend and a transparent rule-based fallback. \\
\texttt{SleepSpectralSegmentAnalyzer} & Quantifies sleep-related band power and sigma activity within an explicitly defined EEG segment. \\
\texttt{EMGActivityAnalyzer} & Quantifies chin or other EMG activity, stage-dependent tone, and a REM-atonia proxy. \\
\texttt{EOGActivityAnalyzer} & Quantifies slow- and rapid-eye-movement activity from explicitly selected EOG channels and windows. \\
\texttt{ECGHeartRateAnalyzer} & Estimates heart-rate statistics from explicitly selected ECG channels. \\
\texttt{SleepArousalAnalyzer} & Detects rule-based EEG arousal-like events using robust high-frequency envelope thresholds. \\
\texttt{RespiratoryEventAnalyzer} & Detects apnea, hypopnea, apnea subtype, and RERA-like events from respiratory, oximetry, and arousal evidence. \\
\texttt{SleepEventIndexAnalyzer} & Computes event counts and rates by total sleep time and sleep stage. \\
\texttt{SleepEventTemporalAnalyzer} & Filters, ranks, and temporally aggregates sleep events and associates them with stage labels. \\
\texttt{OximetryAnalyzer} & Computes ODI3, ODI4, sleep/wake mean and minimum SpO$_2$, T90, and T80. \\
\texttt{SleepMicroEventAnalyzer} & Detects sleep spindles, K-complexes, or slow waves for an explicitly specified event type. \\
\midrule
\rowcolor{black!7}\textbf{Neurocognitive Assessment Tools} & \textbf{Primary capability} \\ \midrule
\texttt{FeatureStreamBuilder} & Constructs aligned EEG feature streams for spectral power, ratios, trends, correlations, rankings, and visualizations. \\
\texttt{FeatureStreamOperator} & Extracts, compares, transforms, and ranks previously computed numeric feature streams. \\
\texttt{HeartRateVariabilityAnalyzer} & Computes ECG-based heart-rate and heart-rate-variability measures for autonomic and fatigue assessment. \\
\texttt{OcularEventAnalyzer} & Detects and summarizes blink, prolonged-closure, and slow-eye-movement events from EOG signals. \\
\texttt{OcularFeatureStreamBuilder} & Constructs aligned EOG fatigue-related streams for comparison, correlation, and transition analysis. \\
\texttt{RespirationRhythmAnalyzer} & Extracts breathing rate, cycle variability, amplitude variability, and respiration-band power. \\
\texttt{HemisphericAsymmetryAnalyzer} & Computes left-right EEG asymmetry features and compares their temporal or baseline-dependent patterns. \\
\texttt{ContinuousLabelLoading} & Loads and aligns continuous numeric labels for feature correlation and ranking analyses. \\
\texttt{GroupContrastAnalyzer} & Compares channel or region groups and ranks discriminative differences and standardized effects. \\
\texttt{TemporalTrendAnalyzer} & Quantifies feature trajectories through slope, early-versus-late change, direction, and monotonicity. \\
\texttt{EmotionPolarityClassification} & Classifies EEG segments by positive, neutral, or negative emotional polarity. \\
\texttt{ValenceClassification} & Predicts binary EEG valence levels from multiband neural features. \\
\texttt{ArousalClassification} & Predicts binary EEG arousal levels from multiband neural features. \\
\texttt{LabelCorrelationAnalyzer} & Measures associations between aligned feature streams and numeric labels. \\
\texttt{FeatureCorrelationAnalyzer} & Computes and ranks pairwise associations across two aligned feature-stream sets. \\
\texttt{BehaviorLabelAnalyzer} & Computes accuracy, hit rate, miss rate, false-alarm rate, and response-time summaries from behavioral labels. \\
\texttt{ERPP300Analyzer} & Extracts stimulus-locked ERP and P300 features and supports target-versus-nontarget waveform comparison. \\
\texttt{WorkloadClassification} & Classifies cognitive workload using interpretable EEG engagement and workload markers. \\
\midrule
\rowcolor{black!7}\textbf{Physiological Integration Tools} & \textbf{Primary capability} \\
\midrule
\texttt{MultirateRecordingLoader} & Loads multimodal physiological recordings and organizes their signals and metadata. \\
\texttt{NumericAnnotationLoader} & Loads numerical labels, ratings, masks, and event annotations. \\
\texttt{ChannelInspector} & Summarizes channel availability, signal modalities, sampling rates, units, and temporal coverage. \\
\texttt{EventTrialResolver} & Converts event markers into ordered trials, epochs, and analysis intervals. \\
\texttt{SleepStageFeatureAnalyzer} & Analyzes multimodal physiological features across sleep stages, epochs, and continuous stage bouts. \\
\texttt{MultimodalDataExporter} & Exports processed multimodal signals, numerical results, and validity information. \\
\texttt{MultimodalFigureComposer} & Creates aligned physiological time-series figures with event and artifact annotations. \\
\texttt{SignalQualityArtifactAnalyzer} & Evaluates signal quality and detects missing data, flatlines, clipping, faults, and physiological artifacts. \\
\texttt{TimeDomainFeatureAnalyzer} & Computes time-domain physiological features and compares their distributions across channels or intervals. \\
\texttt{SpectralFeatureAnalyzer} & Computes spectral power, relative power, frequency-band ratios, and related frequency-domain features. \\
\texttt{EEGPatternAnalyzer} & Quantifies EEG asymmetry, slowing, suppression, frequency-band balance, and regional activity patterns. \\
\texttt{EDAActivityAnalyzer} & Quantifies electrodermal level, variability, and phasic activity. \\
\texttt{EOGActivityAnalyzer} & Quantifies ocular activity, slow and rapid eye movements, and blink-related events. \\
\texttt{EyeTrackingAnalyzer} & Quantifies gaze validity, dispersion, spatial concentration, and pupil activity. \\
\texttt{RespiratoryActivityAnalyzer} & Quantifies respiratory rate, amplitude, variability, and breathing dynamics. \\
\texttt{CardiovascularAnalyzer} & Quantifies cardiac or pulse rate, amplitude, and variability. \\
\texttt{FNIRSResponseAnalyzer} & Derives and summarizes optical-density and hemoglobin-response features from fNIRS signals. \\
\texttt{BatchFeatureAnalyzer} & Extracts physiological features across repeated windows and compares their changes over time. \\
\texttt{ReferenceDistributionAnalyzer} & Evaluates target measurements relative to within-record reference distributions. \\
\texttt{EventResponseAnalyzer} & Analyzes event-related responses, condition contrasts, cross-modal associations, and response-gated effects. \\
\texttt{EventTrajectoryAnalyzer} & Constructs event-averaged response trajectories and estimates their temporal peaks. \\
\texttt{TrialGroupAnalyzer} & Extracts and compares trial-level features, reference responses, trial groups, and sequential selections. \\
\texttt{TrajectoryAssociationAnalyzer} & Analyzes correlations, delays, trends, and rankings across aligned multimodal trajectories. \\
\texttt{MultimodalAssociationAnalyzer} & Analyzes spatial relationships among EEG activity, gaze behavior, and regional multimodal responses. \\
\texttt{StatisticalModelAnalyzer} & Performs correlation, group comparison, regression, and physiological response--rating association analysis. \\
\end{longtable}
\endgroup

\subsubsection{Illustrative BrainAgent Execution Traces}
\label{app:brainagent_execution}

As an illustrative BrainAgent execution trace, we present the evaluation of Sleep Assessment instance \texttt{SA-26-Instance1} using Claude Opus 5. The corresponding instruction is:

\begin{quote}
\small\itshape
Load the EEG data file at path \texttt{data/sleep/ISRUC\_01.edf}. Then analyze the following three 1-minute segments: A = minute 274.5 to 275.5, B = minute 63.0 to 64.0, C = minute 437.0 to 438.0. For each segment, quantify the EOG/eye-movement activity strength using the EOG/eye-movement channels present in the current system state (inspect the loaded recording to determine the exact channel names; do not assume names). Compare the three segments and report the segment labels (A, B, C) in descending order of EOG activity strength, including the quantitative metric values used for the ranking. Additionally, determine the sleep stage of the segment with the strongest EOG activity and explicitly state whether that segment is REM sleep, with supporting evidence.
\end{quote}

Listing~\ref{lst:brainagent_trace} presents a faithful condensed trace. It retains the supervisor dispatch, all six planning rounds, all 16 tool calls, and the final analytical output, while omitting repetitive success messages and serialized state snapshots. The colored statements summarize the reasoning that determined each planning segment; color distinguishes rounds only.

\begin{lstlisting}[style=appendixcode,escapeinside={(*@}{@*)},caption={Condensed BrainAgent execution trace for SA-26-Instance1 with Claude Opus 5. Planning summaries are color-coded by round.},label={lst:brainagent_trace}]
[Supervisor dispatch]
Target subagent: SleepAgent
Task: execute SA-26-Instance1 using data/sleep/ISRUC_01.edf

[Planning round 1]
(*@\textcolor{planone}{No recording is registered. Load the file and inspect its channels before selecting EOG inputs.}@*)
  01  RecordingLoader
  02  ChannelInspector

[Planning round 2]
(*@\textcolor{plantwo}{The exact EOG channels are LOC-A2 and ROC-A1. Build the three requested views and quantify EOG activity consistently.}@*)
  03  SignalViewBuilder          [A: 16470--16530 s]
  04  SignalViewBuilder          [B:  3780--3840 s]
  05  SignalViewBuilder          [C: 26220--26280 s]
  06  EOGActivityAnalyzer        [A]
  07  EOGActivityAnalyzer        [B]
  08  EOGActivityAnalyzer        [C]

[Planning round 3]
(*@\textcolor{planthree}{All EOG metrics rank A above B above C. Determine A's stage, collect spectral and EOG evidence, and formalize the ranking.}@*)
  09  SleepStageEstimator        [A, 1-min view]
  10  SleepSpectralSegmentAnalyzer [A]
  11  EOGActivityAnalyzer        [A, 0.3--10 Hz check]
  12  FeatureRankerAggregator    [A, B, C]

[Planning round 4]
(*@\textcolor{planfour}{A is staged as W/N1 and is delta-dominant. Add chin-EMG evidence to test for the muscle atonia expected in REM.}@*)
  13  SignalViewBuilder          [A, X1 channel]
  14  EMGActivityAnalyzer        [A]

[Planning round 5]
(*@\textcolor{planfive}{The 1-min staging context is limited. Re-estimate A within a 10-min EEG/EOG/EMG context before drawing the sleep-stage conclusion.}@*)
  15  SignalViewBuilder          [16200--16800 s context]
  16  SleepStageEstimator        [10-min context]

[Planning round 6]
(*@\textcolor{plansix}{The state now contains convergent EOG, staging, spectral, and EMG evidence. No further tool call is needed; generate the report.}@*)

[Final output]
EOG activity ranking (descending): A > B > C

Segment  Window (s)       EOG RMS (V)  Line length  Derivative RMS
A        16470--16530      4.44e-06     2.16e-07     3.72e-07
B         3780--3840       1.81e-06     1.47e-07     1.97e-07
C        26220--26280      1.10e-06     9.22e-08     1.25e-07

The strongest segment is A, and it is not REM sleep. Its two 30-s epochs were initially 
estimated as W and N1. In the 10-min context, the epochs covering A were N2 and W, whereas 
the REM epochs occurred outside A. The EEG was delta-dominant (66.9% relative delta power), 
and the X1 EMG retained measurable tone (RMS 4.68e-06 V), providing no evidence of REM atonia.
Conclusion: A > B > C; segment A is a wake/light-NREM transition, not REM sleep.
\end{lstlisting}

A second trace illustrates the evaluation of Foundational Analysis instance \texttt{FA-33-Instance15} on a SEED-V recording. The corresponding instruction is:

\begin{quote}
\small\itshape
Load the EEG data file at path \texttt{data/core/SEED-V-05.cnt}. Select the standard EEG channels and compute, for each channel, alpha-band relative power and signal variance. Exclude channels with variance below $Q_1-1.5\times\mathrm{IQR}$ or above $Q_3+1.5\times\mathrm{IQR}$. Among the remaining channels, return those whose alpha relative power is greater than the global mean across the retained channels. Use Welch PSD estimation, with relative power defined as the target-band power divided by the total power from 0.5 to 45 Hz.
\end{quote}

Listing~\ref{lst:brainagent_trace_fa33} retains all six planning rounds, all 12 successful tool calls, and the final analytical output. Serialized state snapshots, repetitive success messages, and the provider-side retry preceding the final planning round are omitted because they do not alter the analytical workflow. Following the benchmark's agent organization, the trace names the executing subagent \texttt{NeurocogAgent}.

\begin{lstlisting}[style=appendixcode,escapeinside={(*@}{@*)},caption={Condensed BrainAgent execution trace for FA-33-Instance15 with Claude Opus 5. Planning summaries are color-coded by round.},label={lst:brainagent_trace_fa33}]
[Supervisor dispatch]
Target subagent: NeurocogAgent
Task: execute FA-33-Instance15 using data/core/SEED-V-05.cnt

[Planning round 1]
(*@\textcolor{planone}{No recording is registered. Load the CNT file and inspect its channels before selecting the EEG inputs.}@*)
  01  RecordingLoader
  02  ChannelInspector

[Planning round 2]
(*@\textcolor{plantwo}{The recording contains 64 EEG channels after excluding VEO and HEO. Compute channel-wise alpha relative power and variance over the same channel set.}@*)
  03  SpectralFeatureAnalyzer   [64 EEG channels; alpha/[0.5--45] Hz]
  04  TimeDomainFeatureAnalyzer [64 EEG channels; variance]

[Planning round 3]
(*@\textcolor{planthree}{Per-channel previews are truncated. Apply IQR-based variance screening and expose the complete variance and alpha-power rankings.}@*)
  05  ArtifactQualityAnalyzer   [variance outliers; 1.5 x IQR]
  06  FeatureRankerAggregator   [variance; ascending]
  07  FeatureRankerAggregator   [alpha relative power; descending]

[Planning round 4]
(*@\textcolor{planfour}{The upper variance tail is still hidden by the state summary. Retrieve a descending variance ranking to identify the high-side outliers.}@*)
  08  FeatureRankerAggregator   [variance; descending top 20]

[Planning round 5]
(*@\textcolor{planfive}{M1 and M2 are the two high-variance outliers. Recompute the retained-channel mean and expose the remaining per-channel alpha values in manageable groups.}@*)
  09  SpectralFeatureAnalyzer   [62 retained channels; global mean]
  10  SpectralFeatureAnalyzer   [retained channel group A]
  11  SpectralFeatureAnalyzer   [retained channel group B]
  12  SpectralFeatureAnalyzer   [retained channel group C]

[Planning round 6]
(*@\textcolor{plansix}{The retained-channel mean is 8.9181\%. Compare every retained channel with this threshold and report the qualifying set.}@*)

[Final output]
Recording: 66 channels at 1000 Hz; 64 channels identified as EEG.
Variance-IQR exclusions: M2 (7.782e-06), M1 (3.405e-07).
Retained EEG channels: 62.
Global mean alpha relative power over retained channels: 8.9181%.

Qualified channels (alpha relative power > 8.9181%; 32 reported):
C3, C1, C2, C4, TP7, CP5, CP3, CP1, CP2, CP4, CP6, TP8,
P7, P5, P3, P1, PZ, P2, P4, P6, P8, PO7, PO5, PO3, POZ,
PO4, PO6, PO8, CB1, O1, OZ, O2.
\end{lstlisting}

\subsection{CodeAct Execution Protocol}
\label{app:codeact_protocol}

\subsubsection{Interaction Loop}
\label{app:codeact_loop}

CodeAct provides the autonomous code-execution counterpart to the structured BrainAgent workflow. It receives the same instance-level instruction and permitted input files through the unified interface described in Section~\ref{app:black_box_evaluation_protocol}, but it is not given the tool inventory, structured intermediate state, or task-specific helper functions. Instead, the target LLM independently selects the analysis method, Python libraries, preprocessing operations, intermediate computations, and artifact-generation procedure needed to complete the instruction.

For each \texttt{instance}, CodeAct initializes an instance-scoped persistent Python kernel. At every interaction round, the model returns one of two actions: an \texttt{<execute>} block containing Python code or a \texttt{<solution>} block containing the final report. Code inside an \texttt{<execute>} block is evaluated in the persistent kernel, allowing loaded recordings, intermediate variables, and generated files to be reused across subsequent rounds of the same instance. Textual stream output, \texttt{text/plain} execution results, and exception tracebacks are collected as an \emph{Observation} and returned to the model. The model may then inspect the result, correct erroneous code, revise its analytical strategy, or continue the analysis. This loop terminates when the model emits a valid \texttt{<solution>} block or when an execution-control limit is reached. Only the content of \texttt{<solution>} is treated as the target system's final response. Intermediate code, printed values, and error messages are retained as execution information but are not interpreted as final answers and do not directly contribute to the benchmark score. The final report and any reported artifacts are subsequently processed by the same external Parser Agent and validation pipeline used for BrainAgent.

\subsubsection{Prompt and Execution Contract}
\label{app:codeact_prompt}

Listing~\ref{lst:codeact_prompt} presents the complete system prompt used by CodeAct. The prompt defines the executable action format, the persistent Python environment, the observation feedback mechanism, and the distinction between intermediate execution and the final response.

\begin{lstlisting}[
	style=appendixcode,
	caption={System prompt used for the CodeAct execution protocol.},
	label={lst:codeact_prompt}
	]
	You are a helpful assistant assigned a problem-solving task. You have access to
	an interactive Python environment to inspect data and calculate the answer.
	
	Return exactly one action block per turn:
	<execute>...</execute> or <solution>...</solution>.
	Do not output plans, explanations, or text outside the selected block.
	
	Then choose exactly one of these actions:
	
	1) Execute Python code by enclosing it in <execute>...</execute>. The code will
	run in a persistent Python kernel and the output will be returned as an
	Observation. Top-level variables from earlier snippets remain available.
	2) When the task is complete, provide the requested final report enclosed in
	<solution>...</solution>. The text inside <solution> is returned to the user,
	so it must follow the task's requested output format and contain all results.
	
	Use code to inspect the provided files instead of guessing. Paths named in the
	task are accessible from the current workspace. To conserve context, never print
	an entire long signal, label sequence, dataframe, or file.
	
	---
	Example task:
	The file input/HR.npy contains a 1 Hz heart-rate signal in BPM. Calculate how
	many seconds are in the inclusive range 60 to 100 BPM and return JSON with the
	key time_in_range.
	
	Assistant:
	<thought>I will load the signal and inspect its shape.</thought>
	<execute>
	import numpy as np
	hr = np.load("input/HR.npy")
	print(hr.shape)
	</execute>
	
	Observation:
	(300,)
	
	Assistant:
	<thought>I will count samples in range; at 1 Hz the count equals seconds.</thought>
	<execute>
	time_in_range = float(np.sum((hr >= 60) & (hr <= 100)))
	print(time_in_range)
	</execute>
	
	Observation:
	240.0
	
	Assistant:
	<thought>The calculation is complete, so I will return the requested JSON.</thought>
	<solution>
	{"time_in_range": 240.0}
	</solution>
	
	---
	The actual task follows in the user message.
\end{lstlisting}

On the final permitted interaction round, the controller retains the preceding system prompt, instance query, execution actions, and observations, and appends an additional user message to force finalization. The appended message is shown verbatim in Listing~\ref{lst:codeact_final_prompt}. 
\begin{lstlisting}[
	style=appendixcode,
	caption={Verbatim finalization prompt appended on the last CodeAct interaction round.},
	label={lst:codeact_final_prompt}
	]
	The execution budget is exhausted. You must return the best available final
	answer now inside <solution>...</solution>. Do not execute more code.
\end{lstlisting}

The instance-specific user message is constructed separately from the system prompt. Before dispatch, the evaluator projects the permitted \texttt{data\_path} and optional \texttt{label\_path} into the isolated execution environment and rewrites them as paths under \texttt{/input/}. The user message then identifies these paths, presents the natural-language \texttt{instruction}, and appends any additional fields explicitly included in \texttt{agent\_input}. The instruction is augmented with a runtime rule stating that input files under \texttt{/input/} are read-only and that requested artifacts must be saved under \texttt{/workspace/file\_check/} or a relative path under \texttt{file\_check/}.

The benchmark information boundary is enforced by the evaluator rather than encoded as additional task text in the system prompt. CodeAct receives only the prepared \texttt{agent\_input} and the permitted input-file mounts. The ground truth, Parser Prompt, {Validation Configuration}, metric definitions, metric weights, and evaluator-side scores are not included in the model messages or execution workspace. The model also has no interactive mechanism for requesting additional files during execution. After CodeAct returns its final \texttt{<solution>}, the external Parser Agent extracts the required fields, and the evaluator applies the hidden validation units and scoring configuration.

Although the worked example includes \texttt{<thought>} blocks, the executable protocol recognizes only \texttt{<execute>} and \texttt{<solution>} as actions. Text outside these two blocks is neither executed nor treated as the final answer. Accordingly, intermediate Python output and observations are used only to support subsequent interaction rounds, whereas the content enclosed by \texttt{<solution>} constitutes the final response submitted to the evaluation pipeline.

\subsubsection{Execution Controls}
\label{app:codeact_controls}
Table~\ref{tab:codeact_controls} reports the CodeAct-specific interaction controls. Execution output is returned to the model as a merged textual observation. This observation may contain standard-stream text, expression results, textual display representations, or a Python traceback. When an execution succeeds without textual output, the runtime returns an explicit success message; when it exceeds the per-execution limit, a timeout observation is returned. These observations provide the model with an opportunity to diagnose and revise failed analyses, while the repeated-failure controls prevent unproductive execution loops. On the final permitted round, CodeAct is instructed to stop executing code and return the best available result inside \texttt{<solution>}. If the hard instance deadline or an unrecoverable infrastructure error occurs earlier, the instance terminates with a structured error rather than receiving evaluator feedback or a reference answer.
\begin{table}[!ht]
	\centering
	\small
	\setlength{\tabcolsep}{4pt}
	\renewcommand{\arraystretch}{1.12}
	\caption{CodeAct interaction and execution controls.}
	\label{tab:codeact_controls}
	\begin{tabularx}{\linewidth}{@{}>{\raggedright\arraybackslash}p{0.27\linewidth}>{\raggedright\arraybackslash}p{0.25\linewidth}X@{}}
		\toprule
		\rowcolor{black!7}
		\textbf{Control dimension} & \textbf{Budget} & \textbf{Operational rule} \\
		\midrule
		Interaction-round budget & {20 rounds} & Rounds 1--19 permit \texttt{<execute>}; round 20 is reserved for \texttt{<solution>}. \\
		Execution-time budget & {150 s / 720 s} & Per Python execution / complete instance. \\
		Response-length budget & {2,048 / 1,024 tokens} & Maximum length of non-final / final responses. \\
		Observation history & {8K chars / 3 pairs} & Per-observation limit / recent execution--observation pairs retained verbatim. \\
		Stall termination & {3 consecutive events} & Triggered by missing actions, identical code, or the same execution-error signature. \\
		Decoding configuration & {Thinking off / $T=0.7$} & Temperature is applied only when supported by the endpoint. \\
		\bottomrule
	\end{tabularx}
\end{table}

The fixed Python environment contains general numerical and scientific-computing packages, including NumPy, SciPy, pandas, Matplotlib, seaborn, and scikit-learn, together with EEG- and physiological-signal packages such as MNE, YASA, pyEDFlib, edfio, and WFDB. CodeAct may compose these libraries freely but cannot assume access to uninstalled task-specific software. Variables and temporary files persist across interaction rounds only within the current instance; different instances receive independent kernels, message histories, and workspaces. Requested artifacts remain available for evaluator-side validation before cleanup, while the retained audit record stores action types, token usage, elapsed time, code hashes, observation summaries, and termination status. CPU, memory, image, filesystem, and network isolation are described separately in Appendix \ref{app:runtime_failure_handling}.

\subsection{Containerized Runtime and Failure Handling}
\label{app:runtime_failure_handling}

This section specifies the containerized execution environment used by BrainAgent and CodeAct and defines how infrastructure-induced interruptions are distinguished from target-system failures. Both paradigms operate through the same instance-level input--output interface, while their container configurations differ where required by their execution mechanisms.

\subsubsection{Container Configuration}
\label{app:container_configuration}

Each \texttt{instance} is executed in a fresh Docker container with an independent runtime state. Both paradigms use a versioned Python 3.9 scientific-computing environment and receive the same instruction and permitted input files. The containerization layer does not alter the expected report or artifact requirements of the instance. Table~\ref{tab:container_configuration} summarizes the principal runtime settings.

\begin{table*}[!ht]
	\centering
	\small
	\setlength{\tabcolsep}{6pt}
	\renewcommand{\arraystretch}{1.12}
	\caption{Principal container configurations used for BrainAgent and CodeAct.}
	\label{tab:container_configuration}
	\begin{tabularx}{\textwidth}{@{}>{\raggedright\arraybackslash}p{0.22\textwidth}>{\raggedright\arraybackslash}X>{\raggedright\arraybackslash}X@{}}
		\toprule
		\rowcolor{black!7}
		\textbf{Configuration} & \textbf{BrainAgent} & \textbf{CodeAct} \\
		\midrule
		Runtime image & Versioned Python 3.9 scientific-computing image & Versioned Python 3.9 scientific-computing image \\
		Execution backend & Supervisor--subagent workflow with controlled analytical tools & Persistent IPython kernel for model-generated Python code \\
		CPU allocation & 4 CPU cores per instance & 4 CPU cores per instance \\
		Memory limit & 32 GB per instance & 8 GB per instance \\
		Network access & Enabled because target-model requests originate inside the BrainAgent container & Disabled for generated code; target-model requests are issued outside the execution container \\
		Total instance deadline & 720 seconds & 720 seconds \\
		Per-execution limit & Not separately constrained beyond the instance deadline & 150 seconds for each Python execution \\
		\bottomrule
	\end{tabularx}
\end{table*}

The larger memory allocation for BrainAgent accommodates its multi-agent runtime, hierarchical intermediate state, and domain-oriented analytical toolset. CodeAct instead executes generated code in a smaller scientific-computing sandbox. Its execution container has no network access and receives no target-model credentials, whereas BrainAgent requires network access because model requests and tool-mediated planning are performed inside the container. Within each execution paradigm, the corresponding resource and time constraints are held fixed across evaluated models.

\subsubsection{Failure Classification and Retry Policy}
\label{app:failure_retry_policy}
We distinguish verified infrastructure interruptions from failures attributable to the target system. Transient API or container-runtime failures are retried under a predefined policy; if recovery is unsuccessful, the affected instance is rerun after the external condition is restored using the same instruction, inputs, model configuration, execution paradigm, and container constraints. The interrupted execution is excluded from score aggregation and replaced by the valid rerun. In contrast, failures arising from model planning, code generation, tool use, execution strategy, or incomplete outputs are retained as evaluation outcomes and are not rerun by the evaluator. Any self-correction performed by BrainAgent or CodeAct within their allotted execution budgets is considered part of the evaluated paradigm; once that budget is exhausted, the resulting report, missing output, or termination state is scored as produced. Table~\ref{tab:failure_retry_policy} summarizes the resulting policy described above.

\begin{table*}[!ht]
	\centering
	\small
	\setlength{\tabcolsep}{6pt}
	\renewcommand{\arraystretch}{1.12}
	\caption{Failure classification and instance-level rerun policy.}
	\label{tab:failure_retry_policy}
	
	\begingroup
	\renewcommand{\tabularxcolumn}[1]{m{#1}}
	\begin{tabularx}{\textwidth}{@{}
			>{\raggedright\arraybackslash}m{0.25\textwidth}
			>{\raggedright\arraybackslash}X
			>{\centering\arraybackslash}m{0.10\textwidth}
			>{\centering\arraybackslash}m{0.17\textwidth}
			@{}}
		\toprule
		\rowcolor{black!7}
		\textbf{Failure type} & \textbf{Representative examples} & \textbf{Rerun} & \textbf{Model outcome} \\
		\midrule
		Provider-side communication failure & Connection interruption, request timeout, rate limiting, or HTTP 5xx response & \yesmark & \nomark \\
		\midrule
		Provider access or service-state failure & Verified HTTP 400/401/402/403/404 caused by access, quota, routing, endpoint, or content-inspection restrictions & \yesmark & \nomark \\
		\midrule
		Container initialization failure & Docker image startup failure or Python-kernel initialization failure & \yesmark & \nomark \\
		\midrule
		Invalid generated code & Syntax errors, runtime exceptions, or invalid data-processing operations in CodeAct & \nomark & \yesmark \\
		\midrule
		Incorrect tool invocation & Inappropriate tool selection, invalid arguments, or unresolved tool dependencies in BrainAgent & \nomark & \yesmark \\
		\midrule
		Execution-budget exhaustion & Excessive computation, hard instance timeout, maximum interaction rounds, or repeated execution failures & \nomark & \yesmark \\
		\midrule
		Incomplete report or artifact & Missing final report, omitted requested result, absent artifact, or malformed deliverable & \nomark & \yesmark \\
		\bottomrule
	\end{tabularx}
	\endgroup
\end{table*}

\clearpage
\section{Additional Experimental Results and Analyses}
\label{app:more_results}

\subsection{Overall Score Distributions}
\label{app:overall_score_distributions}
\begin{figure}[!th]
	\centering
	\includegraphics[width=\columnwidth]{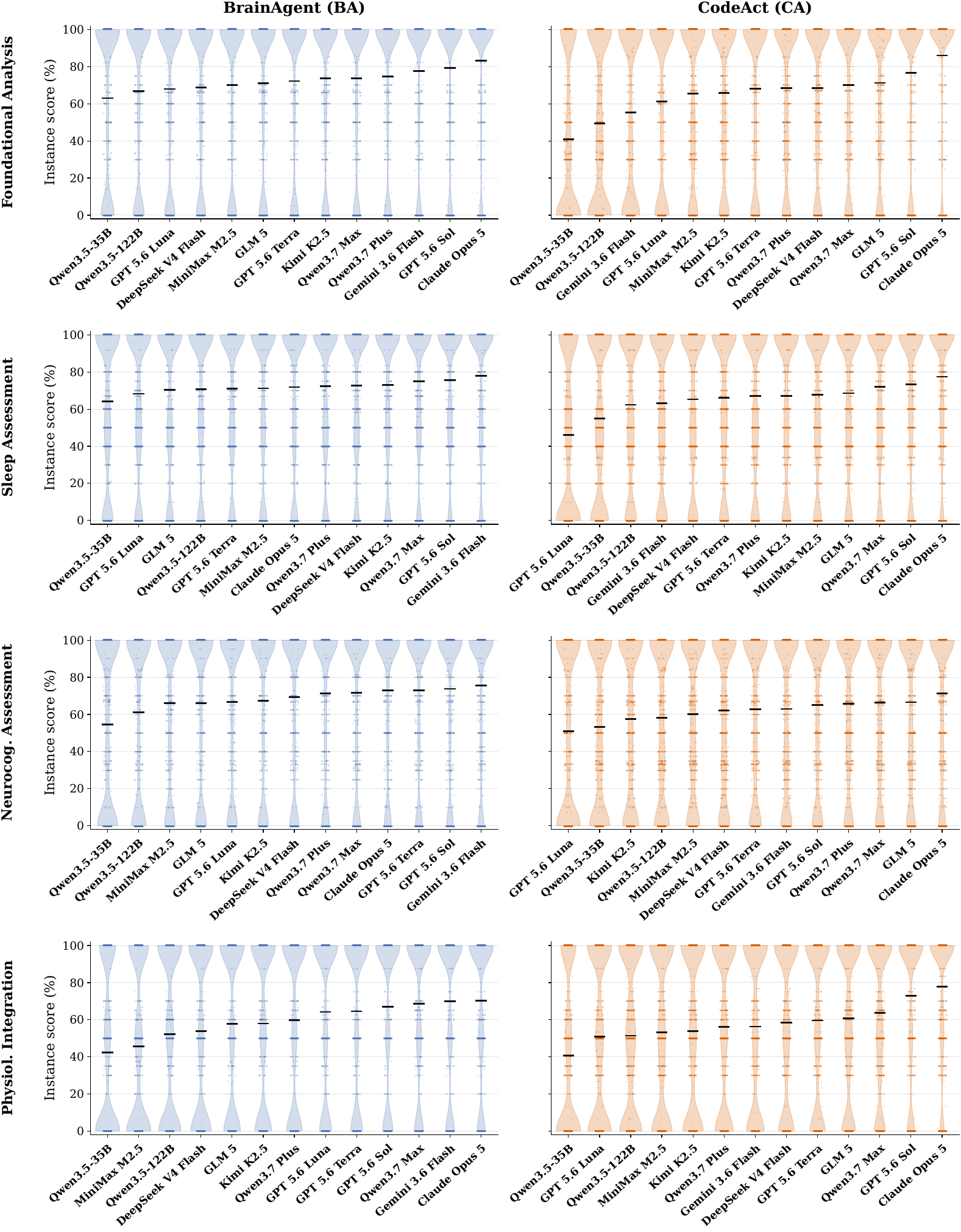}
	\caption{\textbf{Instance-level score distributions across models and execution paradigms.} Violin densities and translucent points show the unweighted normalized instance scores across four subsets. Black horizontal markers denote arithmetic means. Models are sorted independently within each panel by their mean instance score; task-difficulty weights are not applied.}
	\label{fig:model_score_distributions}
\end{figure}
Figure~\ref{fig:model_score_distributions} reveals that aggregate performance masks substantial heterogeneity at the instance level across all four \benchmarkname{} subsets. Both execution paradigms exhibit broad score distributions, indicating that strong average performance does not translate into uniformly reliable behavior across individual analyses. Compared with flexible LLM-driven coding, structured agentic execution generally raises the performance of weaker models and reduces between-model dispersion, making overall performance less sensitive to the choice of underlying LLM. This effect is particularly evident in domains where execution can be effectively organized through predefined analytical capabilities. CodeAct, in contrast, shows stronger model dependence: weaker models degrade more substantially, while stronger models can match or surpass the leading structured-agent configurations in some domains. Overall, structured workflows primarily improve robustness across model capabilities, whereas autonomous coding preserves greater performance variability together with a higher potential ceiling for sufficiently capable models.
\subsection{Difficulty-Conditioned Effects of Execution Paradigms}
\label{app:difficulty_conditioned_effects}
\begin{figure}[!ht]
	\centering
	\includegraphics[width=\columnwidth]{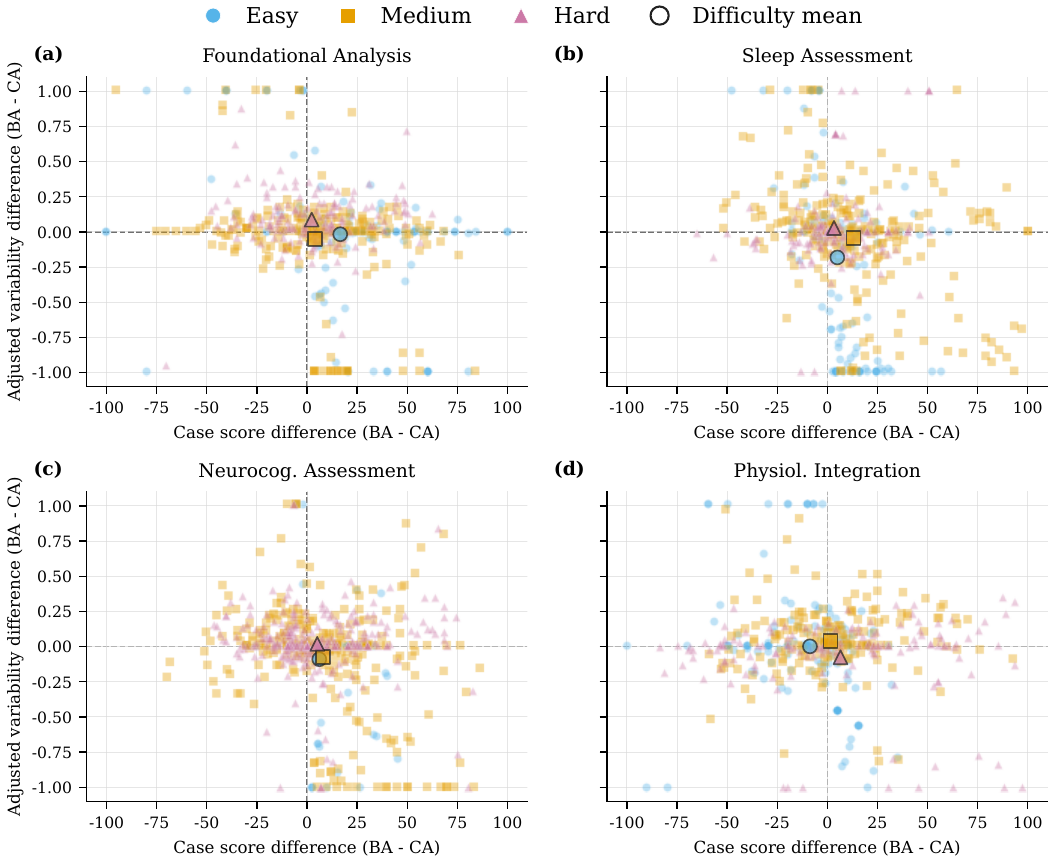}
	\caption{\textbf{Difficulty-conditioned performance--stability trade-off between BrainAgent (BA) and CodeAct (CA).} Results are shown across the four \benchmarkname{} subsets. Each translucent marker represents a paired model--task comparison, with color and shape denoting task difficulty; outlined markers indicate difficulty-level means. Positive horizontal values indicate higher task scores under BA, while negative vertical values indicate lower adjusted within-task variability under BA.}
	\label{fig:difficulty_performance_stability_tradeoff}
\end{figure}
To quantify how the relative benefit of structured execution changes with difficulty, let $\bar{S}*{p,d}$ denote the task-macro score under paradigm $p\in{\mathrm{BA},\mathrm{CA}}$ at difficulty $d\in{\mathrm{E},\mathrm{M},\mathrm{H}}$, and define the paradigm gap as $G_d=\bar{S}*{\mathrm{BA},d}-\bar{S}*{\mathrm{CA},d}$. The \textbf{Difficulty Advantage Index} (DAI) is
\begin{equation}
	\mathrm{DAI}=\frac{(G*{\mathrm{M}}-G_{\mathrm{E}})+(G_{\mathrm{H}}-G_{\mathrm{M}})}{2}
	=\frac{G_{\mathrm{H}}-G_{\mathrm{E}}}{2}.
\end{equation}
DAI characterizes how the relative advantage of BrainAgent over CodeAct evolves with task difficulty, with positive and negative values indicating widening and narrowing advantages, respectively.

Figure~\ref{fig:difficulty_performance_stability_tradeoff} further examines how this difficulty-conditioned performance effect relates to within-task stability. Across the four subsets, the difficulty-level means reveal no universal trajectory for structured execution. In Foundational Analysis, the BrainAgent advantage weakens as difficulty increases, whereas Sleep Assessment shows its clearest benefit at intermediate difficulty and substantially less separation on the hardest tasks. Neurocognitive Assessment exhibits a comparatively stable advantage across difficulty levels, accompanied by generally favorable stability. In contrast, Physiological Integration displays an increasing benefit from structured execution as task difficulty rises, suggesting that explicit workflow organization becomes more valuable as multimodal coordination demands grow. The broad dispersion around these aggregate trends further highlights substantial heterogeneity across individual model--task combinations. Taken together, these results show that task difficulty does more than reduce overall performance: it changes how execution strategy affects both analytical effectiveness and stability. The benefit of structured agentic workflows is therefore not monotonic with difficulty, but depends on the nature of the analytical demands. \textbf{Structured execution can provide substantial support when complexity can be organized through domain-oriented workflows, whereas its advantage may diminish when the dominant bottleneck shifts toward long-horizon reasoning and evidence integration.}

\subsection{Additional Validation-Unit-Level Analysis}\label{app:addtional_validation_unit}
\begin{figure}[!ht]
	\centering
	\includegraphics[width=\columnwidth]{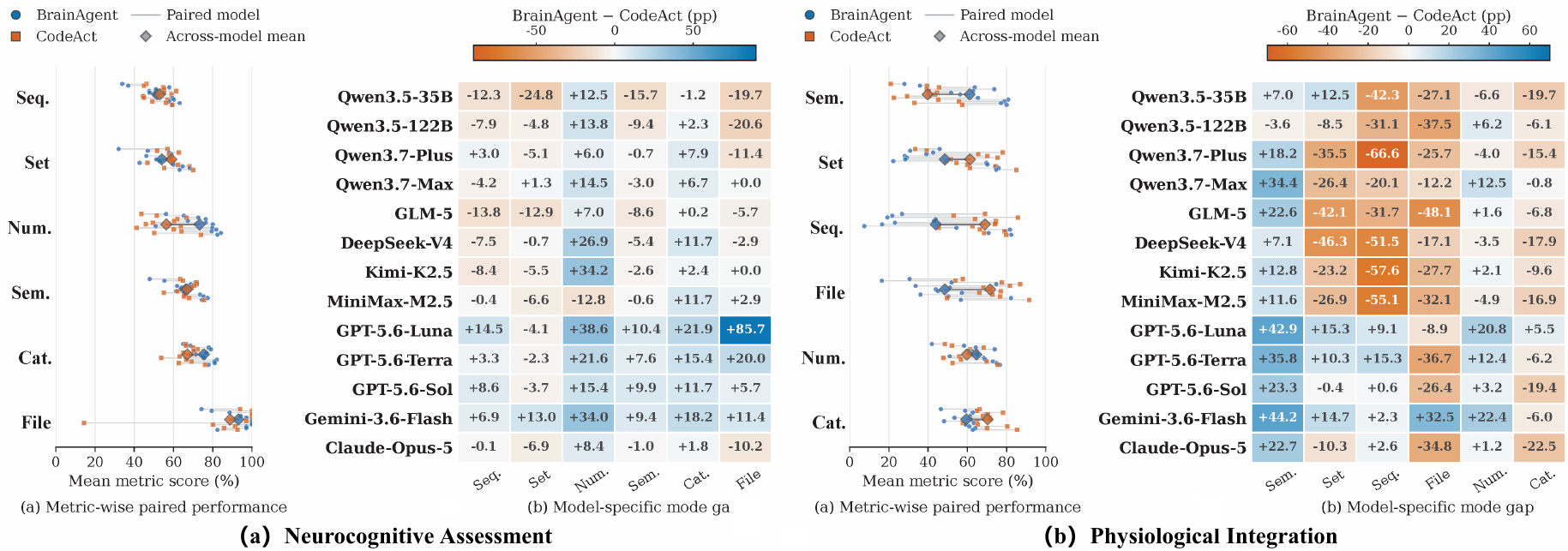}
	\caption{\textbf{Validation-unit-specific effects of execution paradigms.} Within each subset, paired markers show model-wise unweighted mean validation scores under BA and CA, diamonds indicate across-model means, and heatmaps report the corresponding \(\Delta\) in percentage points.)}
	\label{fig:validation_unit_addtitonal}
\end{figure}
Figure~\ref{fig:validation_unit_addtitonal} extends the validation-unit analysis to Neurocognitive Assessment and Physiological Integration. The overall pattern is consistent with the \textbf{\textit{Finding 3}}: BrainAgent retains a clear and broadly consistent advantage in numerical validation across models, further supporting the finding that structured agentic execution most reliably strengthens quantitative EEG analysis. In contrast, the remaining validation units exhibit substantially more heterogeneous paradigm effects, with the relative benefit of BrainAgent and CodeAct varying across models and analytical domains. This heterogeneity is particularly evident in semantic, categorical, set, sequence, and artifact validation, where neither execution paradigm shows a uniform advantage across the two subsets. Taken together with the Foundational Analysis and Sleep Assessment results, these findings indicate that the benefit of structured execution is selective rather than universal, with quantitative analysis emerging as its most consistent cross-domain strength.
\subsection{Additional Within-Task Stability Analysis}\label{app:E4}
\begin{figure}[!ht]
	\centering
	\includegraphics[width=\columnwidth]{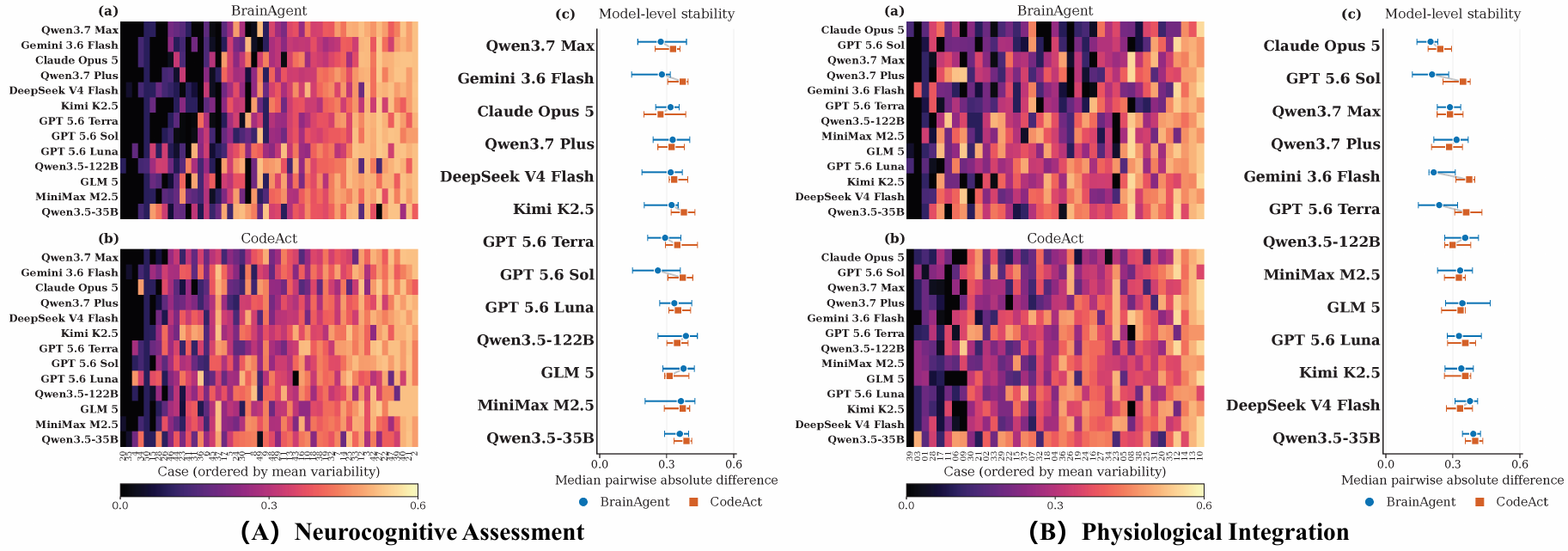}
	\caption{\textbf{Cross-instance stability within reusable tasks.} For each model--task pair, the heatmap shows the mean pairwise absolute difference among within-task normalized instance scores. Model-level panels report the median across tasks with task-bootstrap 95\% confidence intervals.}
	\label{fig:inner_stability_addtitonal}
\end{figure}
For a given model, execution paradigm, and task, let $s_1,\ldots,s_n\in[0,100]$ denote the normalized scores of its $n$ instances. We quantify within-task variability using the \emph{mean pairwise absolute difference} (MPAD):
\begin{equation}
	\mathrm{MPAD}
	=
	\frac{2}{n(n-1)}
	\sum_{i<j}
	\frac{|s_i-s_j|}{100}.
\end{equation}
MPAD ranges from 0 to 1, with lower values indicating more consistent performance across instances sharing the same analytical objective and higher values indicating greater within-task variability. Because MPAD measures consistency rather than correctness, it should be interpreted together with task-level performance. For model-level reporting, we take the median MPAD across tasks and estimate 95\% confidence intervals using 10,000 task-level bootstrap resamples.

Figure~\ref{fig:inner_stability_addtitonal} extends the within-task stability analysis to Neurocognitive Assessment and Physiological Integration. Consistent with the Foundational Analysis and Sleep Assessment results in the Sec. \ref{sec4.4}, structured agentic execution generally exhibits lower within-task variability than flexible LLM-driven coding across both additional subsets. Although individual model--task pairs remain heterogeneous and occasional reversals occur, the model-level distributions show an overall shift toward lower MPAD under BrainAgent. Importantly, this pattern persists across substantially different analytical settings, from neurocognitive inference to multimodal physiological analysis, suggesting that the stability benefit of structured workflows is not confined to a particular EEG domain. Taken together with the \textbf{\textit{Finding 4}}, these findings further support that structured agentic workflows improve the cross-context reliability of EEG understanding by promoting more consistent execution across recordings that share the same analytical objective.

\subsection{Effects of Reasoning Effort and Token Efficiency}\label{app:E5}
\begin{figure}[!ht]
	\centering
	\includegraphics[width=\columnwidth]{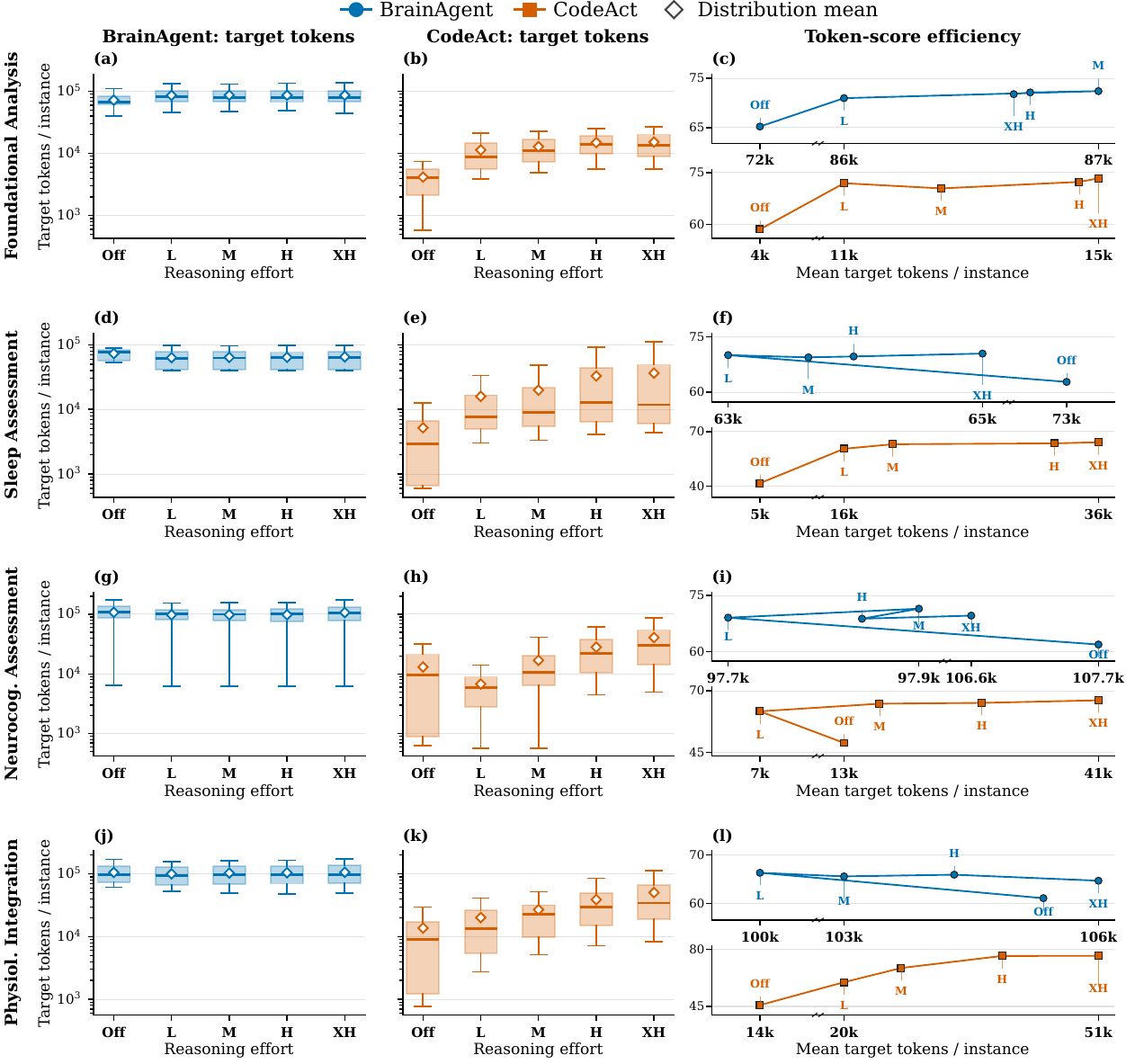}
	\caption{\textbf{Reasoning-effort scaling and token--score efficiency across execution paradigms.} Results are shown for all four \benchmarkname{} subsets. The left and middle columns report the distributions of target-token usage per instance across reasoning-effort levels for BrainAgent and CodeAct, respectively, with diamonds indicating distribution means. The right column relates mean target-token usage to the corresponding EEG understanding score, illustrating how additional inference-time computation translates into performance under each execution paradigm.}
	\label{fig:reasoning_addtitonal}
\end{figure}

Figure~\ref{fig:reasoning_addtitonal} further examines how increasing reasoning effort affects test-time computation across all four \benchmarkname{} subsets. BrainAgent and CodeAct exhibit markedly different token-scaling behaviors: under BrainAgent, target-token usage remains comparatively stable across reasoning-effort levels, whereas CodeAct shows a substantially stronger increase in token consumption as reasoning effort grows. This difference reflects how the two execution paradigms organize additional inference-time computation, with structured agentic workflows constraining reasoning within a relatively stable execution process and flexible coding allowing additional reasoning budget to expand autonomous reasoning and code-generation trajectories more freely. When considered together with the corresponding performance trends, \textbf{enabling reasoning provides the dominant performance gain, while further scaling generally requires greater computation for smaller and increasingly task-dependent improvements.} This effect is particularly evident under CodeAct, where higher reasoning effort can continue to improve performance in several settings but is accompanied by pronounced growth in token usage. Reasoning efficiency is therefore strongly execution-dependent: structured workflows stabilize computational demand, whereas flexible coding exposes a clearer trade-off between reasoning budget and analytical performance. These results provide a computational perspective on \textbf{\textit{Finding 5}}, further supporting that \textbf{LLM reasoning is an effective but bounded amplifier of EEG understanding rather than a capability that improves proportionally with inference-time computation.}
\subsection{Model-Family Structure in Task-Level Performance}
\label{app:within_family_consistency}
\begin{figure}[!th]
	\centering
	\includegraphics[width=\columnwidth]{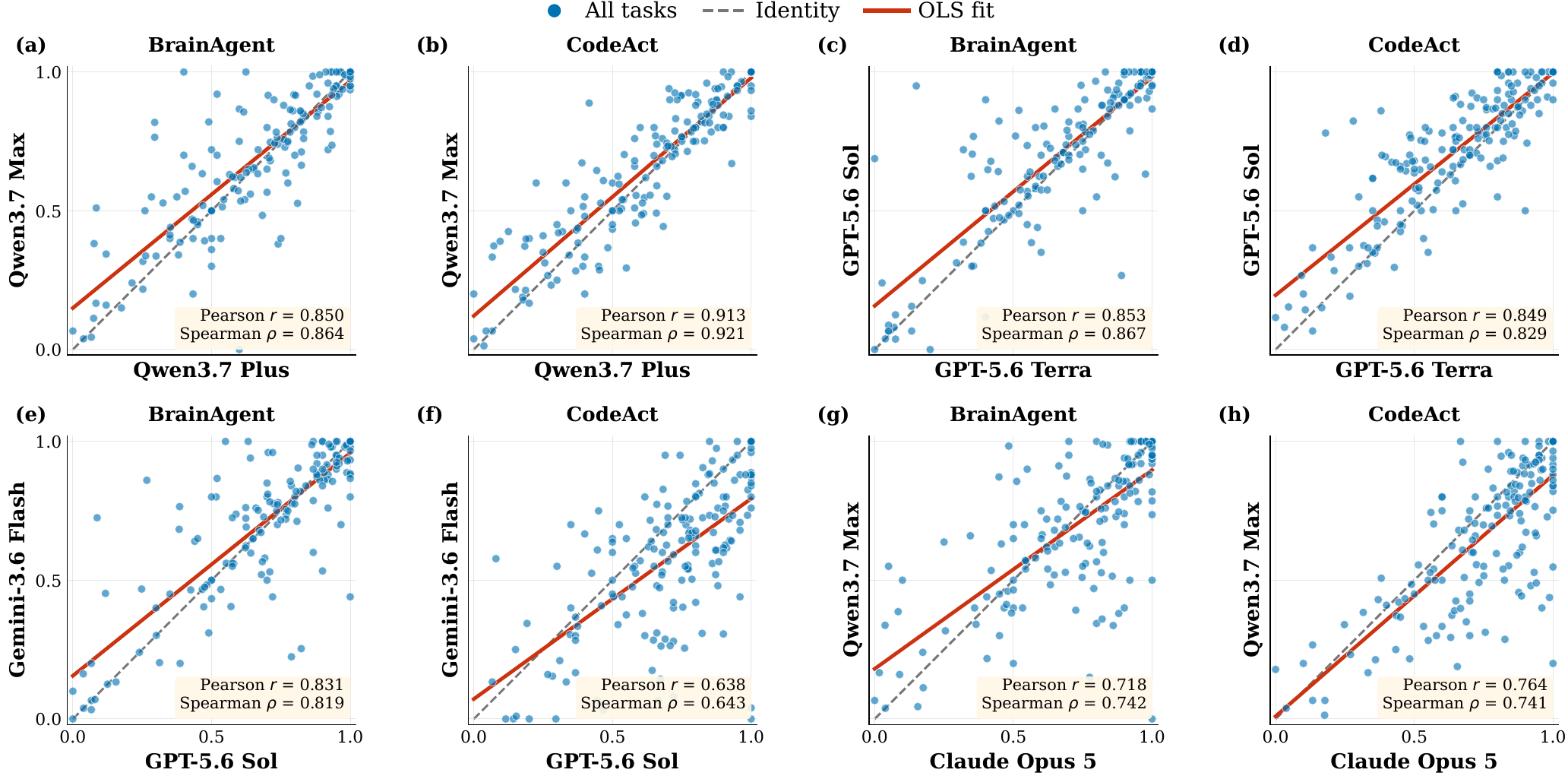}
	\caption{\textbf{Model-family consistency of task-level EEG understanding.} The top row evaluates consistency between model variants from the same family, whereas the bottom row compares strong models from different families. Each point represents the unweighted mean instance score of the same task for a pair of models. The gray dashed line denotes identity, and the red line shows the ordinary least-squares fit. Insets report Pearson and Spearman correlations computed over the 172 pooled tasks. Panels (a)--(d) therefore characterize within-family consistency, while panels (e)--(h) provide a cross-family reference for comparison.}
	\label{fig:model_family}
\end{figure}
Figure~\ref{fig:model_family} examines the consistency of task-level performance profiles both within and across model families. Across both BrainAgent and CodeAct, variants from the same model family exhibit consistently strong agreement, indicating that tasks that are relatively easy or difficult for one family member tend to retain similar relative difficulty for another. Cross-family comparisons remain positively correlated but show noticeably weaker and more variable agreement, suggesting that different model families share a common task structure while exhibiting more distinct capability profiles. This contrast persists across execution paradigms, although the strength of agreement can vary with how the model is executed. Overall, task-level EEG understanding exhibits a clear model-family structure: \textbf{capability profiles are more strongly preserved within families than across families, while execution paradigms further modulate how these underlying capabilities are expressed.} At the same time, deviations from the identity line indicate that stronger models do not improve uniformly across all tasks, reinforcing that model scaling changes the magnitude of performance without simply preserving a fixed task-wise advantage.
\subsection{Consistency of Results Across Repeated Runs}
\begin{figure}[!ht]
	\centering
	\includegraphics[width=\columnwidth]{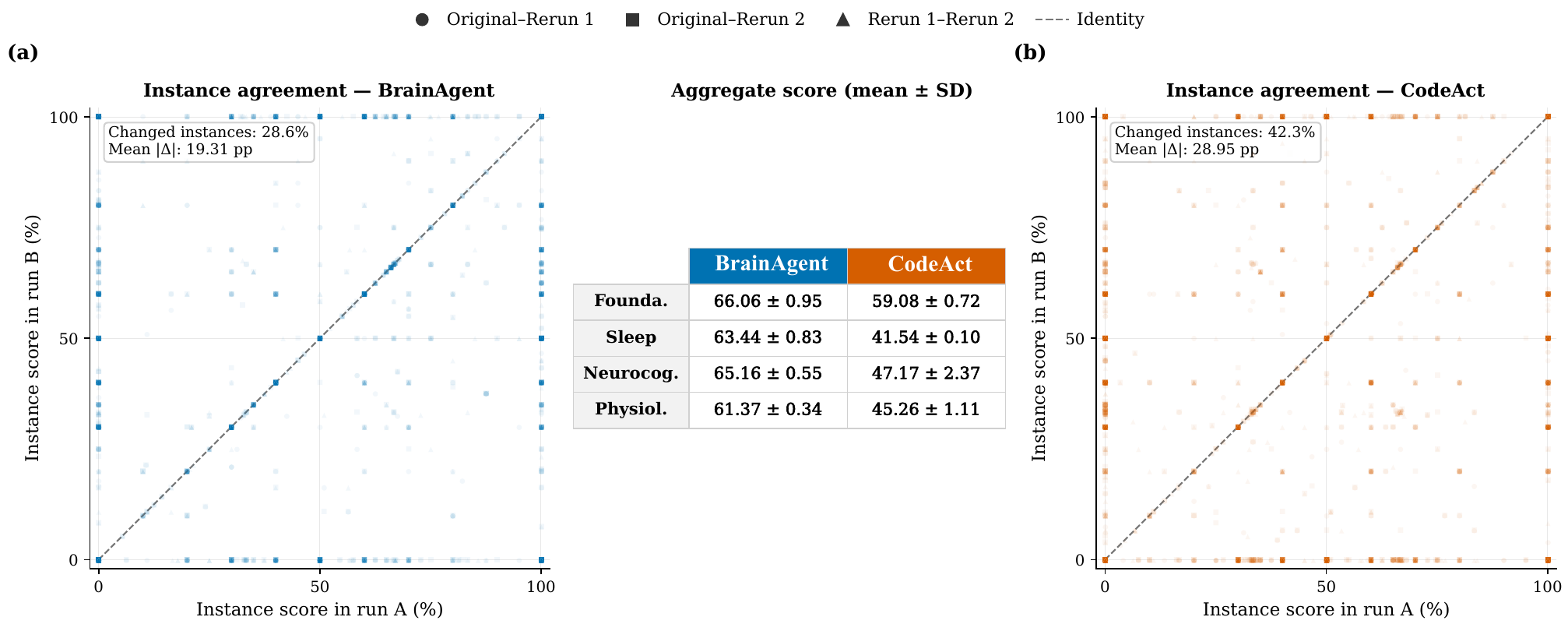}
	\caption{\textbf{Consistency across repeated runs.} Pairwise instance-level agreement across three runs under (a) BrainAgent and (b) CodeAct. The central table reports aggregate benchmark scores as mean ± SD across the three runs.}
	\label{fig:select_rerun}
\end{figure}
Repeated runs produced observable instance-level variation, but BrainAgent showed greater consistency than CodeAct, with fewer changed instances (28.6\% vs. 42.3\%) and a lower mean absolute score difference (19.31 vs. 28.95 percentage points). Nevertheless, aggregate benchmark scores remained closely clustered across the three runs, with standard deviations of 0.10–2.37 points across subsets and execution modes. These results indicate that aggregation over diverse benchmark instances attenuates run-level sampling variability, while BrainAgent’s constrained workflow further improves instance-level reproducibility.
\subsection{Token Usage and Execution Efficiency}
\label{app:token_usage_efficiency}

\begin{figure}[!ht]
	\centering
	\includegraphics[width=\columnwidth]{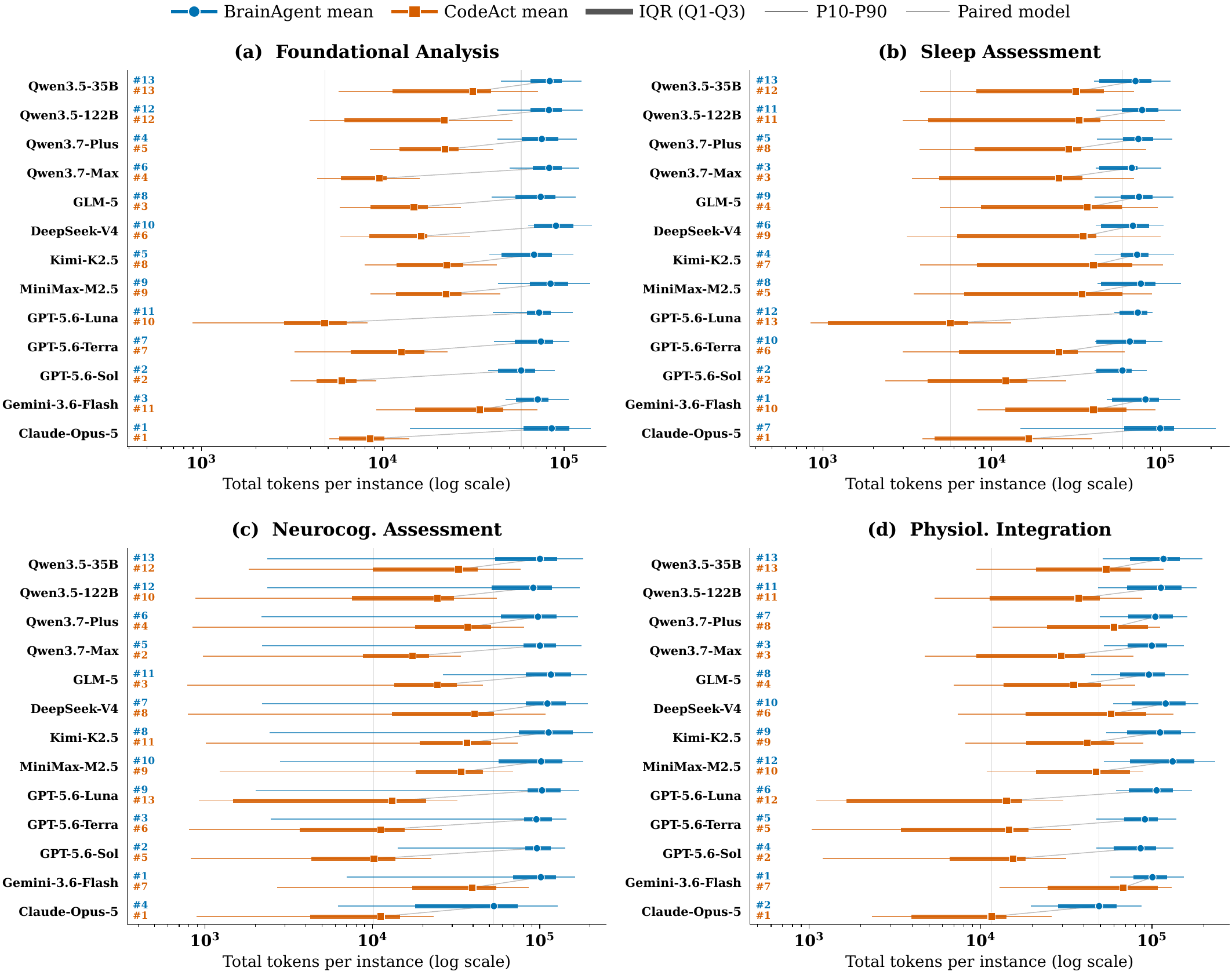}
	\caption{\textbf{Token usage and execution efficiency.} Total tokens per instance are compared between BrainAgent and CodeAct across four subsets. Markers denote means, thick bars show the interquartile range, and thin whiskers indicate the 10th--90th percentiles. Gray lines connect the two execution paradigms for the same model, and the colored annotations report the corresponding performance ranks. The horizontal axis uses a logarithmic scale.}
	\label{fig:token_usage_efficiency}
\end{figure}

Figure~\ref{fig:token_usage_efficiency} compares total token consumption under the two execution paradigms. Across both subsets and all evaluated models, BrainAgent consistently uses more tokens than CodeAct, reflecting the additional communication and context required for agent coordination, tool selection, and intermediate result synthesis. CodeAct is substantially more token-efficient because much of the analytical computation is delegated directly to executable code. Nevertheless, token consumption is not monotonically associated with benchmark performance: several highly ranked CodeAct configurations achieve strong scores with comparatively modest token budgets, whereas larger token usage does not necessarily yield a higher rank. These results reveal a clear performance--efficiency trade-off between the two paradigms: structured agentic execution generally provides stronger and more reliable EEG analysis at higher token overhead, while autonomous coding reduces token consumption but exhibits less consistent performance. As token accounting and pricing differ across model providers, the comparison should be interpreted as execution overhead rather than a direct estimate of monetary cost.

\clearpage

\subsection{Parser Stability and Scoring Robustness}
\label{app:parser_scoring_robustness}

The Parser Agent separates free-form reporting from structured validation, but extraction errors could introduce evaluator-side noise into the final scores. We therefore audit its outputs on the complete Sleep Assessment subset across ten evaluation runs. For each of the 1,025 instances in a run, an independent auditor compares the extracted JSON with the parser prompt and original report, judging only whether the extraction faithfully represents the explicitly reported answer without assessing its scientific correctness.

\begin{table}[!ht]
	\centering
	\small
	\setlength{\tabcolsep}{10pt}
	\renewcommand{\arraystretch}{1.10}
	\caption{Parser-audit results on Sleep Assessment across ten complete evaluation runs. Agree and Agreement are reported as mean $\pm$ standard deviation across runs.}
	\label{tab:parser_audit}
	\begin{tabular}{@{}ccc@{}}
		\toprule
		\textbf{Audited} & \textbf{Agree} & \textbf{Agreement (\%)} \\
		\midrule
		1,025  & $1{,}015.8 \pm 8.9$ & $99.10 \pm 0.87$ \\
		\bottomrule
	\end{tabular}
\end{table}

Across 10,250 audited extractions, 10,158 agree with the independent audit, yielding an aggregate agreement of 99.10\%. The consistently high run-level agreement indicates that converting flexible reports into structured fields contributes limited evaluator-side noise in the Sleep Assessment results and supports the separation between EEG understanding and rigid format compliance. As this audit is restricted to one subset, it should be interpreted as a focused robustness check rather than a benchmark-wide estimate of parser accuracy.

\newpage

\subsection{Representative Failure Cases Across \benchmarkname{}}
To complement the aggregate evaluation, we present representative failure cases across the four \benchmarkname{} subsets under both BrainAgent and CodeAct. All cases use Claude Opus 5 to control for differences in underlying model capability, allowing the analysis to focus on how analytical demands and execution paradigms shape failure behavior. Rather than treating failures as isolated scoring errors, we examine where the analytical pipeline breaks and how the error propagates to the final interpretation. Errors are highlighted in bold red within each listing for clarity.

\paragraph{BrainAgent failure case1.}

This trace illustrates the evaluation of Physiological Integration instance \texttt{PI-27-Instance26} on a SEED-VII recording. The corresponding instruction is:

\begin{quote}
	\small\itshape
	Analyze the third chronologically ordered event\_code=1--event\_code=2 interval.
	Use the following EEG ROIs:
	\{
	``frontal'': [``FP1'', ``FP2'', ``F3'', ``F4''],
	``posterior'': [``P3'', ``P4'', ``O1'', ``O2'']
	\}.
	Compare mean alpha power in [8, 13) Hz between the two ROIs.
	Report the ROI with lower alpha power as lower\_alpha\_roi.
	Report the proportion of finite gaze samples in the central 50\% $\times$ 50\% screen region as central\_gaze\_fraction.
	Briefly state which ROI has lower alpha power, whether gaze is centrally concentrated or relatively peripheral using the 0.5 threshold, and describe the two findings together as a descriptive EEG--gaze spatial profile.
	Do not interpret lower alpha as a direct measure of engagement or imply causal EEG--gaze coupling.
	Report only lower\_alpha\_roi and central\_gaze\_fraction, together with the brief descriptive interpretation.
\end{quote}

\begin{lstlisting}[style=appendixcode,escapeinside={(*@}{@*)},caption={Condensed BrainAgent execution trace for PI-27-Instance26 with Claude Opus 5. Planning summaries are color-coded by round.},label={lst:brainagent_case27_26}]
	[Supervisor dispatch]
	Target subagent: MultiAgent
	Task: analyze the third event-code 1--2 interval using
	SEEDVII EEG and eye-tracking recordings
	
	[Planning round 1]
	(*@\textcolor{planone}{No data are registered. Load the EEG, eye-tracking recording, and event annotations while preserving their native sampling rates. The initial eye-tracking modality hint is rejected, so retry the EDF without that hint.}@*)
	01  MultirateRecordingLoader  [EEG CNT, 1000 Hz]
	02  MultirateRecordingLoader  [eye-tracking EDF, 250 Hz]
	03  NumericAnnotationLoader   [40 event rows]
	
	[Planning round 2]
	(*@\textcolor{plantwo}{The event table contains alternating codes 1 and 2. Resolve the intervals and select the third interval, [483.170, 672.749) s. Compute alpha power for both EEG ROIs and obtain an initial gaze summary.}@*)
	04  EventTrialResolver        [codes 1--2, third interval]
	05  EEGPatternAnalyzer        [alpha 8--13 Hz, frontal/posterior]
	06  EyeTrackingAnalyzer       [GazePointX, GazePointY]
	
	[Planning round 3]
	(*@\textcolor{planthree}{The EEG result is frontal=-25.2661 and posterior=-26.0357, so posterior has lower alpha. The initial gaze call returned validity and dispersion features rather than central fraction; request central\_fraction explicitly.}@*)
	07  EyeTrackingAnalyzer       [feature=central_fraction, fraction=0.5]
	
	[Planning round 4]
	(*@\textcolor{planfour}{The second gaze result contains valid\_ratio=18447/47395=0.3892. Interpret this value as central\_gaze\_fraction; because it is below 0.5, classify gaze as relatively peripheral and generate the report.} \textcolor{red}{\textbf{ERROR: valid\_ratio is not central\_gaze\_fraction; usable=false and insufficient\_valid\_eye\_samples are ignored.}}@*)
	
	[Final output]
	lower_alpha_roi: posterior
	central_gaze_fraction: 0.3892
	
	The posterior ROI exhibits lower mean alpha power (-26.04 dB) than
	the frontal ROI (-25.27 dB). With a central gaze fraction of 0.39,
	gaze is relatively peripheral rather than centrally concentrated.
	Together, these findings provide a descriptive EEG--gaze spatial
	profile without implying engagement or causal EEG--gaze coupling.
\end{lstlisting}

\paragraph{Error analysis.}
The EEG branch is correct: BrainAgent selects the intended interval and correctly identifies the posterior ROI as having lower alpha power. The failure occurs in the eye-tracking branch, where \texttt{EyeTrackingAnalyzer} does not compute the requested \texttt{central\_fraction} because the required screen dimensions are missing. BrainAgent nevertheless interprets the returned \texttt{valid\_ratio=0.3892} as the central-gaze fraction and ignores the associated validity warning, leading to the opposite gaze interpretation. The correct central-gaze fraction is approximately 0.9360, indicating centrally concentrated gaze. This case therefore reflects a \emph{tool-output interpretation error}. The parser and evaluator operate correctly. \textbf{Score: 35/100.} The lower-alpha ROI metric scores 35/35, while the central-gaze fraction and semantic interpretation metrics score 0/35 and 0/30, respectively.

\paragraph{BrainAgent failure case2.}

This trace illustrates the evaluation of Sleep Assessment instance
\texttt{SA-32-Instance01} on an ISRUC overnight PSG recording. The corresponding
instruction is:

\begin{quote}
	\small\itshape
	Please use the provided sleep EEG/PSG file and paired sleep-stage label
	file to calculate the whole-night Apnea-Hypopnea Index (AHI). Count all
	apnea events and all hypopnea events across the full recording from the
	sleep signal. Compute total sleep time (TST) from the label file as the
	number of non-Wake epochs (N1, N2, N3, or R/REM) multiplied by 30 seconds,
	converted to hours. AHI is defined as (apnea\_count + hypopnea\_count)
	divided by TST in hours, reported as events per hour of total sleep time.
	Respiratory channels for ISRUC: X5 = snore signal; X6 = Flow-1,
	pressure-based airflow; DC3 = Flow-2, second airflow channel; X7 and X8 =
	abdominal respiratory-effort channels. Please clearly report the final
	AHI value.
\end{quote}

\begin{lstlisting}[style=appendixcode,escapeinside={(*@}{@*)},
	caption={Condensed BrainAgent execution trace for SA-32-Instance01 with Claude Opus 5. Planning summaries are color-coded by round.},
	label={lst:brainagent_isruc_01}]
	[Supervisor dispatch]
	Target subagent: SleepAgent
	Task: compute whole-night AHI from ISRUC PSG respiratory
	signals and paired sleep-stage labels
	
	[Planning round 1]
	(*@\textcolor{planone}{No data are registered. Load the PSG recording and sleep-stage labels. Detect apnea and hypopnea events using X6/DC3 airflow and X7/X8 respiratory-effort channels, and compute TST from the non-Wake sleep-stage epochs.}@*)
	01  RecordingLoader             [19 channels, 200 Hz, 7.33 h]
	02  SleepLabelLoader            [880 epochs, 30 s each]
	03  RespiratoryEventAnalyzer    [X6, DC3 airflow; X7, X8 effort]
	04  SleepArchitectureAnalyzer   [full-night labels]
	
	Respiratory events:
	apnea     = 130
	hypopnea  = 334
	total     = 464
	
	Sleep-stage counts:
	W=264, N1=73, N2=194, N3=231, R=118
	non-Wake epochs = 616
	TST = 308.0 min
	
	[Planning round 2]
	(*@\textcolor{plantwo}{The respiratory analyzer reports 130 apneas and 334 hypopneas, while the labels give 616 non-Wake epochs and TST=5.133 h. Compute AHI=(130+334)/5.133=90.4 events/h and report the result.} \textcolor{red}{\textbf{ERROR: the respiratory-event count is substantially overestimated; BrainAgent accepts the 464 detected events without validating the event count against the underlying respiratory evidence.}}@*)
	
	[Final output]
	AHI = 90.4 events/hour
	
	Apneas: 130
	Hypopneas: 334
	TST: 5.133 h
	
	The resulting AHI indicates a very severe level of
	sleep-disordered breathing.
\end{lstlisting}

\paragraph{Error analysis.}
The sleep-time estimation and AHI formula are correct, but the failure arises
from the respiratory-event count. \texttt{RespiratoryEventAnalyzer} reports
464 apnea-plus-hypopnea events, which substantially overestimates the reference
event burden and consequently inflates the final AHI. Accurate whole-night
respiratory-event counting is itself a challenging problem because event
detection must remain reliable over long recordings with variable signal
quality and respiratory morphology. In this case, BrainAgent accepts the
intermediate event count without further validation and propagates it directly
to the final result, representing an \emph{intermediate-result validation
	failure}. The parser and evaluator operate correctly.
\textbf{Score: 0/100.} The reported AHI of \(90.4\) events/h falls outside
the tolerance around the reference value of approximately \(32.14\) events/h.

\paragraph{CodeAct failure case1.}

This trace illustrates the evaluation of Physiological Integration instance \texttt{PI-32-Instance02} on a Simultaneous Dataset B recording. The corresponding instruction is:

\begin{quote}
	\small\itshape
	Compare \texttt{Go} (event code \texttt{16}) and \texttt{NoGo} (event code \texttt{32}) events within $[184.307, 253.202)$ seconds, using only events with complete EEG and fNIRS epochs.
	For EEG channels \texttt{FP1}, \texttt{AFz}, and \texttt{F1}, compare the baseline-corrected theta power in $[4, 8)$ Hz between the two conditions, using $[-1, 0)$ seconds as baseline and $[0, 1)$ seconds as response. Report the stronger condition as \texttt{eeg\_stronger\_condition}.
	For fNIRS pair \texttt{S03D02}, compare the baseline-corrected HbO response between the two conditions, using $[-2, 0)$ seconds as baseline and $[4, 8)$ seconds as response. Report the stronger condition as \texttt{hbo\_stronger\_condition}.
	Briefly state whether EEG and HbO favor the same condition or different conditions. Treat this as descriptive cross-modal evidence without making causal or inhibition-specific claims.
\end{quote}

\begin{lstlisting}[style=appendixcode,escapeinside={(*@}{@*)},caption={Condensed CodeAct execution trace for PI-32-Instance02 with Claude Opus 5. Execution summaries are color-coded by round.},label={lst:codeact_case32_02}]
	[Task assignment]
	Execution backend: persistent IPython kernel
	Task: compare Go and NoGo EEG theta and S03D02 HbO responses
	within [184.307, 253.202) s
	
	[Execution round 1]
	(*@\textcolor{planone}{Load the event labels as a serialized dictionary and begin the complete EEG and fNIRS analysis.} \textcolor{red}{\textbf{ERROR: the label file is an array, so .item() raises ValueError.}}@*)
	01  PythonExecutor            [np.load(...).item()]
	ValueError: can only convert an array of size 1 to a scalar
	
	[Execution round 2]
	(*@\textcolor{plantwo}{The label file is a numerical array rather than a dictionary. Inspect its structure, extract event times and codes, and load both EDF recordings.}@*)
	02  PythonExecutor            [inspect labels: shape=(128,4)]
	03  PythonExecutor            [time=column 0, code=column 2]
	04  PythonExecutor            [load EEG and fNIRS EDF files]
	
	[Execution round 3]
	(*@\textcolor{planthree}{Select codes 16 and 32 within the requested interval and retain events with complete EEG and fNIRS epochs. Six Go events and fourteen NoGo events satisfy the requirements.}@*)
	05  PythonExecutor            [filter 184.307--253.202 s]
	06  PythonExecutor            [Go=6, NoGo=14, all complete]
	
	[Execution round 4]
	(*@\textcolor{planfour}{For FP1, AFz, and F1, compute Welch theta power in 4--8 Hz for the baseline and response windows. The response-minus-baseline mean is larger for NoGo.}@*)
	07  PythonExecutor            [EEG theta calculation]
	08  PythonExecutor            [print high-precision means]
	
	Go theta mean   = -1.452706e-11
	NoGo theta mean =  8.800841e-12
	EEG stronger    = NoGo
	
	[Execution round 5]
	(*@\textcolor{planfive}{Treat the S03D02 760-nm channel as HbO. Compare its mean response-minus-baseline value between conditions; the resulting value is larger for Go.} \textcolor{red}{\textbf{ERROR: raw 760-nm intensity is not HbO; optical-density conversion and MBLL are omitted.}}@*)
	09  PythonExecutor            [S03D02_760 as HbO]
	
	fn_hbo = fn.get_data(picks=["S03D02_760"])[0]
	
	Go "HbO" mean   = -1.151377e-04
	NoGo "HbO" mean = -2.765590e-04
	HbO stronger    = Go
	
	[Execution round 6]
	(*@\textcolor{plansix}{EEG favors NoGo while the computed fNIRS quantity favors Go. Report that the two modalities favor different conditions.}@*)
	
	[Final output]
	eeg_stronger_condition: NoGo
	hbo_stronger_condition: Go
	
	EEG and fNIRS favor different conditions. Frontal theta power is
	stronger for NoGo trials, while the reported S03D02 HbO response is
	stronger for Go trials. The two modalities therefore provide
	divergent descriptive cross-modal evidence.
\end{lstlisting}

\paragraph{Error analysis.}
The event selection and EEG analysis are correct: CodeAct retains the valid Go and NoGo epochs and correctly identifies stronger frontal theta responses for NoGo. The failure occurs in the fNIRS branch, where the raw \texttt{S03D02\_760} intensity signal is incorrectly treated as HbO. Proper HbO estimation requires both 760- and 850-nm channels, conversion to optical density, and a modified Beer--Lambert transformation. Omitting these steps causes CodeAct to analyze raw intensity rather than HbO, leading to the wrong condition assignment and an incorrect cross-modal conclusion. The files, execution environment, parser, and evaluator operate normally; the error originates from the model-generated analysis code. \textbf{Score: 30/100.} The EEG condition metric scores 30/30, while the HbO condition and cross-modal interpretation metrics score 0/30 and 0/40, respectively.

\paragraph{CodeAct failure case2.}

This trace illustrates the evaluation of Sleep Assessment instance
\texttt{SA-34-Instance05} on an SHHS recording. The corresponding instruction is:

\begin{quote}
	\small\itshape
	Use the provided sleep EEG/PSG recording and paired sleep-stage labels to
	identify the longest apnea event over the complete recording. Report its
	duration in seconds, its start time in minutes from the beginning of the
	recording, and the sleep stage at event onset. The stage must be one of
	\texttt{W}, \texttt{N1}, \texttt{N2}, \texttt{N3}, or \texttt{R}.
	The available respiratory channels are \texttt{THOR RES},
	\texttt{ABDO RES}, \texttt{NEW AIR}, and \texttt{SaO2}.
\end{quote}

\begin{lstlisting}[style=appendixcode,escapeinside={(*@}{@*)},caption={Condensed CodeAct execution trace for SA-34-Instance05 with Claude Opus 5. Execution summaries are color-coded by round.},label={lst:codeact_case34_05}]
	[Task assignment]
	Execution backend: persistent IPython kernel
	Task: identify the longest apnea event and report its duration,
	start time, and sleep stage at onset
	
	[Execution round 1]
	(*@\textcolor{planone}{Inspect the sleep-stage labels, load the EDF recording, and identify the respiratory channels and their sampling rate.}@*)
	01  PythonExecutor            [load sleep-stage labels]
	02  PythonExecutor            [load EDF recording]
	03  PythonExecutor            [inspect respiratory channels]
	
	Recording duration = 32519.992 s
	Sampling frequency = 125 Hz
	Sleep-stage epochs = 1084
	
	[Execution round 2]
	(*@\textcolor{plantwo}{Construct a per-second airflow-amplitude signal from NEW AIR and compare it with a local rolling baseline to identify sustained reductions.}@*)
	04  PythonExecutor            [filter and downsample NEW AIR]
	05  PythonExecutor            [compute airflow envelope]
	06  PythonExecutor            [estimate rolling baseline]
	07  PythonExecutor            [group consecutive low-amplitude seconds]
	
	Threshold   Events   Maximum duration
	0.10        0        0 s
	0.20        0        0 s
	0.25        0        0 s
	0.30        1        25 s
	
	[Execution round 3]
	(*@\textcolor{planthree}{The initial detector finds no event under the strict thresholds. Test alternative rolling baselines and inspect airflow, respiratory-effort, and oximetry signals.}@*)
	08  PythonExecutor            [test five-minute percentile baseline]
	09  PythonExecutor            [recompute airflow amplitude]
	10  PythonExecutor            [inspect THOR RES and ABDO RES]
	11  PythonExecutor            [detect SaO2 desaturations]
	
	Trailing-baseline threshold 0.20: maximum duration = 29 s
	Trailing-baseline threshold 0.25: maximum duration = 30 s
	Trailing-baseline threshold 0.30: maximum duration = 34 s
	
	[Execution round 4]
	(*@\textcolor{planfour}{Bandpass-filter NEW AIR, compute a Hilbert envelope, average it into one-second bins, and retain runs below 30\% of a trailing median baseline.}@*)
	12  PythonExecutor            [bandpass NEW AIR at 0.05--1 Hz]
	13  PythonExecutor            [compute and smooth Hilbert envelope]
	14  PythonExecutor            [apply 30\% rolling-baseline threshold]
	
	Candidate events:
	start=8147 s,  duration=34 s
	start=28061 s, duration=15 s
	start=3 s,     duration=12 s
	
	(*@\textcolor{red}{\textbf{ERROR: the detector relaxes the airflow criterion to 30\% of the rolling baseline and quantizes event boundaries into one-second bins. This heuristic is not boundary-equivalent to the annotated apnea interval and underestimates its duration.}}@*)
	
	[Execution round 5]
	(*@\textcolor{planfive}{Inspect the longest candidate, verify the associated desaturation and respiratory-effort reduction, and map its onset to the paired sleep-stage label.}@*)
	15  PythonExecutor            [inspect SaO2 around 8147 s]
	16  PythonExecutor            [inspect thoracic effort]
	17  PythonExecutor            [map onset to sleep-stage epoch]
	
	Detected start = 8147 s = 135.78 min
	Detected duration = 34.0 s
	Sleep stage = R
	
	[Execution round 6]
	(*@\textcolor{plansix}{Report the longest detected airflow-cessation event together with its onset time and sleep stage.}@*)
	
	[Final output]
	duration_sec: 34.0
	start_time_min: 135.78
	sleep_stage: R
\end{lstlisting}

\paragraph{Error analysis.}
CodeAct correctly identifies the target event region, onset time, and sleep
stage; the failure is confined to event duration. The model-generated detector
uses relaxed airflow thresholds and one-second temporal aggregation, which
coarsens the event boundaries and yields a duration of 34.0 s instead of the
39.4-s reference interval. This case therefore reflects a \emph{temporal
	discretization error} introduced by the model-generated detection procedure.
The files, execution environment, parser, and evaluator operate normally.
\textbf{Score: 60/100.} The duration metric scores 0/40, while the onset-time
and sleep-stage metrics score 40/40 and 20/20, respectively.

\section{Limitations and Future Work}

Evaluation cost is a key practical consideration in large-scale LLM benchmarking. Inspired by lightweight benchmark maintenance strategies such as HeaRT-Lighting, we sample five subjects per dataset to balance subject diversity with evaluation efficiency. This design keeps large-scale evaluation tractable while retaining broad coverage across datasets, recordings, and analytical tasks. Nevertheless, the resulting subject coverage remains insufficient to characterize population-level variability. Future releases could adopt more scalable evaluation tracks that broaden subject coverage without substantially increasing evaluation cost.
More broadly, we plan to extend \benchmarkname{} toward a more comprehensive benchmark for brain-science understanding by incorporating richer analytical tasks and additional neural modalities, including MEG and fMRI. Such extensions will enable systematic evaluation of whether LLM-based systems can generalize analytical competence across modalities, cohorts, experimental settings, and scientific workflows.

\end{document}